%% file: ICML.tex
\documentclass{article}

\usepackage{microtype}
\usepackage{graphicx}
\usepackage{subfigure}
\usepackage{booktabs} 
\usepackage{wrapfig}
\usepackage{bbm, dsfont}

\usepackage{hyperref}

 \usepackage[accepted]{icml2023}

\usepackage{amsmath}
\usepackage{amssymb}
\usepackage{mathtools}
\usepackage{amsthm}
\newtheorem{hyp}{Hypothesis}
\newtheorem{prop}{Proposition}
\usepackage{float}

\definecolor{gd_green}{rgb}{0.111766,0.563365,0.388828}
\definecolor{tf_purple}{rgb}{0.384629,0.350935,0.640238}

\input{math_commands.tex}

\usepackage[capitalize,noabbrev]{cleveref}

\theoremstyle{plain}

\theoremstyle{definition}

\theoremstyle{remark}

\usepackage[textsize=tiny]{todonotes}

\icmltitlerunning{Transformers Learn In-Context by Gradient Descent}

\begin{document}

\twocolumn[
\icmltitle{Transformers Learn In-Context by Gradient Descent}
\vspace{-10pt}


\icmlsetsymbol{equal}{*}

\begin{icmlauthorlist}
\icmlauthor{Johannes von Oswald}{ethz,comp}
\icmlauthor{Eyvind Niklasson}{comp}
\icmlauthor{Ettore Randazzo}{comp}
\icmlauthor{João Sacramento}{ethz}\\
\icmlauthor{Alexander Mordvintsev}{comp}
\icmlauthor{Andrey Zhmoginov}{comp}
\icmlauthor{Max Vladymyrov}{comp}
\end{icmlauthorlist}

\icmlaffiliation{ethz}{Department of Computer Science, ETH Zürich, Zürich, Switzerland}
\icmlaffiliation{comp}{Google Research}

\icmlcorrespondingauthor{Johannes von Oswald}{voswaldj@ethz.ch}

\icmlkeywords{Transformers, in-context learning, mechansitic interpretability, mesa-optimization, Machine Learning, ICML}

\vskip 0.3in
]



\printAffiliationsAndNotice{}  

\begin{abstract}

At present, the mechanisms of in-context learning in Transformers are not well understood and remain mostly an intuition. In this paper, we suggest that training Transformers on auto-regressive objectives is closely related to gradient-based meta-learning formulations. We start by providing a simple weight construction that shows the equivalence of data transformations induced by 1) a single linear self-attention layer and by 2) gradient-descent (GD) on a regression loss. Motivated by that construction, we show empirically that when training self-attention-only Transformers on simple regression tasks either the models learned by GD and Transformers show great similarity or, remarkably, the weights found by optimization match the construction. Thus we show how trained Transformers become mesa-optimizers i.e. learn models by gradient descent in their forward pass. This allows us, at least in the domain of regression problems, to mechanistically understand the inner workings of in-context learning in optimized Transformers. Building on this insight, we furthermore identify how Transformers surpass the performance of plain gradient descent by learning an iterative curvature correction and learn linear models on deep data representations to solve non-linear regression tasks. Finally, we discuss intriguing parallels to a mechanism identified to be crucial for in-context learning termed \textit{induction-head} \citep{induction_heads} and show how it could be understood as a specific case of in-context learning by gradient descent learning within Transformers.
\end{abstract}

\section{Introduction}

In recent years Transformers \citep[TFs;][]{transformers} have demonstrated their superiority in numerous benchmarks and various fields of modern machine learning,
and have emerged as the \textit{de-facto} neural network architecture used for modern AI \citep{transformers_vision, NEURIPS2019_9d63484a, https://doi.org/10.48550/arxiv.2005.12872, convormer}. It has been hypothesised that their success is due in part to a phenomenon called \textit{in-context learning}~\citep{transformers_few_shot,pre_train_prompt}: an ability to flexibly adjust their prediction based on additional data given \emph{in context} (i.e.\ in the input sequence itself). In-context learning offers a seemingly different approach to few-shot and meta-learning \citep{transformers_few_shot}, but as of today the exact mechanisms of how it works are not fully understood.
It is thus of great interest to understand what makes Transformers pay attention to their context, what the mechanisms are, and under which circumstances, they come into play \citep{data_dis_in_context, induction_heads}.

\begin{figure*}
\vspace{-5pt}
\hspace{20pt}
\begin{minipage}{.63\textwidth}
    \includegraphics[width=1.\textwidth]{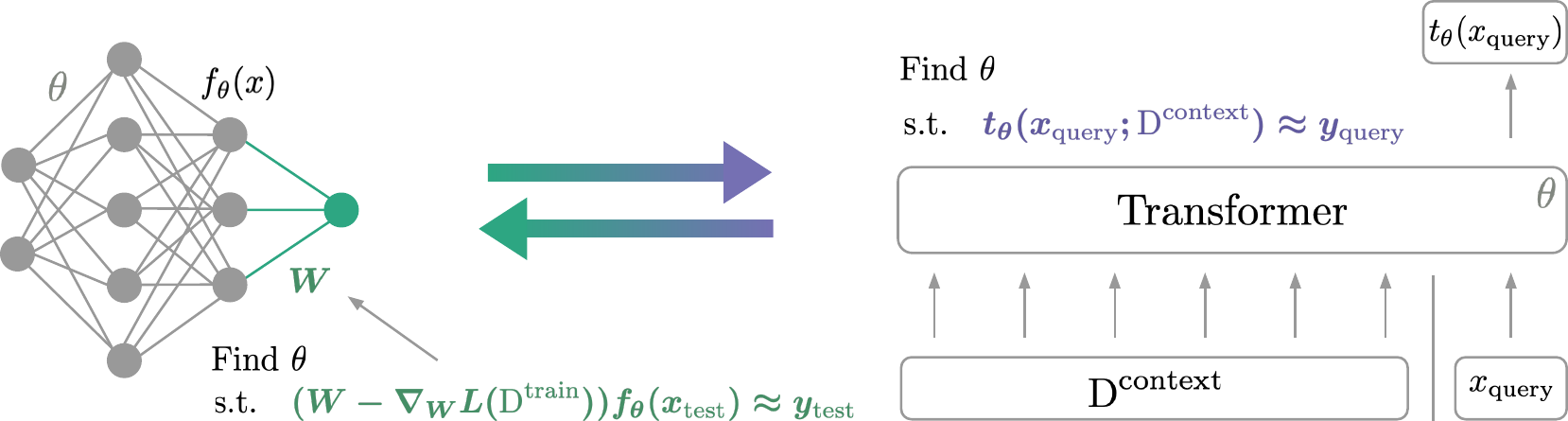}
\end{minipage}
\hspace{10pt}
\begin{minipage}{.3\textwidth}
    \includegraphics[width=.99\textwidth]{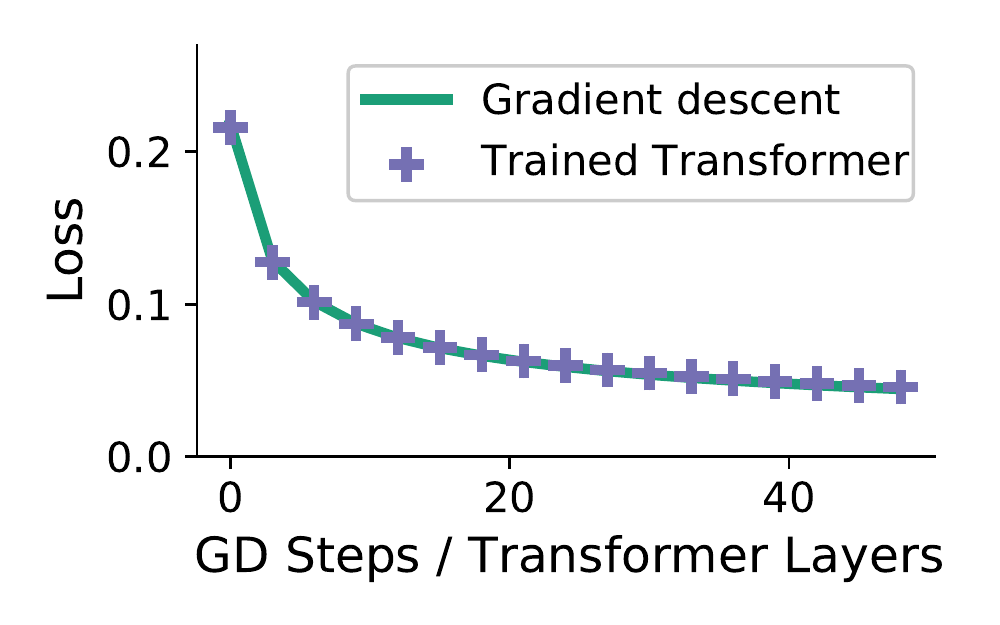}
\end{minipage}
\vspace{-5pt}
  \caption{\textbf{Illustration of our hypothesis: {\color{gd_green} gradient-based optimization} and {\color{tf_purple} attention-based in-context learning} are equivalent.} \textit{Left}: Learning a neural network output layer by gradient descent on a dataset $\text{D}^{\text{train}}$. The task-shared meta-parameters $\theta$ are obtained by meta-learning with the goal that after adjusting the neural network output layer, the model generalizes well on unseen data. \textit{Center}: Illustration of a Transformer that adjusts its query prediction on the data given in-context i.e.\ $t_{\theta}(x_{\text{query}}; \text{D}^{\text{context}})$. The weights of the Transformer are optimized to predict the next token $y_{\text{query}}$.
  \textit{Right}: Our results confirm the hypothesis that learning with $K$ steps of gradient descent on a dataset $\text{D}^{\text{train}}$ (\emph{green part of the left plot}) matches trained Transformers with $K$ linear self-attention layers (\emph{central plot}) when given $\text{D}^{\text{train}}$ as in-context data $\text{D}^{\text{context}}$.}
  \label{fig:illus}
  \vspace{-13pt}
\end{figure*}
 
In this paper, we aim to bridge the gap between in-context and meta-learning, and show that in-context learning in Transformers can be an emergent property approximating gradient-based few-shot learning within its forward pass, see Figure \ref{fig:illus}. For this to be realized, we show how Transformers \textbf{(1)} construct a loss function dependent on the data given in sequence and \textbf{(2)} learn based on gradients of that loss. We will first focus on the latter, the more elaborate learning task, in sections \ref{sect:construction} and \ref{sect:trained_tf}, after which we provide evidence for the former in section \ref{sect:softmax-builds-tokens}.

We summarize our contributions as follows\footnote{Main experiments can be reproduced with notebooks provided under the following link: \url{https://github.com/google-research/self-organising-systems/tree/master/transformers_learn_icl_by_gd}}:
\begin{itemize}
    \item We construct explicit weights for a linear self-attention layer that induces an update identical to a single step of gradient descent (GD) on a mean squared error loss. Additionally, we show how several self-attention layers can iteratively perform curvature correction improving on plain gradient descent.
    \item When optimized on linear regression datasets, we demonstrate that linear self-attention-only Transformers either converge to our weight construction and therefore implement gradient descent, or generate linear models that closely align with models trained by GD, both in in- and out-of-distribution validation tasks.
    \item By incorporating multi-layer-perceptrons (MLPs) into the Transformer architecture, we enable solving \textit{non}linear regression tasks within Transformers by showing its equivalence to learning a linear model on deep representations. We discuss connections to kernel regression as well as nonparametric kernel smoothing methods. Empirically, we compare meta-learned MLPs and a single step of GD on its output layer with trained Transformers and demonstrate striking similarities between the identified solutions.
    \item We resolve the dependency on the specific token construction by providing evidence that learned Transformers first encode incoming tokens into a format amenable to the in-context gradient descent learning that occurs in the later layers of the Transformer.
\end{itemize}

These findings allow us to connect \textit{learning} Transformer weights and the concept of \textit{meta-learning} a learning algorithm \citep{schmidhuber_evolutionary_1987,Hinton1987UsingFW,bengio_learning_1990, chalmers_evolution_1991,fast_weights,thrun_learning_1998,meta_hochreiter,andrychowicz_learning_2016, hinton_fast_weights, kirsch2021meta}. In this extensive research field,  meta-learning is typically regarded as learning that takes place on various time scales namely fast and slow. The slowly changing parameters control and prepare for fast adaptation reacting to sudden changes in the incoming data by e.g. a context switch. Notably, we build heavily on the concept of fast weights \cite{fast_weights} which has shown to be equivalent to linear self-attention \cite{linear_transformers_fast_weight} and show how optimized Transformers implement interpretable learning algorithms within their weights.

Another related meta-learning concept, termed MAML \cite{finn_model-agnostic_2017}, aims to meta-learn a deep neural network initialization which allows for fast adaptation on novel tasks. It has been shown that in many circumstances, the solution found can be approximated well when only adapting the output layer i.e. learning a linear model on a meta-learned deep data representations \citep{finn_model-agnostic_2017, finn2018metalearning_universal,DBLP:conf/iclr/GordonBBNT19,meta_opt_net,rusu2019metalearning, raghu_rapid_2020, von_oswald_learning_2021}. 
In section \ref{sect:trained_tf}, we show the equivalence of this framework to in-context learning implemented in a common Transformer block i.e. when combining self-attention layers with a multi-layer-perceptron. 

In the light of meta-learning we show how optimizing Transformer weights can be regarded as learning on two time scales. More concretely, we find that solely through the pressure to predict correctly Transformers discover learning algorithms inside their forward computations, effectively meta-learning a learning algorithm.
Recently, this concept of an emergent optimizer within a learned neural network, such as a Transformer, has been termed ``mesa-optimization'' \citep{mesa}. We find and describe one possible realization of this concept and hypothesize that the in-context learning capabilities of language models emerge through mechanisms similar to the ones we discuss here.

Transformers come in different ``shapes and sizes'', operate on vastly different domains, and exhibit varying forms of phase transitions of in-context learning \citep{kirsch2022generalpurpose, Trans_generalize_differently_weights}, suggesting variance and significant complexity of the underlying learning mechanisms. As a result, we expect our findings on linear self-attention-only Transformers to only explain a limited part of a complex process, and it may be one of many possible methods giving rise to in-context learning. Nevertheless, our approach provides an intriguing perspective on, and novel evidence for, an in-context learning mechanism that significantly differs from existing mechanisms based on associative memory \citep{hopfield}, or by the copying mechanism termed \textit{induction heads} identified by \cite{induction_heads}. We, therefore, state the following 

\begin{hyp}[Transformers learn in-context by gradient descent]
\label{prop:hyp}
When training Transformers on auto-regressive tasks, in-context learning in the Transformer forward pass is implemented by gradient-based optimization of an implicit auto-regressive inner loss constructed from its in-context data.
\vspace{-5pt}
\end{hyp}

We acknowledge work done in parallel, investigating the same hypothesis. \citet{related_work} puts forward a weight construction based on a chain of Transformer layers (including MLPs) that together implement a single step of gradient descent with weight decay. Similar to work done by \citet{simple_case_study}, they then show that trained Transformers match the performance of models obtained by gradient descent. Nevertheless, it is not clear that optimization finds Transformer weights that coincide with their construction. 

Here, we present a much simpler construction that builds on \citet{linear_transformers_fast_weight} and \emph{only requires a single linear self-attention layer} to implement a step of gradient descent. 
This allows us to (1) show that optimizing self-attention-only Transformers finds weights that match our weight construction (Proposition \ref{prop:self_att_gd}), demonstrating its practical relevance, and (2) explain in-context learning in shallow two layer Transformers intensively studied by \citet{induction_heads}. Therefore, although related work provides comprehensive empirical evidence that Transformers indeed seem to implement gradient descent based learning on the data given in-context, we will in the following present mechanistic verification of this hypothesis and provide compelling evidence that our construction, which implements GD in a Transformer forward pass, is found in practice.

\section{Linear self-attention \textit{can} 
emulate gradient descent on a linear regression task}
\label{sect:construction}

We start by reviewing a standard multi-head self-attention (SA) layer with parameters $\theta$. A SA layer updates each element $e_j$ of a set of tokens $\{e_1,\ldots,e_N\}$ according to
\begin{align}
\begin{split}
\label{eq:trans}
e_j &\leftarrow e_j + \sattn_{\theta}(j, \{e_1,\ldots,e_N\}) \\ &= e_j + \sum_{h} P_hV_h \text{softmax}(K_h^{T}q_{h,j})
\end{split}
\end{align}
with $P_h, V_h, K_h$ the projection, value and key matrices, respectively, and $q_{h,i}$ the query, all for the $h$-th head. To simplify the presentation, we omit bias terms here and throughout. The columns of the value
$V_h  = [v_{h,1}, \dots, v_{h,N}]$ and key $K_h=[k_{h,1}, \dots, k_{h,N}]$ matrices consist of vectors $v_{h,i} = W_{h,V}e_i$ and  $k_{h,i} = W_{h,K}e_i$; likewise, the query is produced by linearly projecting the tokens, $q_{h,j} = W_{h,Q}e_j$. The parameters  $\theta = \{P_h, W_{h,V}, W_{h,K}, W_{h,Q}\}_h$ of a SA layer consist of all the projection matrices, of all heads.

The self-attention layer described above corresponds to the one used in the standard Transformer model. Following \citet{linear_transformers_fast_weight}, we now introduce our first (and only) departure from the standard model, and omit the $\text{softmax}$ operation in \eqref{eq:trans}, leading to the \textit{linear} self-attention (LSA) layer $e_j  \leftarrow e_j + \lsattn_{\theta}(j, \{e_1,\ldots,e_N\}) = e_j + \sum_{h} P_hV_hK_h^{T}q_{h,j}$
We next show that with some simple manipulations we can relate the update performed by an LSA layer to one step of gradient descent on a linear regression loss.

\subsection*{Data transformations induced by gradient descent}
 
We now introduce a reference linear model $y(x) = W x$ parameterized by the weight matrix $W \in \mathbb{R}^{N_y \times N_x}$, and a training dataset $D = \{(x_i, y_i)\}_{i=1}^{N}$ comprising of input samples $x_i \in \mathbb{R}^{N_x}$ and respective labels $y_i \in \mathbb{R}^{N_y}$. The goal of learning is to minimize the squared-error loss:
\begin{equation}
\label{eq:lin_reg}
L(W) = \frac{1}{2N}\sum_{i=1}^{N} \|Wx_i - y_i\|^2. 
\end{equation}
One step of gradient descent on $L$ with learning rate $\eta$ yields the weight change
\begin{align}
\label{eq:gd_lin_reg}
    \Delta W = - \eta \nabla_W L(W) = - \frac{\eta}{N} \sum_{i=1}^{N}(Wx_i - y_i)x_i^T.
\end{align} Considering the loss after changing the weights, we obtain
\begin{align}
\begin{split}
\label{eq:lin_reg_changed}
L(W + \Delta W) &= \frac{1}{2N} \sum_{i=1}^{N}\left\|(W + \Delta W)x_i - y_i\right\|^2 \\
&= \frac{1}{2N} \sum_{i=1}^{N} \|Wx_i  - (y_i - \Delta y_i)\|^2 
\end{split}
\end{align}
where we introduced the transformed targets $y_i - \Delta y_i$ with $\Delta y_i = \Delta W x_i$. Thus, we can view the outcome of a gradient descent step as an update to our regression loss (\eqref{eq:lin_reg}), where data, and not weights, are updated. Note that this formulation is closely linked to predicting based on nonparametric kernel smoothing, see Appendix \ref{app:ker_reg} for a discussion.

Returning to self-attention mechanisms and Transformers, we consider an in-context learning problem where we are given $N$ context tokens together with an extra query token, indexed by $N+1$. In terms of our linear regression problem, the $N$ context tokens $e_j = (x_j, y_j) \in \mathbb{R}^{N_x+N_y}$ correspond to the $N$ training points in $D$, and the $N$+1-th token $e_{N+1} = (x_{N+1}, y_{N+1}) = (x_{\text{test}}, \hat{y}_{\text{test}}) = e_{\text{test}}$ to the test input $x_{\text{test}}$ and the corresponding
prediction $\hat{y}_{\text{test}}$. We use the terms training and in-context data interchangeably, as well as query and test token/data, as we establish their equivalence now. 

\subsection*{Transformations induced by gradient descent and a linear self-attention layer can be equivalent}

We have re-cast the task of learning a linear model as directly modifying the data, instead of explicitly computing and returning the weights of the model (\eqref{eq:lin_reg_changed}). We proceed to establish a connection between self-attention and gradient descent. We provide a construction where learning takes place simultaneously by directly updating all tokens, including the test token, through a linear self-attention layer. In other words, the token produced in response to a query (test) token is transformed from its initial value $W_0 x_{\text{test}}$, where $W_0$ is the initial value of $W$, to the post-learning prediction $\hat{y} = (W_0 + \Delta W) x_{\text{test}}$ obtained after one gradient descent step.

\begin{prop}
\label{prop:self_att_gd}
Given a 1-head linear attention layer and the tokens $e_j = (x_j, y_j)$, for $j=1,\ldots,N$, one can construct key, query and value matrices $W_K, W_Q, W_V$ as well as the projection matrix $P$ such that a Transformer step on every token $e_j$ is identical to the gradient-induced dynamics $ e_j \leftarrow (x_j, y_j) + (0, -\Delta W x_j) = (x_j, y_j) + P \,V K^{T}q_{j}$ such that $e_j = (x_j, y_j - \Delta y_j)$. For the test data token $(x_{N+1}, y_{N+1})$ the dynamics are identical.
\end{prop}
The simple construction can be found in Appendix~\ref{app:prop1} and we denote the corresponding self-attention weights by  $\theta_{\text{GD}}$.

Below, we provide some additional insights on what is needed to implement the provided LSA-layer weight construction, and further details on what it can achieve:
\begin{itemize}
\item \textbf{Full self-attention.} Our dynamics model training is based on in-context tokens only, i.e., only $e_1, \ldots, e_N$ are used for computing key and value matrices; the query token $e_{N+1}$ (containing test data) is excluded. This leads to a linear function in $x_{\text{test}}$ as well as to the correct $\Delta W$, induced by gradient descent on a loss consisting only of the training data. This is a minor deviation from full self-attention. In practice, this modification can be dropped, which corresponds to assuming that the underlying initial weight matrix is zero, $W_0 \approx 0$, which makes $\Delta W$ in equation \ref{eq:trans_update} independent of the test token even if incorporating it in the key and value matrices. In our experiments, we see that these assumptions are met when initializing the attention weights $\theta$ to small values.

\item \textbf{Reading out predictions.} When initializing the $y$-entry of the test-data token with $-W_0 x_{N+1}$, i.e.~$e_\text{test} = (x_{\text{test}}, -W_0 x_{\text{test}})$, the test-data prediction $\hat{y}$ can be easily read out by simply multiplying again by $-1$ the updated token, since  $-y_{N+1} + \Delta y_{N+1} = - (y_{N+1} - \Delta y_{N+1}) = y_{N+1} + \Delta W x_{N+1}$. This can easily be done by a final projection matrix, which incidentally is usually found in Transformer architectures. Importantly, we see that a single head of self-attention is sufficient to transform our training targets as well as the test prediction simultaneously. 

\item \textbf{Uniqueness.} We note that the construction is not unique; in particular, it is only required that the products $PW_V$ as well as $W_KW_Q$ match the construction. Furthermore, since no nonlinearity is present,  any rescaling $s$ of the matrix products, i.e., $PW_Vs$  and $W_KW_Q/s$, leads to an equivalent result. If we correct for these equivalent formulations, we can experimentally verify that weights of our learned Transformers indeed match the presented construction. 

\item \textbf{Meta-learned task-shared learning rates.} When training self-attention parameters $\theta$ across a family of in-context learning tasks $\tau$, where the data $(x_{\tau,i},y_{\tau,i})$ follows a certain distribution, the learning rate can be implicitly (meta-)learned such that an optimal loss reduction (averaged over tasks) is achieved given a fixed number of update steps. In our experiments, we find this to be the case. This kind of meta-learning to improve upon plain gradient descent has been leveraged in numerous previous approaches for deep neural networks   \citep{DBLP:journals/corr/LiZCL17,lee2018gradient,DBLP:conf/nips/ParkO19,zhao_meta_learning_hypernetworks,flennerhag2020metalearning}.

\item \textbf{Task-specific data transformations.} A self-attention layer is in principle further capable of exploiting statistics in the current training data samples, beyond modeling  task-shared curvature information in $\theta$. More concretely, a LSA layer updates an input sample according to a data transformation $x_j \leftarrow x_j+ \Delta x_j = (I + P(X)V(X)K(X)^TW_Q)x_j = H_\theta(X) x_j$, with $X$ the $N_x \times N$ input training data matrix, when neglecting influences by target data $y_i$. Through $H_\theta(X)$, a LSA layer can encode in $\theta$ an algorithm for carrying out data transformations which depend on the actual input training samples in $X$. In our experiments, we see that trained self-attention learners employ a simple form of $H(X)$ and that this leads to substantial speed ups in for GD and TF learning.
\end{itemize}

\section{Trained Transformers \textit{do} mimic gradient descent on linear regression tasks}
\label{sect:trained_tf}

\begin{figure*}
\vspace{-7pt}
\centering
\begin{minipage}{.23\textwidth}
  \centering
  \begin{center}
    \includegraphics[width=1.\textwidth]{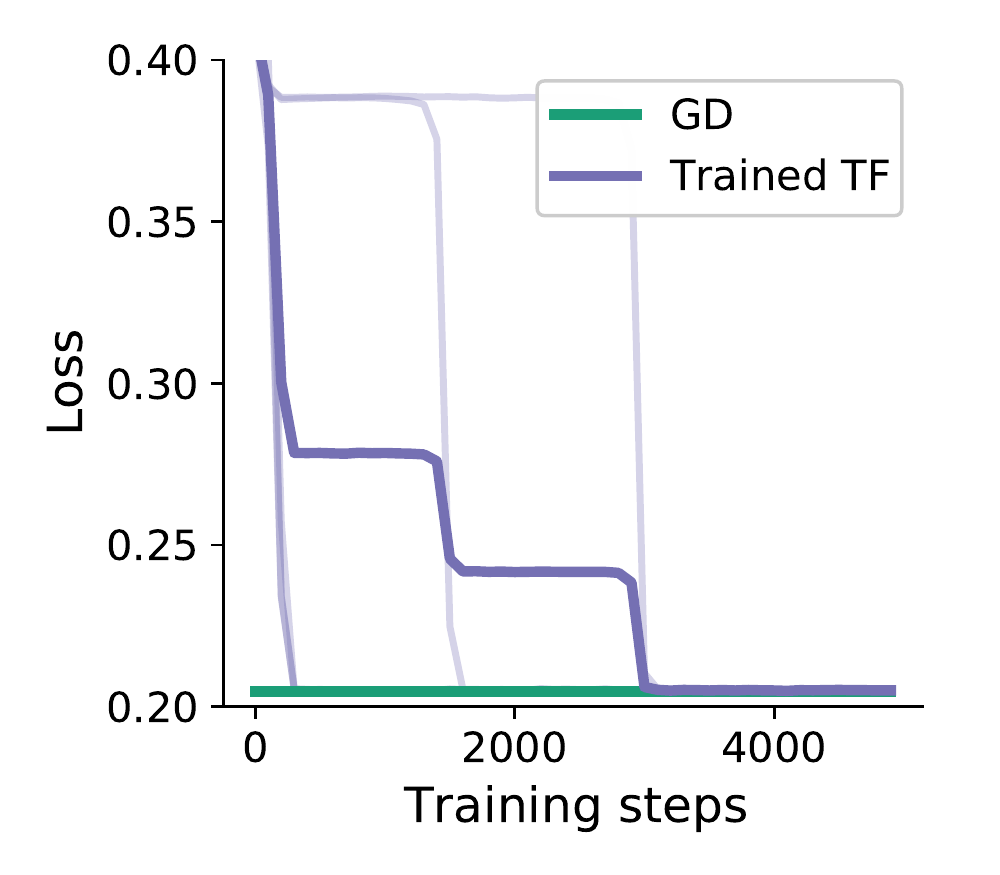}
  \end{center}
  \vspace{-10pt}
\end{minipage}
\begin{minipage}{.28\textwidth}
  \centering
  \begin{center}
    \includegraphics[width=1.\textwidth]{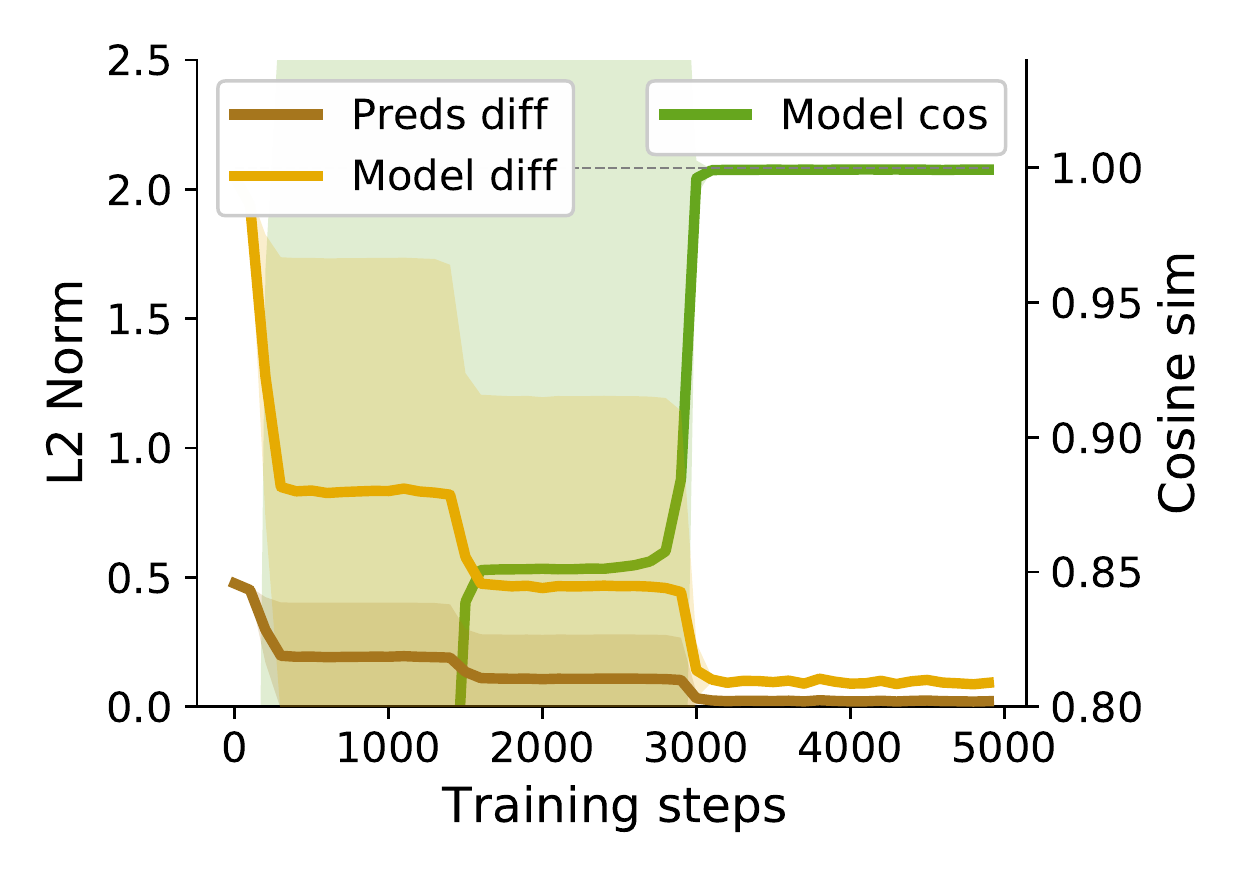}
  \end{center}
  \vspace{-10pt}
\end{minipage}
\begin{minipage}{.23\textwidth}
  \centering
  \begin{center}
    \includegraphics[width=1.\textwidth]{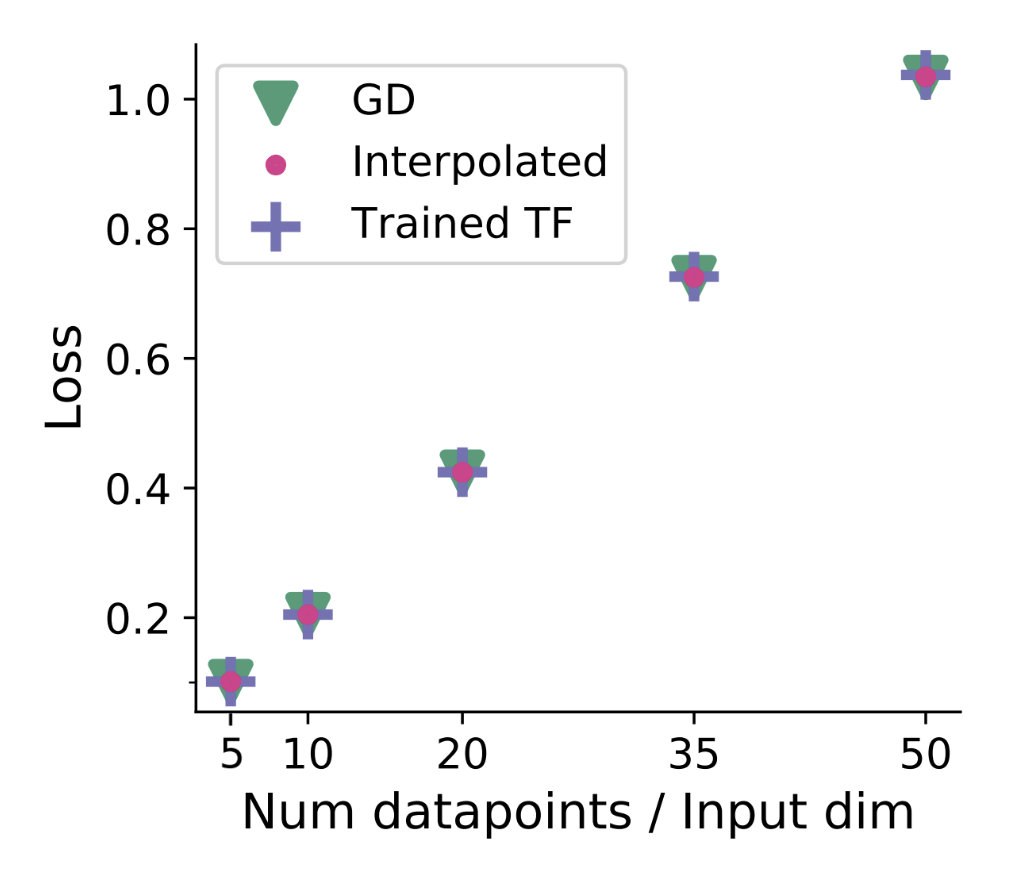}
  \end{center}
  \vspace{-10pt}
\end{minipage}
\begin{minipage}{.23\textwidth}
  \centering
  \begin{center}
    \includegraphics[width=1.\textwidth]{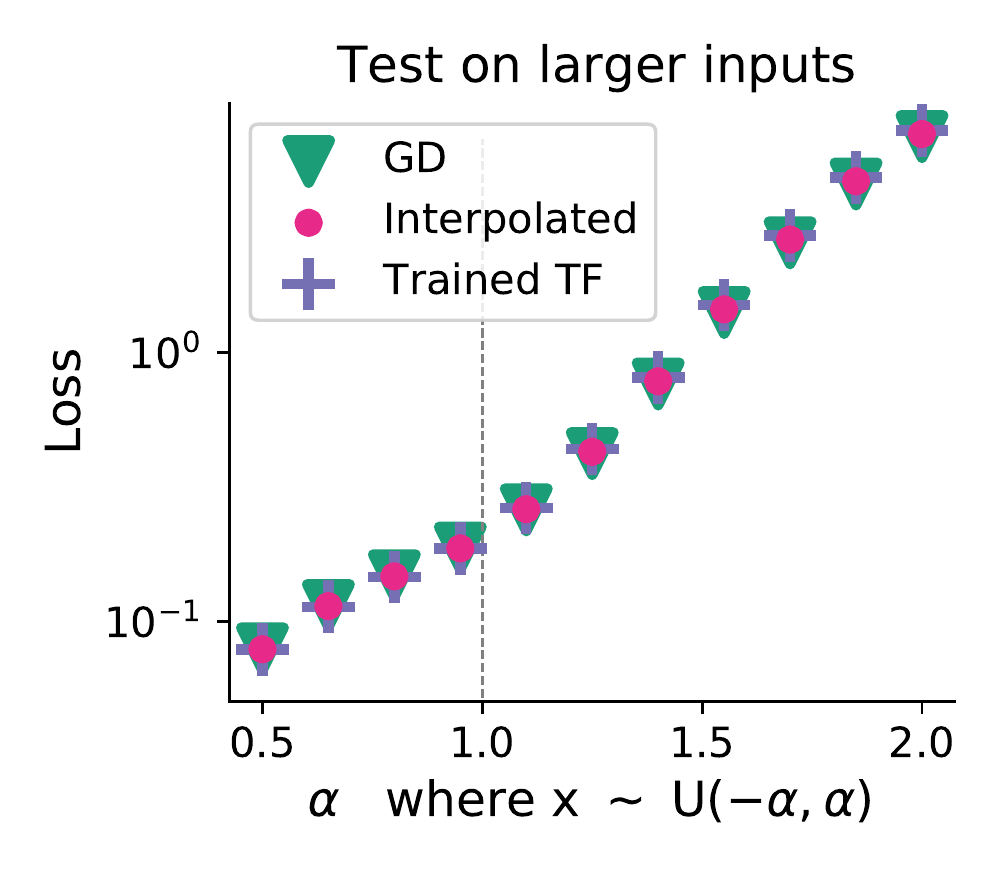}
  \end{center}
  \vspace{-10pt}
\end{minipage}
\hspace{-0pt}
  \caption{\textbf{Comparing one step of GD with a trained single linear self-attention layer.} \textit{Outer left:} Trained single LSA layer
  performance is identical to the one of gradient descent. 
  \textit{Center left}: Almost perfect alignment of GD and the model generated by the SA layer after training, measured by cosine similarity and the L2 distance between models as well as their predictions. \textit{Center right:} Identical loss of GD, the LSA layer model as well as the model obtained by interpolating between the construction and the optimized LSA layer weights for different $N=N_x$.
  \textit{Outer right:} The trained LSA layer, gradient descent and their interpolation show identically loss (in log-scale) when provided input data different than during training i.e. with scale of 1. We display the mean/std. or the single runs of 5 seeds.}
  \label{fig:training_trans_lin}
   \vspace{-13pt}
\end{figure*}

We now experimentally investigate whether trained attention-based models implement gradient-based in-context learning in their forward passes. We gradually build up from single linear self-attention layers to multi-layer nonlinear models, approaching full Transformers. In this section, we follow the assumption of Proposition \ref{prop:self_att_gd} tightly and construct our tokens by concatenating input and target data, $e_j = (x_j, y_j)$ for $1\leq j \leq N$, and our query token by concatenating the test input and a zero vector, $e_{N+1} = (x_{\text{test}}, 0)$. We show how to lift this assumption in the last section of the paper. The prediction $\hat{y}_{\theta}(\{e_{\tau,1}, \ldots, e_{\tau,N}\}, e_{\tau, N+1})$ of the attention-based model, which depends on all tokens and on the parameters $\theta$, is read-out from the $y$-entry of the updated $N+1$-th token as explained in the previous section. 

The objective of training, visualized in Figure \ref{fig:illus}, is to minimize the expected squared prediction error, averaged over tasks $\min_\theta \mathbb{E}_\tau [||\hat{y}_{\theta}(\{e_{\tau,1}, \ldots, e_{\tau,N}\}, e_{\tau, N+1}) - y_{\tau,\text{test}}||^2]$. We achieve this by minibatch online minimization (by Adam \citep{adam}): At every optimization step, we construct a batch of novel training tasks and take a step of stochastic gradient descent on the loss function:
\begin{equation}
\label{eq:trans_loss}
     \mathcal{L}(\theta) = \frac{1}{B} \sum_{\tau=1}^B ||\hat{y}_{\theta}(\{e_{\tau,i}\}_{i=1}^N, e_{\tau, N+1}) - y_{\tau,\text{test}}||^2
\end{equation}
where each task (context) $\tau$ consists of in-context training data  $D_\tau=\{(x_{\tau,i}, y_{\tau,i})\}_{i=1}^{N}$ and test point $(x_{\tau,N+1}, y_{\tau,N+1})$, which we use to construct our tokens $\{e_{\tau,i}\}_{i=1}^{N+1}$ as described above. We denote the optimal parameters found by this optimization process by $\theta^*$. In our setup, finding $\theta^*$ may be thought of as meta-learning, while learning a particular task $\tau$ corresponds to simply evaluating the model $\hat{y}_{\theta}(\{e_{\tau,1}, \ldots, e_{\tau,N}\}, e_{\tau, N+1})$. Note that we therefore never see the exact same training task twice during training. See Appendix~\ref{app:ex_det}, especially Figure \ref{fig:cycling} for an analyses when using a fixed dataset size which we cycle over during training.

We focus on solvable tasks and similarly to \citet{simple_case_study} generate data for each task using a teacher model with parameters $W_\tau \sim \mathcal{N}(0, I)$. We then sample $x_{\tau,i} \sim U(-1, 1)^{n_I}$ and construct targets using the task-specific teacher model, $y_{\tau,i} = W_\tau x_{\tau,i}$. In the majority of our experiments we set the dimensions to $N=n_I=10$ and $n_O=1$. Since we use a noiseless teacher for simplicity, we can expect our regression tasks to be well-posed and analytically solvable as we only compute a loss on the Transformers last token, which stands in contrast to usual autoregressive training and the training setup of \citet{simple_case_study}. 
Full details and results for training with a fixed training set size may be found in Appendix~\ref{app:ex_det}.

\subsection*{One-step of gradient descent vs.~a single trained self-attention layer}

Our first goal is to investigate whether a trained single, linear self-attention layer can be explained by the provided weight construction that implements GD. To that end, we compare the predictions made by a LSA layer with trained weights $\theta^*$ (which minimize~\eqref{eq:trans_loss}) and with constructed weights $\theta_\text{GD}$ (which satisfy~Proposition~\ref{prop:self_att_gd}).

Recall that a LSA layer yields the prediction $\hat{y}_{\theta}(x_{\text{test}}) = e_{N+1} + \lsattn_{\theta}(\{e_1,\ldots,e_N\},e_{N+1}) = \Delta W_{\theta,D} x_{\text{test}}$, which is linear in $x_{\text{test}}$. We denote by $\Delta W_{\theta,D}$ the matrix generated by the LSA layer following the construction provided in Proposition \ref{prop:self_att_gd}, with query token $e_{N+1}$ set such that the initial prediction is set to zero, $\hat{y}_{\text{test}} = 0$. We compare $\hat{y}_{\theta}(x_{\text{test}})$ to the prediction of the control LSA $\hat{y}_{\theta_\text{GD}}(x_{\text{test}})$, which under our token construction corresponds to a linear model trained by one step of gradient descent starting from $W_0=0$. For this control model, we determine the optimal learning rate $\eta$ by minimizing $\mathcal{L}(\eta)$ over a training set of $10
^4$ tasks through line search, with $\mathcal{L}(\eta)$ defined analogously to \eqref{eq:trans_loss}. 

More concretely, to compare trained and constructed LSA layers, we sample $T_\text{val}=10^4$ validation tasks and record the following quantities, averaged over validation tasks: (1) the difference in predictions measured with the L2 norm, $\|\hat{y}_\theta(x_{\tau,\text{test}}) - \hat{y}_{\theta_\text{GD}}(x_{\tau,\text{test}})\|$, (2) the cosine similarity between the sensitivities $\frac{\partial \hat{y}_{\theta_{\text{GD}}}(x_{\tau,\text{test}}) }{\partial x_{\text{test}}} $ and $ \frac{\partial \hat{y}_{\theta}(x_{\tau,\text{test}})}{\partial x_{\text{test}}}$ as well as (3) their difference $\|\frac{\partial \hat{y}_{\theta_{\text{GD}}}(x_{\tau,\text{test}}) }{\partial x_{\text{test}}} - \frac{\partial \hat{y}_{\theta}(x_{\tau,\text{test}})}{\partial x_{\text{test}}}\|$ again according to the L2 norm, which in both cases yields the explicit models computed by the algorithm. We show the results of these comparisons in Figure~\ref{fig:training_trans_lin}. We find an excellent agreement between the two models over a wide range of hyperparameters. We note that as we do not have direct access to the initialization of $W$ in the attention-based learners (it is hidden in $\theta$), we cannot expect the models to agree exactly.

Although the above metrics are important to show similarities between the resulting learned models (in-context vs.~gradient-based), the underlying algorithms could still be different. We therefore carry out an extended set of analyses:
\begin{enumerate}
 \item \textbf{Interpolation.}  We take inspiration on recent work \citep{permute2, permute1} that showed approximate equivalence of models found by SGD after permuting weights within the trained neural networks. Since our models are deep linear networks with respect to $x_{\text{test}}$ we only correct for scaling mismatches between the two models -- in this case the construction that implements GD and the trained weights. As shown in Figure \ref{fig:training_trans_lin}, we observe (and can actually inspect by eye, see Appendix Figure \ref{fig:weight_vis}) that a simple scaling correction on the trained weights is enough to recover the weight construction implementing GD. This leads to an identical loss of GD, the trained Transformer and the linearly interpolated weights $\theta_{\text{I}} = (\theta + \theta_{\text{GD}})/2$. See details in Appendix \ref{app:mode_connect} on how our weight correction and interpolation is obtained.

\item \textbf{Out-of-distribution validation tasks.} To test if our in-context learner has found a generalizable update rule, we investigate how GD, the trained LSA layer and its interpolation behave when providing in-context data in regimes different to the ones used during training. We therefore visualize the loss increase when (1) sampling the input data from $U(-\alpha, \alpha)^{N_x}$ or (2) scaling the teacher weights by $\alpha$ as $\alpha W$ when sampling validation tasks. For both cases, we set $\alpha=1$ during training. We again observe that when training a single linear self-attention Transformer, for both interventions, the Transformer performs equally to gradient descent outside of this training setups, see Figure \ref{fig:training_trans_lin} as well Appendix Figure \ref{fig:ood_big}. Note that the loss obtained through gradient descent also starts degrading quickly outside the training regime. Since we tune the learning rate for the input range $[-1, 1]$ and one gradient step, tasks with larger input range will have higher curvature and the optimal learning rate for smaller ranges will lead to divergence and a drastic increase in loss also for GD.

\item \textbf{Repeating the LSA update.}  Since we claim that a single trained LSA layer implements a GD-like learning rule, we further test its behavior when applying it repeatedly, not only once as in training.
After we correct the learning rate of both algorithms, i.e. for GD and the trained Transformer with a dampening parameter $\lambda = 0.75$ (details in Appendix \ref{app:dampening}), we see an identical loss decrease of both GD and the Transformer, see Figure \ref{fig:illus}. 
\end{enumerate}

To conclude, we present evidence that optimizing a single LSA layer to solve linear regression tasks finds weights that (approximately) coincide with the LSA-layer weight construction of Proposition \ref{prop:self_att_gd}, hence  implementing a step of gradient descent, leading to the same learning capabilities on in- and out-of-distribution tasks. We comment on the random seed dependent phase transition of the loss during training in Appendix \ref{app:phase_transitions}.

\subsection*{Multiple steps of gradient descent vs.~multiple layers of self-attention}

\begin{figure*}
\textbf{(a) Comparing two steps of gradient descent with trained \textit{recurrent} two-layer Transformers.}
\begin{center}
\begin{minipage}{.24\textwidth}
  \centering
  \begin{center}
    \includegraphics[width=1.\textwidth]{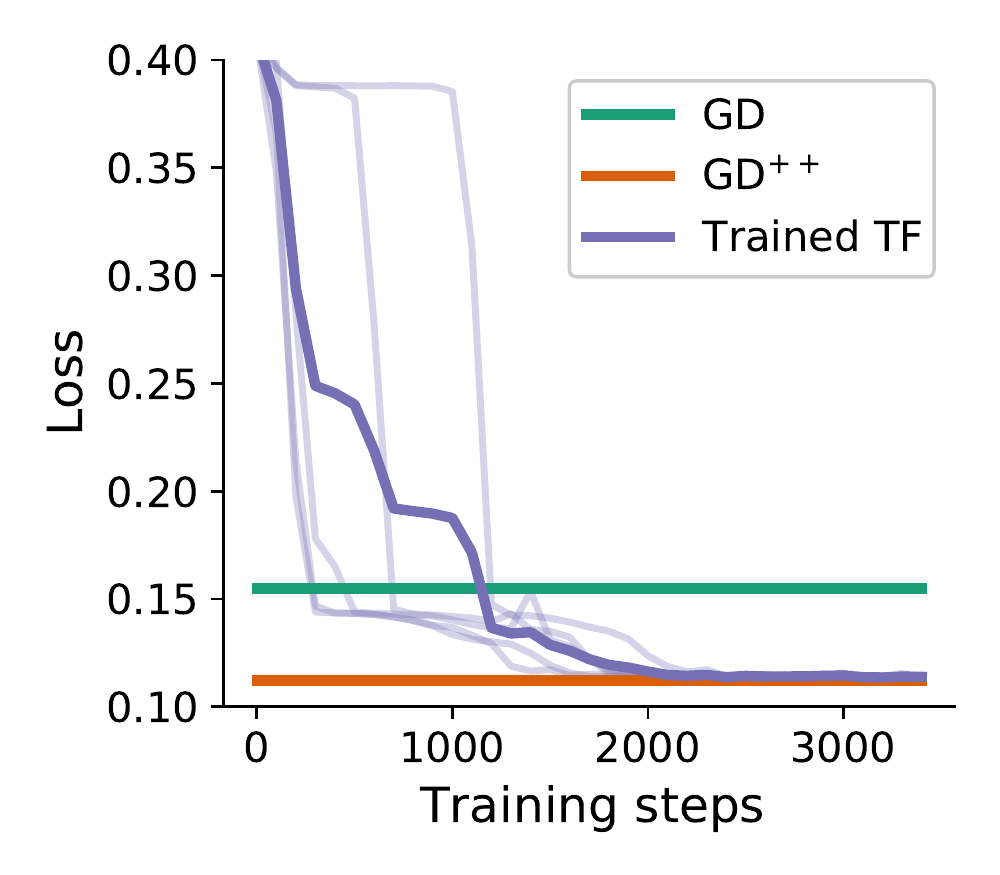}
  \end{center}
  \vspace{-10pt}
\end{minipage}
\begin{minipage}{.24\textwidth}
  \centering
  \begin{center}
    \includegraphics[width=1.\textwidth]{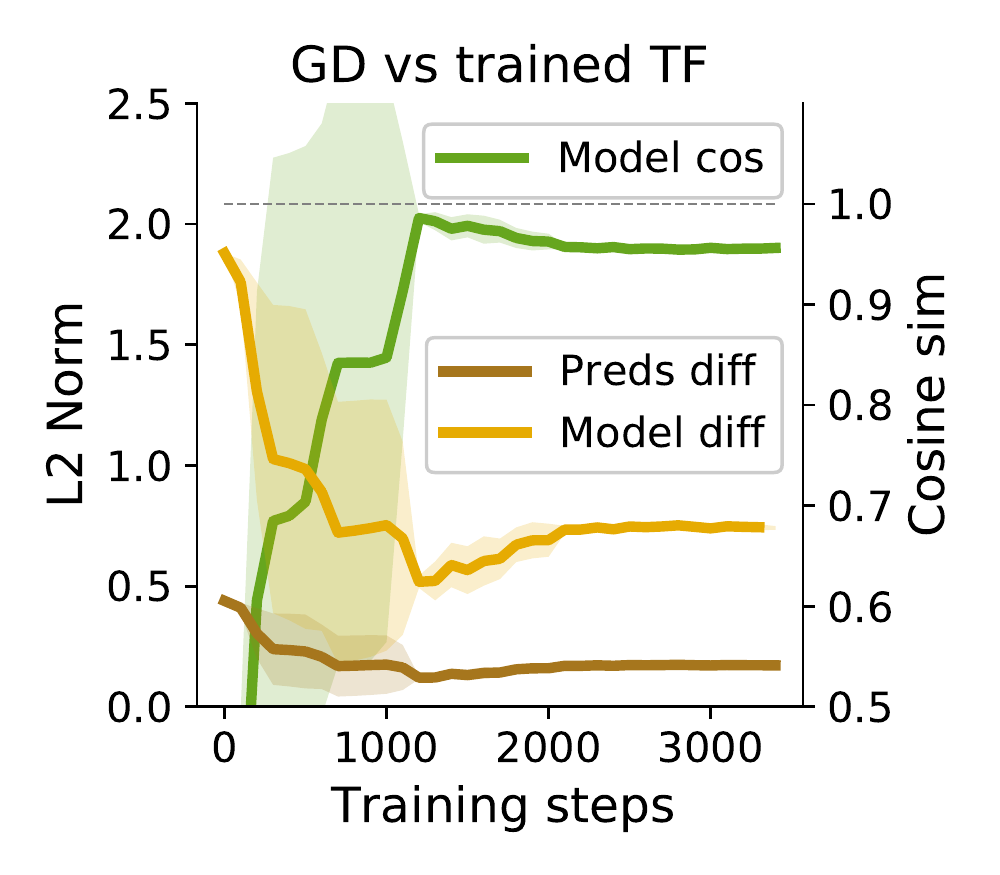}
  \end{center}
  \vspace{-10pt}
\end{minipage}
\begin{minipage}{.24\textwidth}
  \centering
  \begin{center}
    \includegraphics[width=1.\textwidth]{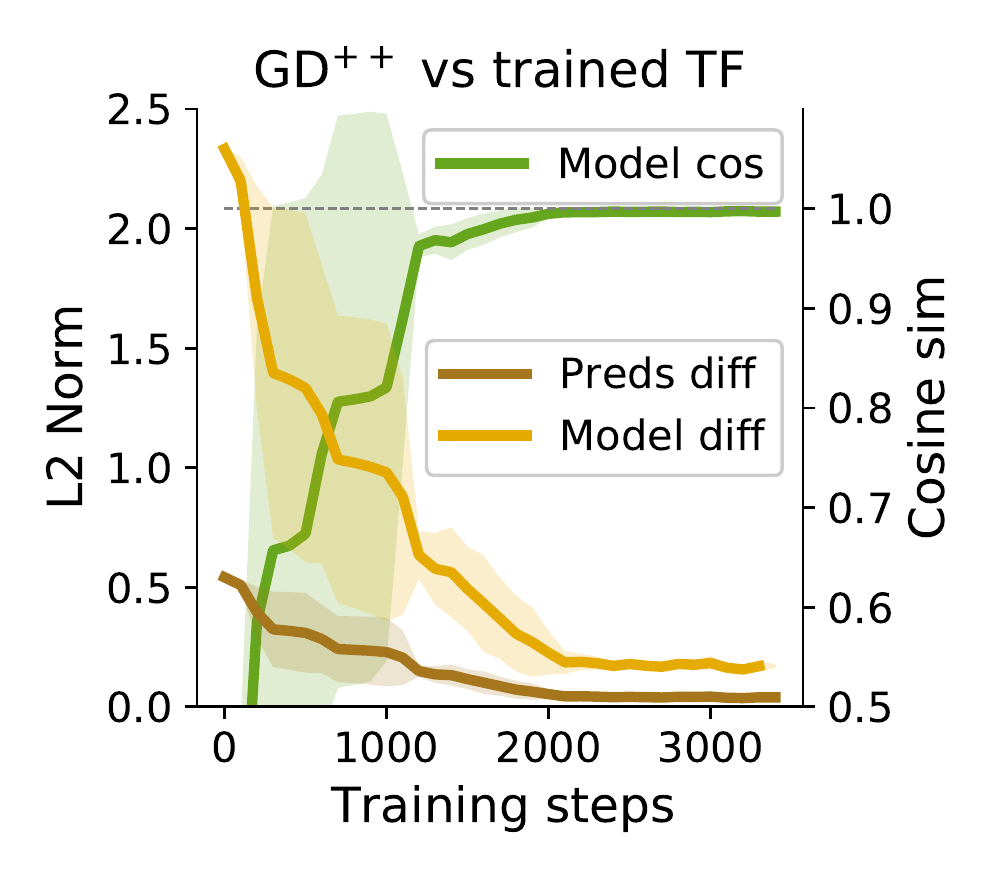}
  \end{center}
  \vspace{-10pt}
\end{minipage}
\begin{minipage}{.24\textwidth}
  \centering
  \begin{center}
    \includegraphics[width=1.\textwidth]{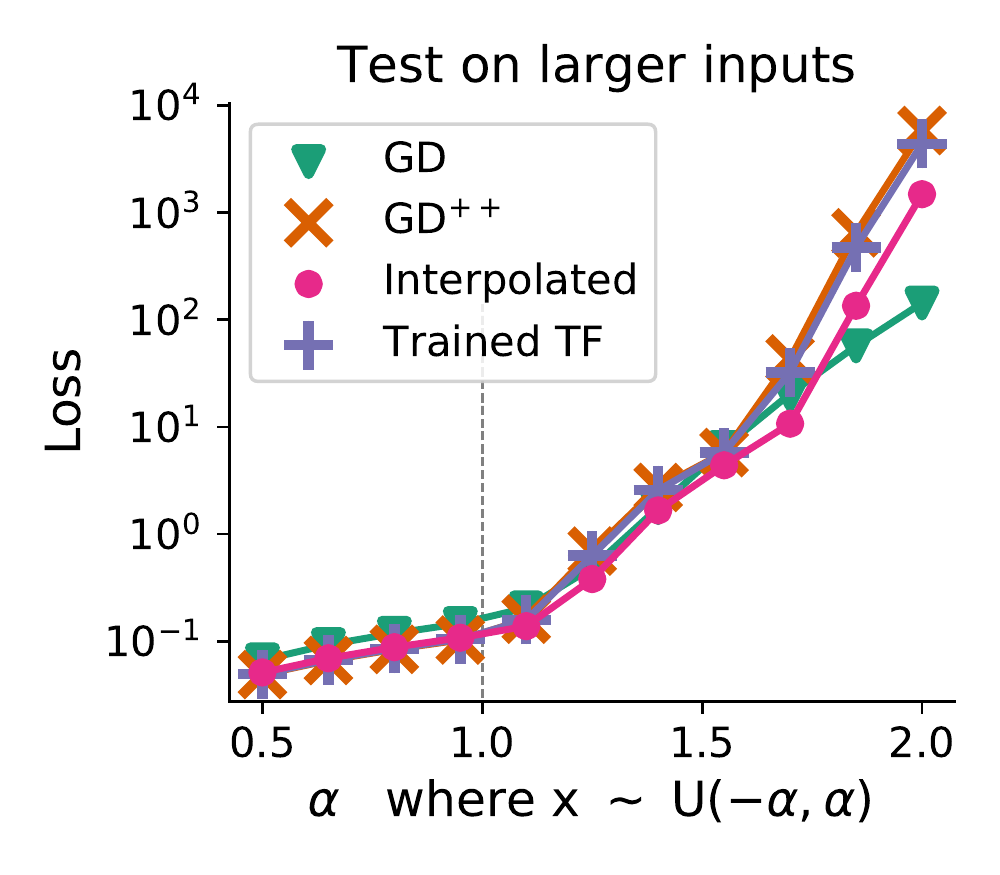}
  \end{center}
  \vspace{-10pt}
\end{minipage}

\end{center}
\textbf{(b) Comparing five steps of gradient descent with trained five-layer Transformers.}

\begin{center}
\begin{minipage}{.24\textwidth}
  \centering
  \begin{center}
    \includegraphics[width=1.\textwidth]{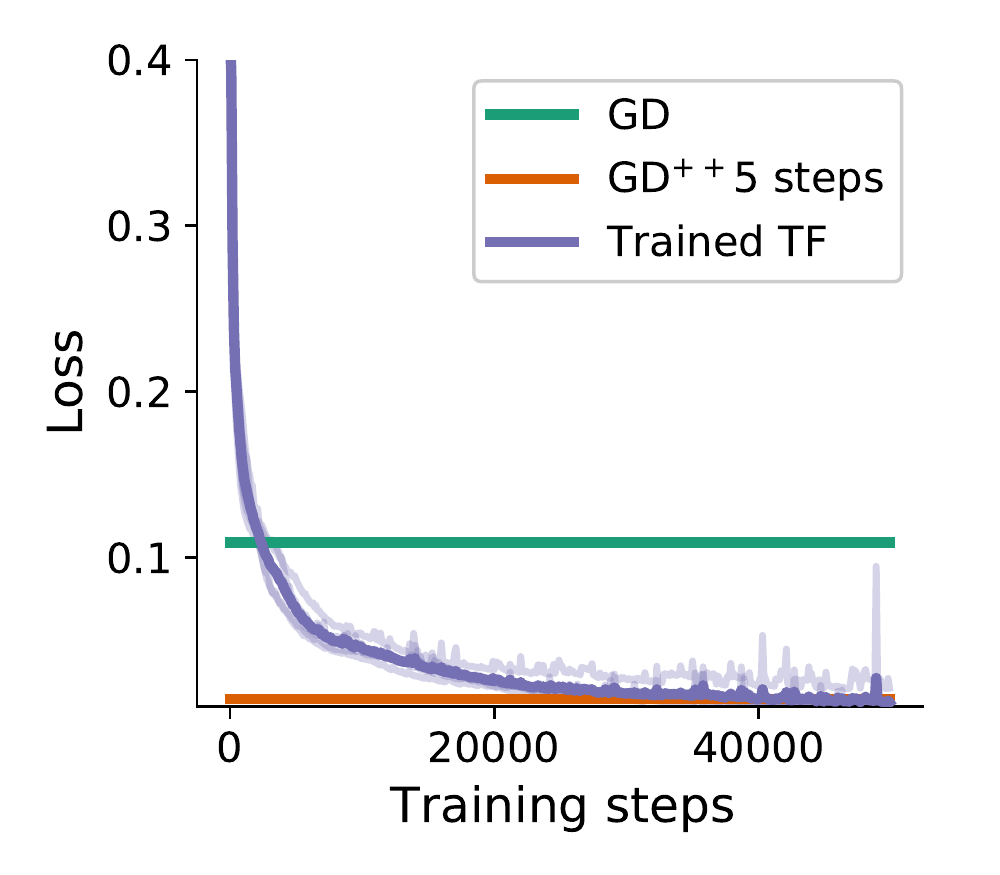}
  \end{center}
  \vspace{-10pt}
\end{minipage}
\begin{minipage}{.24\textwidth}
  \centering
  \begin{center}
    \includegraphics[width=1.\textwidth]{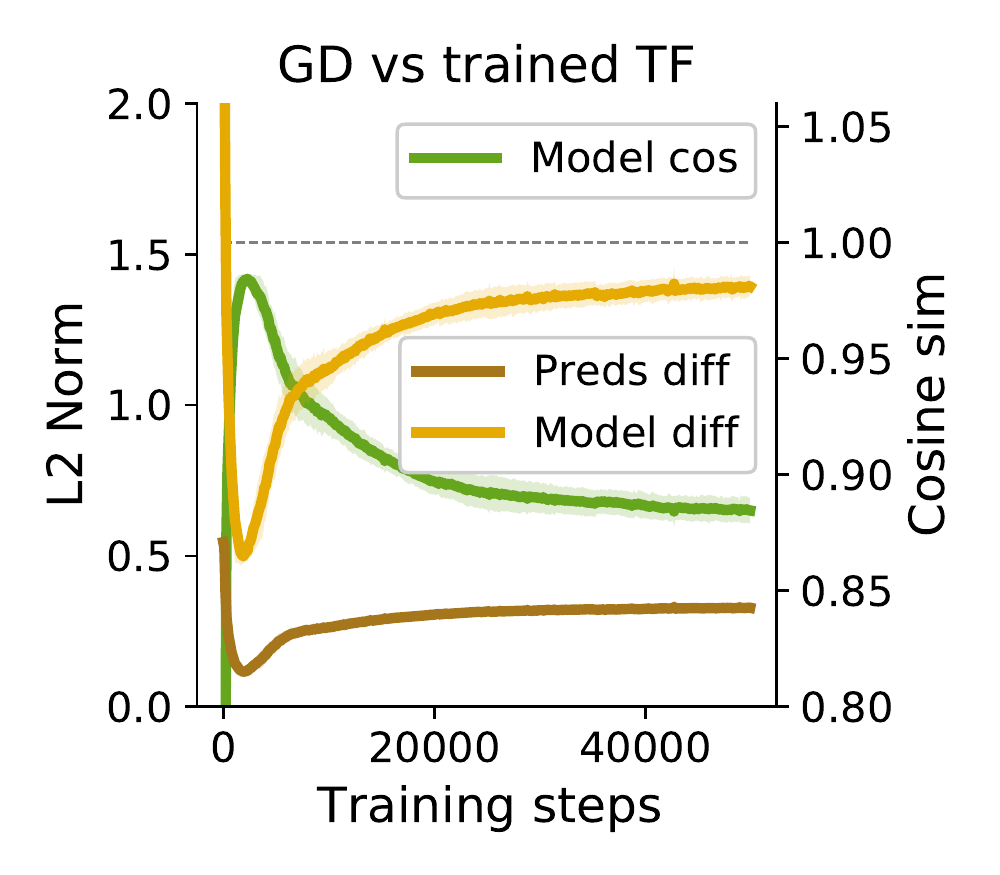}
  \end{center}
  \vspace{-10pt}
\end{minipage}
\begin{minipage}{.24\textwidth}
  \centering
  \begin{center}
    \includegraphics[width=1.\textwidth]{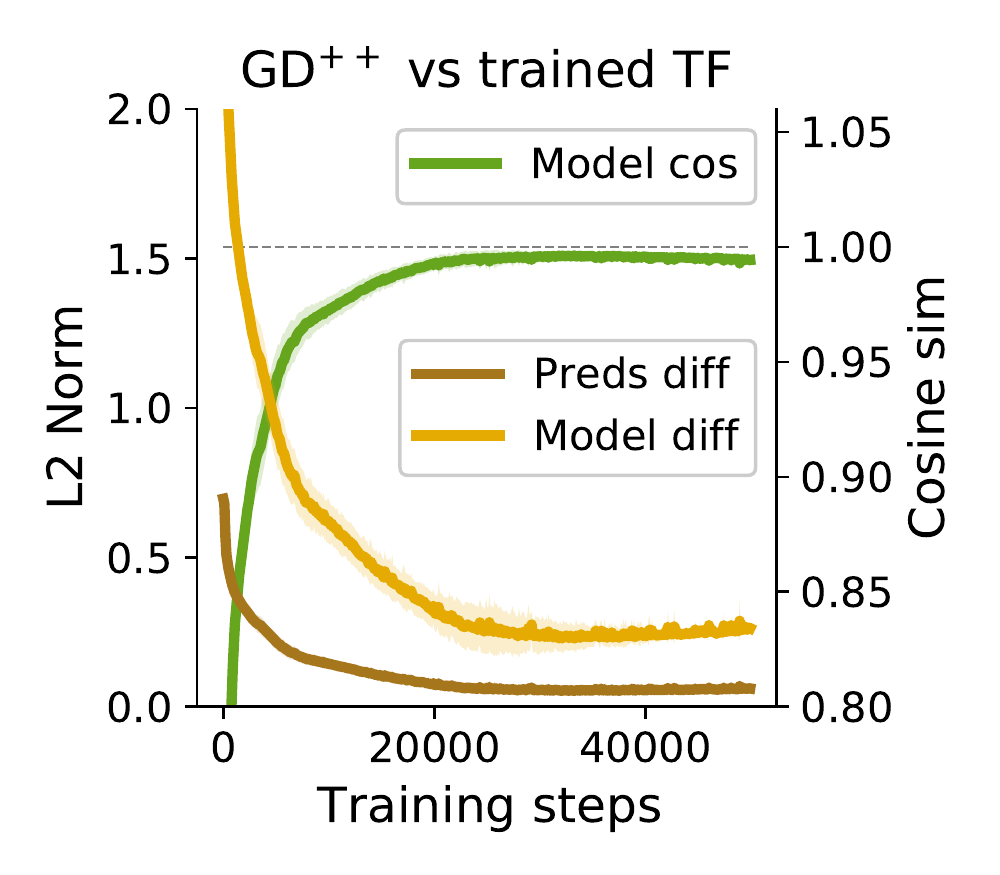}
  \end{center}
  \vspace{-10pt}
\end{minipage}
\begin{minipage}{.24\textwidth}
  \centering
  \begin{center}
    \includegraphics[width=1.\textwidth]{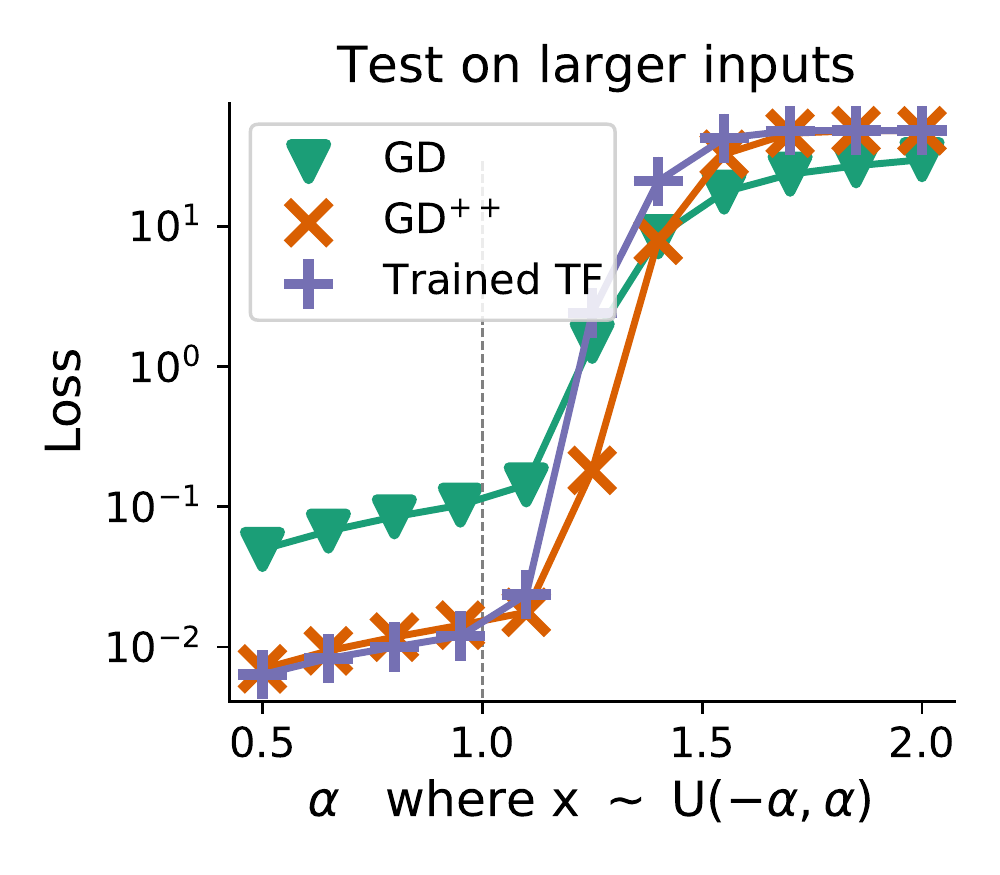}
  \end{center}
\end{minipage}
\end{center}
\vspace{-10pt}
  \caption{\textit{Far left column:}  The trained TF performance surpasses standard  $\text{GD}$ but matches $\text{GD}^{++}$, our GD variant with simple iterative data transformation. On both cases, we tuned the gradient descent learning rates as well as the scalar $\gamma$ which governs the data transformation $H(X)$.
  \textit{Center left \& center right columns}: We measure the alignment between the GD as well as the $\text{GD}^{++}$ models and the trained TF. In both cases the TF aligns well with GD in the beginning of training but aligns much better with $\text{GD}^{++}$ after training. \textit{Far right column:} TF performance (in log-scale) mimics the one of $\text{GD}^{++}$ well when testing on OOD tasks ($\alpha \neq 1$).}
  \label{fig:multi_layer}
   \vspace{-10pt}
\end{figure*}

We now turn to deep linear self-attention-only Transformers. The construction we put forth in Proposition~\ref{prop:self_att_gd}, can be immediately stacked up over $K$ layers; in this case, the final prediction can be read out from the last layer as before by negating the $y$-entry of the last test token: $-y_{N+1} + \sum_{k=1}^K \Delta y_{k,N+1}= - (y_{N+1} - \sum_{k=1}^K \Delta y_{k,N+1}) = y_{N+1} + \sum_{k=1}^K \Delta W_k x_{N+1}$, where $y_{k,N+1}$ are the test token values at layer $k$, and $\Delta y_{k,N+1}$ the change in the $y$-entry of the test token after applying the $k$-th step of self-attention, and $\Delta W_k$ the $k$-th implicit change in the underlying linear model parameters $W$. When optimizing such Transformers with $K$ layers, we observe that these models generally outperform $K$ steps of plain gradient descent, see Figure \ref{fig:multi_layer}. Their behavior is however well described by a variant of gradient descent, for which we tune a single parameter $\gamma$ defined through  the transformation function $H(X)$ which transforms the input data according to $x_j \leftarrow H(X) x_j$, with $H(X) = (I - \gamma XX^T)$. We term this gradient descent variant GD$^{++}$ which we explain and analyze in Appendix \ref{app:curvature_details}.

To analyze the effect of adding more layers to the architecture, we first turn to the arguably simplest extension of a single SA layer and analyze a \textit{recurrent} or \textit{looped} 2-layer LSA model. Here, we simply repeatably apply the same layer (with the same weights) multiple times i.e. drawing the analogy to learning an iterative algorithm that applies the same logic multiple times. 

Somewhat surprisingly, we find that the trained model surpasses plain gradient descent, which also results in decreasing alignment between the two models (see center left column), and the recurrent Transformer realigns perfectly with GD$^{++}$ while matching its performance on in- and out-of distribution tasks. Again, we can interpolate between the Transformer weights found by optimization and the LSA-weight construction with learned $\eta, \gamma$, see Figure \ref{fig:multi_layer} \& \ref{fig:ood_big}.

We next consider deeper, non-recurrent 5-layer LSA-only Transformers, with different parameters per layer (i.e. no weight tying). We see that a different GD learning rate as well as $\gamma$ per step (layer) need to be tuned to match the Transformer performance. This slight modification leads again to almost perfect alignment between the trained TF and GD$^{++}$ with in this case 10 additional parameters and loss close to 0, see Figure \ref{fig:multi_layer}.
Nevertheless, we see that the naive correction necessary for model interpolation used in the aforementioned experiments is not enough to interpolate without a loss increase. We leave a search for better weight corrections to future work. We further study Transformers with different depths for recurrent as well as non-recurrent architectures with multiple heads and equipped with MLPs, and find qualitatively equivalent results, see Appendix Figure~\ref{fig:ten_layer} and Figure~\ref{fig:twelve_layer}. Additionally, in Appendix \ref{app:softmax}, we provide results obtained when using softmax SA layers as well as LayerNorm, thus essentially retrieving the standard Transformer architecture. We again observe and are able to explain (after slight architectural modifications) good learning performance and as well as alignment with the construction of Proposition \ref{prop:self_att_gd}, though worse than when using linear self-attention. These findings suggest that the in-context learning abilities of the standard Transformer with these common architecture choices can be explained by the gradient-based learning hypothesis explored here. Our findings also question the ubiquitous use of softmax attention, and suggest further investigation is warranted into the performance of linear vs.~softmax SA layers in real-world learning tasks, as initiated by \citet{linear_transformers_fast_weight}.

\subsection*{Transformers solve nonlinear regression tasks by gradient descent on deep data representations}

It is unreasonable to assume that the astonishing in-context learning flexibility observed in large Transformers is explained by gradient descent on linear models. We now show that this limitation can be resolved by incorporating one additional element of fully-fledged Transformers: preceding self-attention layers by MLPs enables learning linear models by gradient descent on deep representations which motivates our illustration in Figure \ref{fig:illus}. Empirically, we demonstrate this by solving non-linear sine-wave regression tasks, see Figure \ref{fig:non_linear}. Experimental details can be found in Appendix \ref{app:non_linear}.
We state 
\begin{prop}
\label{prop:kernel}
Given a Transformer block i.e.~a MLP $m(e)$ which transforms the tokens $e_j=(x_j, y_j)$ followed by an attention layer, we can construct weights that lead to gradient descent dynamics descending $\frac{1}{2N}\sum_{i=1}^N||W m(x_i)  -y_i||^2$. Iteratively applying Transformer blocks therefore can solve kernelized least-squares regression problems with kernel function $k(x,y)=m(x)^\top m(y)$ induced by the MLP $m(\cdot)$.
\end{prop}
A detailed discussion on this form of kernel regression as well as kernel smoothing w/wo softmax nonlinearity through gradient descent on the data can be found in Appendix \ref{app:ker_reg}. The way MLPs transform data in Transformers diverges from the standard meta-learning approach, where a task-shared \textit{input} embedding network is optimized by backpropagation-through-training to improve the learning performance of a task-specific readout \citep[e.g.,][]{raghu_rapid_2020,meta_opt_net,bertinetto_meta-learning_2019}. On the other hand, given our token construction in Proposition \ref{prop:self_att_gd}, MLPs in Transformers intriguingly process both inputs \emph{and} targets. The output of this transformation is then processed by a single linear self-attention layer, which, according to our theory, is capable of implementing gradient descent learning. We compare the performance of this Transformer model, where all weights are learned, to a control Transformer where the final LSA weights are set to the construction $\theta_\text{GD}$ which is therefore identical to training an MLP by backpropagation through a GD updated output layer.

Intriguingly, both obtained functions show again surprising similarity on (1) the initial (meta-learned) prediction, read out after the MLP, and (2) the final prediction, after altering the output of the MLP through GD or the self-attention layer. This is again reflected in our alignment measures that now, since the obtained models are nonlinear w.r.t.~$x_{\text{test}}$, only represent the two first parts of the Taylor approximation of the obtained functions. 
Our results serve as a first demonstration of how MLPs and self-attention layers can interplay to support nonlinear in-context learning, allowing to fine-tune deep data representations by gradient descent. Investigating the interplay between MLPs and SA-layer in deep TFs is left for future work.

\begin{figure*}
\vspace{-7pt}
\begin{center}
\begin{minipage}{.36\textwidth}
  \centering
  \begin{center}
    \includegraphics[width=1.\textwidth]{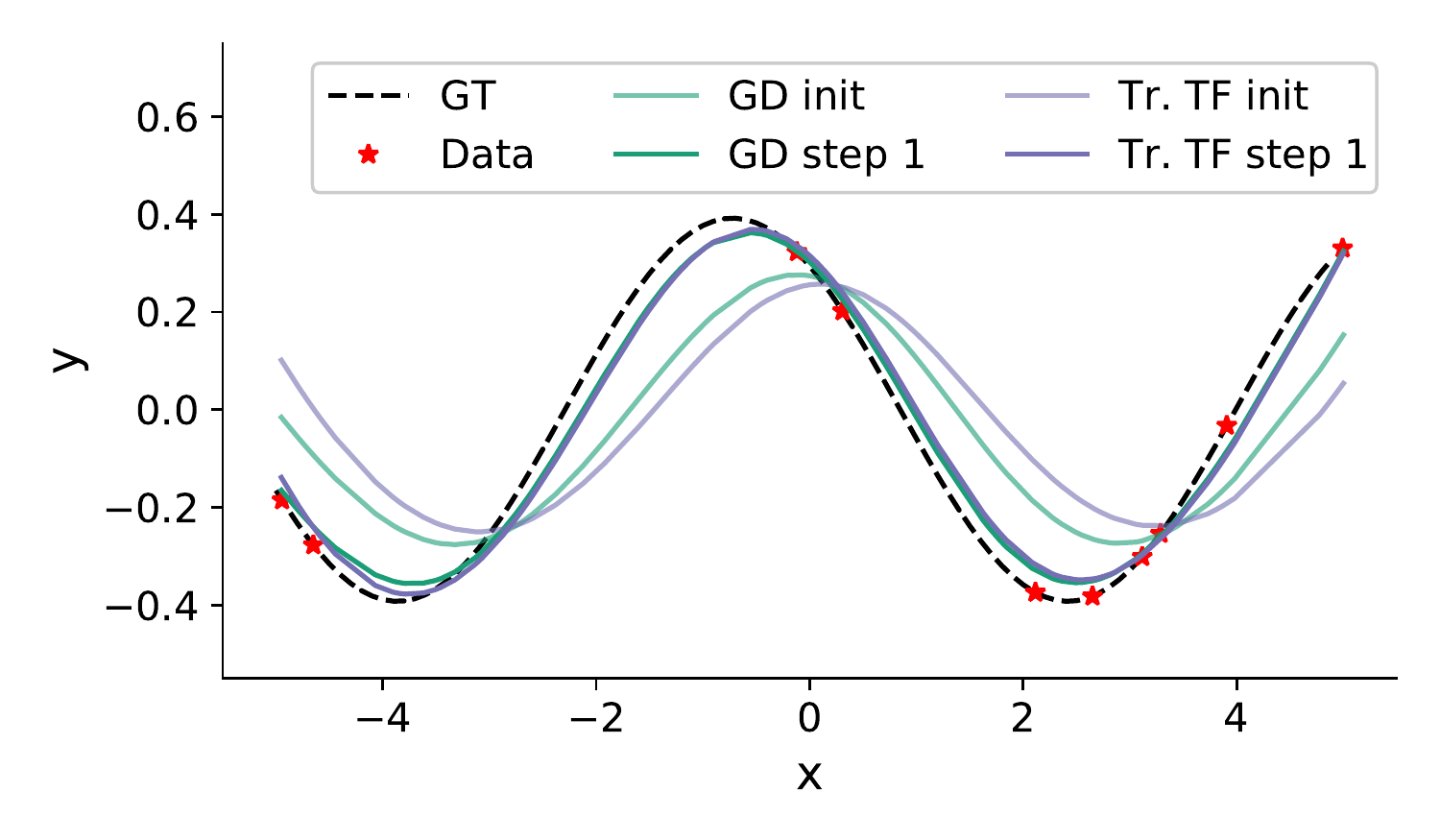}
  \end{center}
\end{minipage}
\hspace{20pt}
\begin{minipage}{.24\textwidth}
  \centering
  \begin{center}
    \includegraphics[width=1.\textwidth]{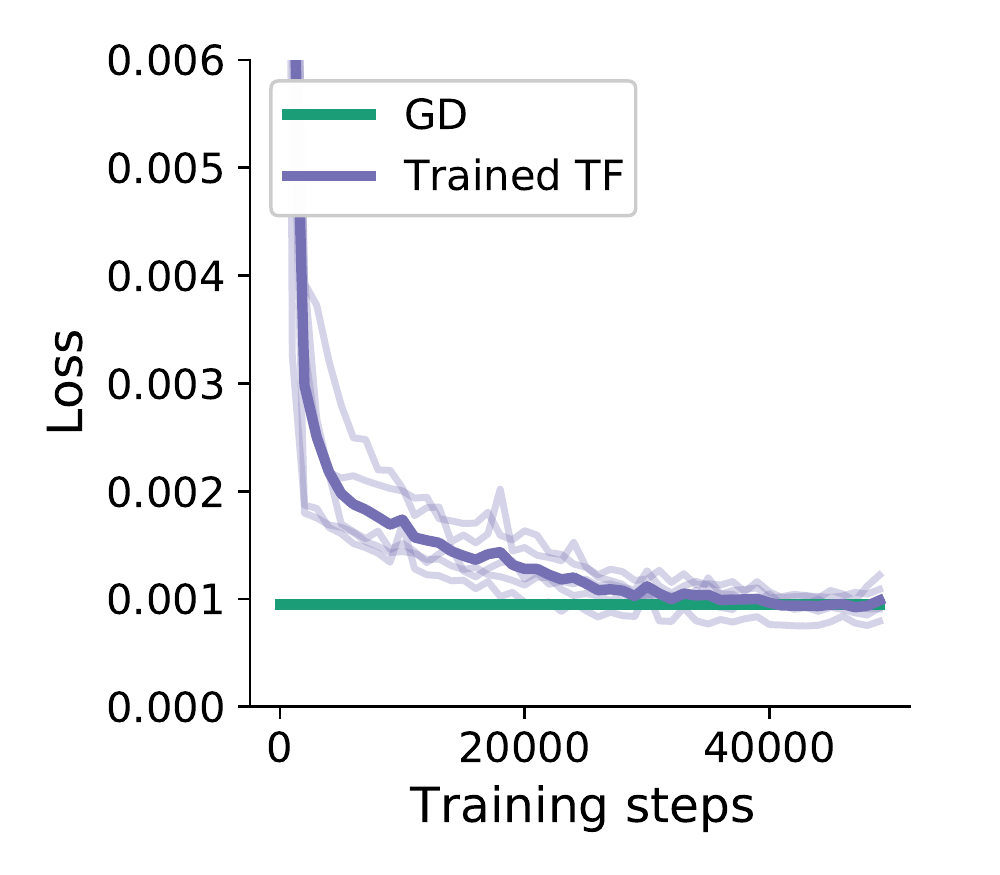}
  \end{center}
\end{minipage}
  \hspace{10pt}
\begin{minipage}{.24\textwidth}
  \centering
  \begin{center}
    \includegraphics[width=1.\textwidth]{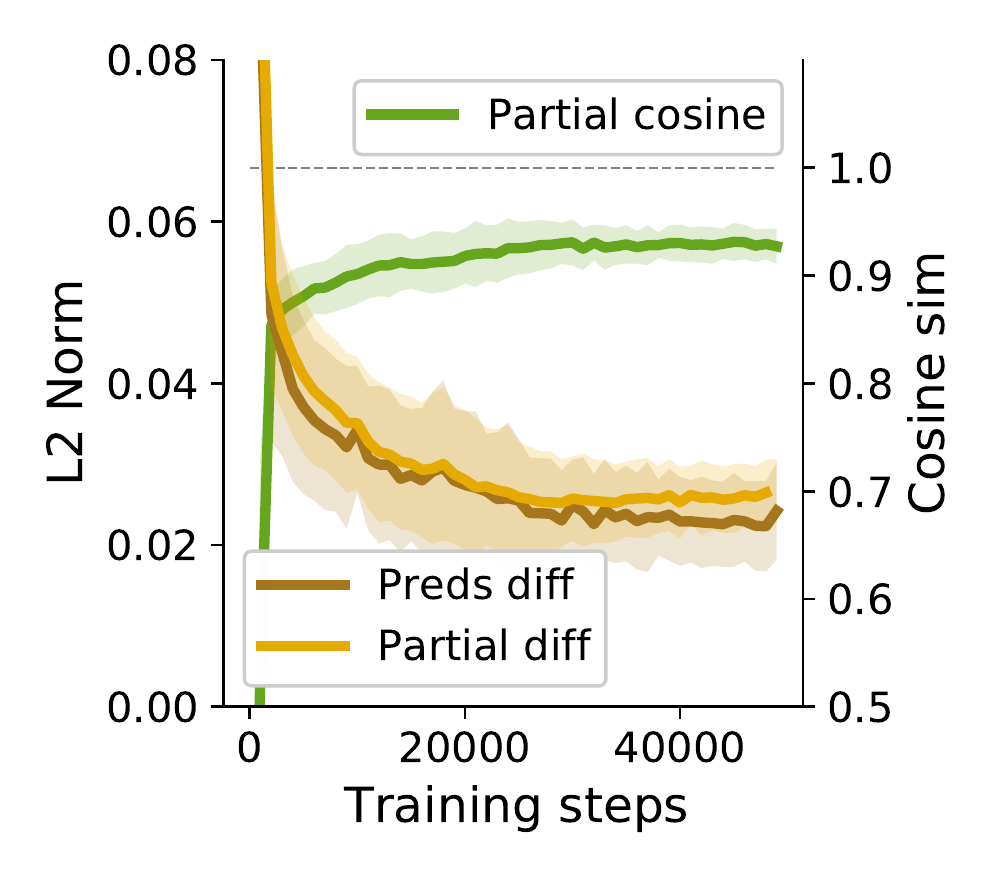}
  \end{center}
\end{minipage}
\vspace{-8pt}
  \caption{\textbf{Sine wave regression: comparing trained Transformers with meta-learned MLPs for which we adjust the output layer with one step of gradient descent.} \textit{Left:} Plots of the learned initial functions as well as the adjusted functions through either a layer of self-attention or a step of GD. We observe similar initial functions as well as solutions for the trained TF compared fine-tuning a meta-learned MLP. \textit{Center}: The performance of the trained Transformer is matched by meta-learned MLPs. \textit{Left}: We observe strong alignment when comparing the prediction as well as the partial derivatives of the the meta-learned MLP and the trained Transformer.}
  \label{fig:non_linear}
 \end{center}
 \vspace{-15pt}
\end{figure*}
\section{Do self-attention layers build regression tasks?}

\label{sect:softmax-builds-tokens}
\begin{figure}[H]
\vspace{-7pt}
\centering
\hspace{-10pt}
\begin{minipage}{.26\textwidth}
  \centering
  \begin{center}
    \includegraphics[width=1.\textwidth]{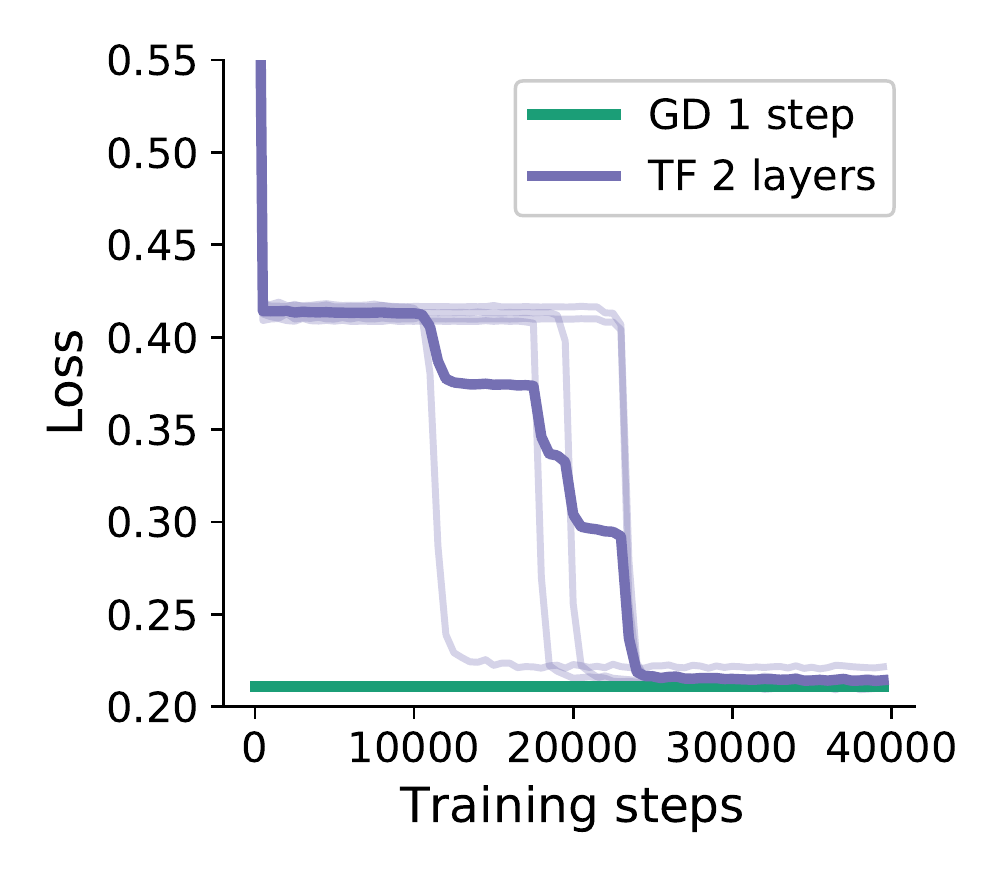}
  \end{center}
\end{minipage}
\hspace{-15pt}
\begin{minipage}{.26\textwidth}
  \centering
  \begin{center}
    \includegraphics[width=1.\textwidth]{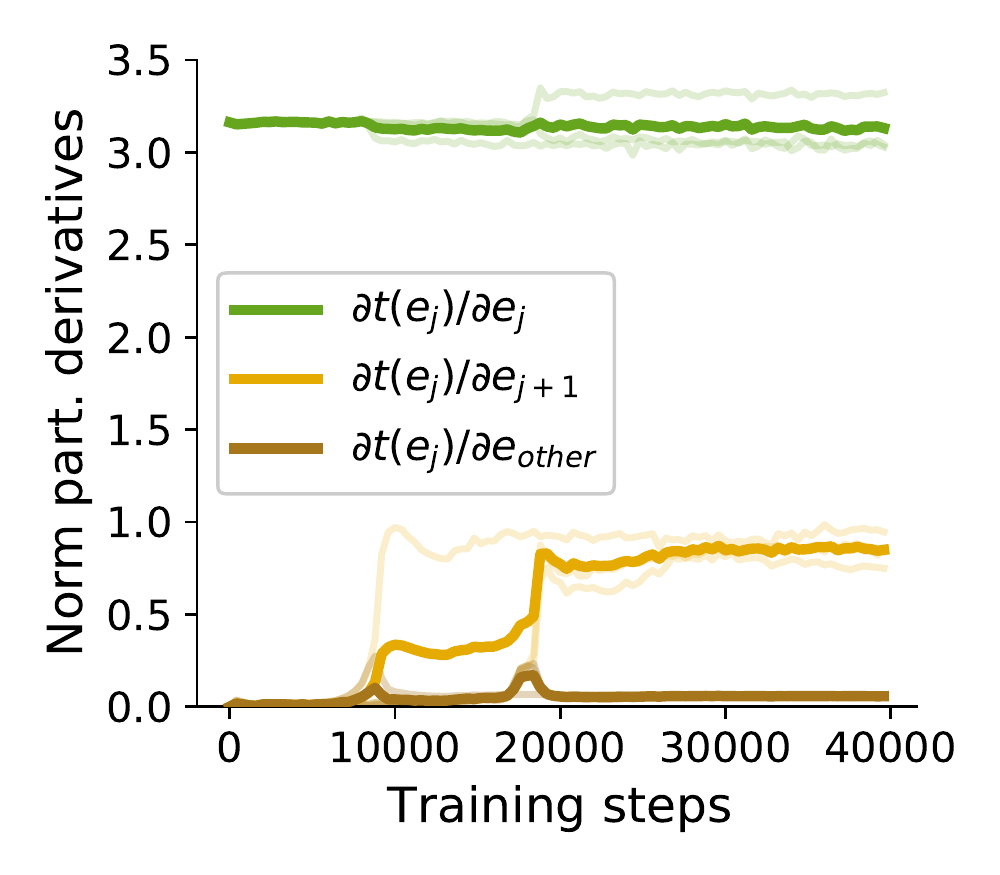}
  \end{center}
\end{minipage}
\hspace{-15pt}
\vspace{-8pt}
  \caption{\textbf{Training a two layer SA-only Transformer using the standard token construction.} \textit{Left:} The loss of trained TFs matches one step of GD, not two, and takes an order of magnitude longer to train. \textit{Right}: Norm of the partial derivatives of the output of the first self-attention layer w.r.t.~input tokens. Before the Transformer performance jumps to the one of GD, the first layer becomes highly sensitive to the next token.}
  \label{fig:copy}
  \vspace{-10pt}
\end{figure}

The construction provided in Proposition \ref{prop:self_att_gd} and the previous experimental section relied on a token structure where both input and output data are concatenated into a single token. This design is different from the way tokens are typically built in most of the related work dealing with simple few-shot learning problems as well as in e.g. language modeling. We therefore ask: Can we overcome the assumption required in Proposition \ref{prop:self_att_gd} and allow a Transformer to build the required token construction on its own? This motivates 
\begin{prop}
\label{prop:token_copy}
Given a 1-head linear or softmax attention layer and the token construction $e_{2j} = (x_j), e_{2j+1} = (0, y_j)$ with a zero vector $0$ of dim $N_x -N_y$ and concatenated positional encodings, one can construct key, query and value matrix $W_K, W_Q, W_V$ as well as the projection matrix $P$ such that all tokens $e_j$ are transformed into tokens equivalent to the ones required in Proposition \ref{prop:self_att_gd}. 
\end{prop}
The construction and its discussion can be found in Appendix \ref{app:prop2}. 
To provide evidence that copying is performed in trained Transformers, we optimize a two-layer self-attention circuit on in-context data where alternating tokens include input or output data i.e. $e_{2j}=(x_j)$ and $e_{2j+1}=(0,y_j)$. We again measure the loss as well as the mean of the norm of the partial derivative of the first layer's output w.r.t.~the input tokens during training, see Figure \ref{fig:copy}. First, the training speeds are highly variant given different training seeds, also reported in \citet{simple_case_study}. Nevertheless, the Transformer is able to match the performance of a \textit{single} (not two) step gradient descent. Interestingly, before the Transformer performance jumps to the one of GD, token $e_j$ transformed by the first self-attention layer becomes notably dependant on the neighboring token $e_{j+1}$ while staying independent on the others which we denote as $e_\text{other}$ in Figure \ref{fig:copy}.

We interpret this as evidence for a copying mechanism of the Transformer's first layer to merge input and output data into single tokens as required by Proposition~\ref{prop:self_att_gd}. Then, in the second layer the Transformer performs a single step of GD. Notably, we were not able to train the Transformer with linear self-attention layers, but had to incorporate the softmax operation in the first layer. These preliminary findings support the study of \citet{induction_heads} showing that softmax self-attention layers easily learn to copy; we confirm this claim, and further show that such copying allows the Transformer to proceed by emulating gradient-based learning in the second or deeper attention layers.

We conclude that copying through (softmax) attention layers is the second crucial mechanism for in-context learning in Transformers. This operation enables Transformers to merge data from different tokens and then to compute dot products of input and target data downstream, allowing for in-context learning by gradient descent to emerge. 

\section{Discussion}

Transformers show remarkable in-context learning behavior. Mechanisms based on attention, associative memory and copying by induction heads are currently the leading explanations for this remarkable feature of learning within the Transformer forward pass. In this paper, we put forward the hypothesis, similar to \citet{simple_case_study} and \citet{related_work}, that Transformer's in-context learning is driven by gradient descent, in short -- \textit{Transformers learn to learn by gradient descent based on their context}. Viewed through the lens of meta-learning, learning Transformer weights corresponds to the outer-loop which then enables the forward pass to transform tokens by gradient-based optimization. 

To provide evidence for this hypothesis, we build on \citet{linear_transformers_fast_weight} that already provide a linear self-attention layer variant with (fast-)inner loop learning by the error-correcting delta rule \cite{widrow:switching}. We diverge from their setting and focus on (in-context) learning where we specifically construct a dataset by considering neighboring elements in the input sequence as input- and target training pairs, see  assumptions of Proposition 1. This construction could be realized, for example, due to the model learning to implement a copying layer, see section \ref{sect:softmax-builds-tokens} and proposition \ref{prop:token_copy}, and allows us to provide a simple and different construction to \citet{linear_transformers_fast_weight} that solely is built on the standard linear, and approximately softmax, self-attention layer but still implements gradient descent based learning dynamics. We, therefore, are able to explain gradient descent based learning in these standard architectures. Furthermore, we extend this construction based on a single self-attention layer and provide an explanation of how deeper K-layer Transformer models implement principled K-step gradient descent learning, which deviates again from Schlag et al. and allows us to identify that deep Transformers implement GD++, an accelerated version of gradient descent.

We highlight that our construction of gradient descent and GD++ is not suggestive but when training multi-layer self-attention-only Transformers on simple regression tasks, we provide strong evidence that the construction is actually found. This allows us, at least in our restricted problems settings, to explain mechanistically in-context learning in trained Transformers and its close resemblance to GD observed by related work. Further work is needed to incorporate regression problems with noisy data and weight regularization into our hypothesis. We speculate aspects of learning in these settings are meta-learned -- e.g., the weight magnitudes to be encoded in the self-attention weights. Additionally, we did not analyze logistic regression for which one possible weight construction is already presented in \citet{hypertransformer}.

Our refined understanding of in-context learning based on gradient descent motives us to investigate how to improve it. We are excited about several avenues of future research. 
First, to exceed upon a single step of gradient descent in every self-attention layer it could be advantageous to incorporate so called  \textit{declarative} nodes \citep{amos_optnet_2017,bai_deep_2019,gould_deep_2021,zucchet_beyond_2022} into Transformer architectures. This way, we would treat a single self-attention layer as the solution of a fully optimized regression loss leading to possibly more efficient architectures.
Second, our findings are restricted to small Transformers and simple regression problems. We are excited to delve deeper into research trying to understand how further mechanistic understanding of Transformers and in-context learning in larger models is possible and to what extend.
Third, we are excited about targeted modifications to Transformer architectures, or their training protocols, leading to improved gradient descent based learning algorithms or allow for alternative in-context learners to be implemented within Transformer weights, augmenting their functionality, as e.g. in \citet{dai2023why}.
Finally, it would be interesting to analyze in-context learning in HyperTransformers \citep{hypertransformer} that produce weights for target networks and already offer a different perspective on merging Transformers and meta-learning. There, Transformers transform weights instead of data and could potentially allow for gradient computations of weights deep inside the target network lifting the limitation of GD on linear models analyzed here.

\subsection*{Acknowledgments}
João Sacramento and Johannes von Oswald deeply thank Angelika Steger for her support and guidance. The authors also thank Seijin Kobayashi, Marc Kaufmann, Nicolas Zucchet, Yassir Akram, Guillaume Obozinski and Mark Sandler for many valuable insights throughout the project and Dale Schuurmans and Timothy Nguyen for their valuable comments on the manuscript. João Sacramento was supported by an Ambizione grant (PZ00P3\_186027) from the Swiss National Science Foundation and an ETH Research Grant (ETH-23 21-1). 

\newpage
\bibliography{example_paper}
\bibliographystyle{icml2023}

\newpage
\appendix
\onecolumn
\section{Appendix}
\label{app:main}
\subsection{Proposition 1}
\label{app:prop1}
First, we highlight the dependency on the tokens $e_i$ of the linear self-attention operation 

\begin{align}
e_j  \leftarrow e_j + \lsattn_{\theta}(\{e_1,\ldots,e_N\}) = e_j + \sum_{h} P_hV_hK_h^{T}q_{h,j} &= e_j + \sum_{h} P_h \sum_{i}  v_{h,i} \otimes k_{h,i}q_{h,j} \nonumber\\
\label{eq:lin_trans}
&= e_j + \sum_{h} P_h W_{h,V}\sum_{i} e_{h,i} \otimes e_{h,i} W_{h,K}^T W_{h,Q}e_{j}
\end{align}
with $\otimes$ the outer product between two vectors. With this we can now easily draw connections to one step of gradient descent on $L(W) = \frac{1}{2N}\sum_{i=1}^{N} \|Wx_i - y_i\|^2$ with learning rate $\eta$ which yields weight change
\begin{align}
    \Delta W = - \eta \nabla_W L(W) = - \frac{\eta}{N} \sum_{i=1}^{N}(Wx_i - y_i)x_i^T.
\end{align}
We first restate

\setcounter{prop}{0}
\begin{prop}

Given a 1-head linear attention layer and the tokens $e_j = (x_j, y_j)$, for $j=1,\ldots,N$, one can construct key, query and value matrices $W_K, W_Q, W_V$ as well as the projection matrix $P$ such that a Transformer step on every token $e_j$ is identical to the gradient-induced dynamics $ e_j \leftarrow (x_j, y_j) + (0, -\Delta W x_j) = (x_i, y_{i}) + P \,V K^{T}q_{j}$ such that $e_j = (x_j, y_j - \Delta y_j)$. For the test data token $(x_{N+1}, y_{N+1})$ the dynamics are identical.
\end{prop}
We provide the weight matrices in block form: $W_{K} = W_{Q} = \left(\begin{array}{@{}c c@{}}
  I_x
  & 0 \\
  0 &
  0
\end{array}\right)
$ with $I_x$ and $I_y$ the identity matrices of size $N_x$ and $N_y$ respectively. Furthermore, we set $W_{V} = \left(\begin{array}{@{}c c@{}}
  0
  & 0 \\
  W_0 &
  -I_y
\end{array}\right)$ with the weight matrix $W_0 \in \mathbb{R}^{N_y \times N_x}$ of the linear model we wish to train and $P = \frac{\eta}{N}I$ with identity matrix of size $N_x + N_y$. With this simple construction we obtain the following dynamics 
\begin{align}
\left(\begin{array}{@{}c@{}}
  x_j\\
  y_j 
\end{array}\right)
  \leftarrow & \left(\begin{array}{@{}c@{}}
  x_j\\
  y_j 
\end{array}\right)
+ \frac{\eta}{N}I \sum_{i=1}^N
\left(\left(\begin{array}{@{}c c@{}}
  0
  & 0 \\
  W_0 &
  -I_y
\end{array}\right)
\left(\begin{array}{@{}c@{}}
  x_i\\
  y_i 
\end{array}\right)\right)
 \otimes\left(
\left(\begin{array}{@{}c c@{}}
  I_x
  & 0 \\
  0 &
  0
\end{array}\right)\left(\begin{array}{@{}c@{}}
  x_i\\
  y_i 
\end{array}\right)\right)
\left(\begin{array}{@{}c c@{}}
  I_x
  & 0 \\
  0 &
  0
\end{array}\right)
\left(\begin{array}{@{}c@{}}
  x_j\\
  y_j 
\end{array}\right)
 \nonumber\\
 \label{eq:trans_update}
&= \left(\begin{array}{@{}c@{}}
  x_j\\
  y_j 
\end{array}\right)
+ \frac{\eta}{N}I \sum_{i=1}^N
\left(\begin{array}{@{}c@{}}
  0\\
W_0 x_i - y_i 
\end{array}\right)
 \otimes\left(\begin{array}{@{}c@{}}
  x_i\\
   0
\end{array}\right)
\left(\begin{array}{@{}c@{}}
  x_j\\
  0
\end{array}\right) = \left(\begin{array}{@{}c@{}}
  x_j\\
  y_j 
\end{array}\right) + \left(\begin{array}{@{}c@{}}
  0\\
  -\Delta W x_j 
\end{array}\right).
\end{align}

for every token $e_j = (x_j, y_j)$ including the query token $e_{N+1} = e_{\text{test}} = (x_{\text{test}}, -W_0 x_{\text{test}})$ which will give us the desired result.

\subsection{Comparing the out-of-distribution behavior of trained Transformers and GD}

\begin{figure*}
\textbf{(a) Comparing one step of gradient descent with trained one layer Transformers on OOD data.}
\begin{center}
\begin{minipage}{.24\textwidth}
  \centering
  \begin{center}
    \includegraphics[width=1.\textwidth]{Final_figures/linear/linear_one_layer_one_head_0_002/normal.pdf}
  \end{center}
  \vspace{-10pt}
\end{minipage}
\begin{minipage}{.24\textwidth}
  \centering
  \begin{center}
    \includegraphics[width=1.\textwidth]{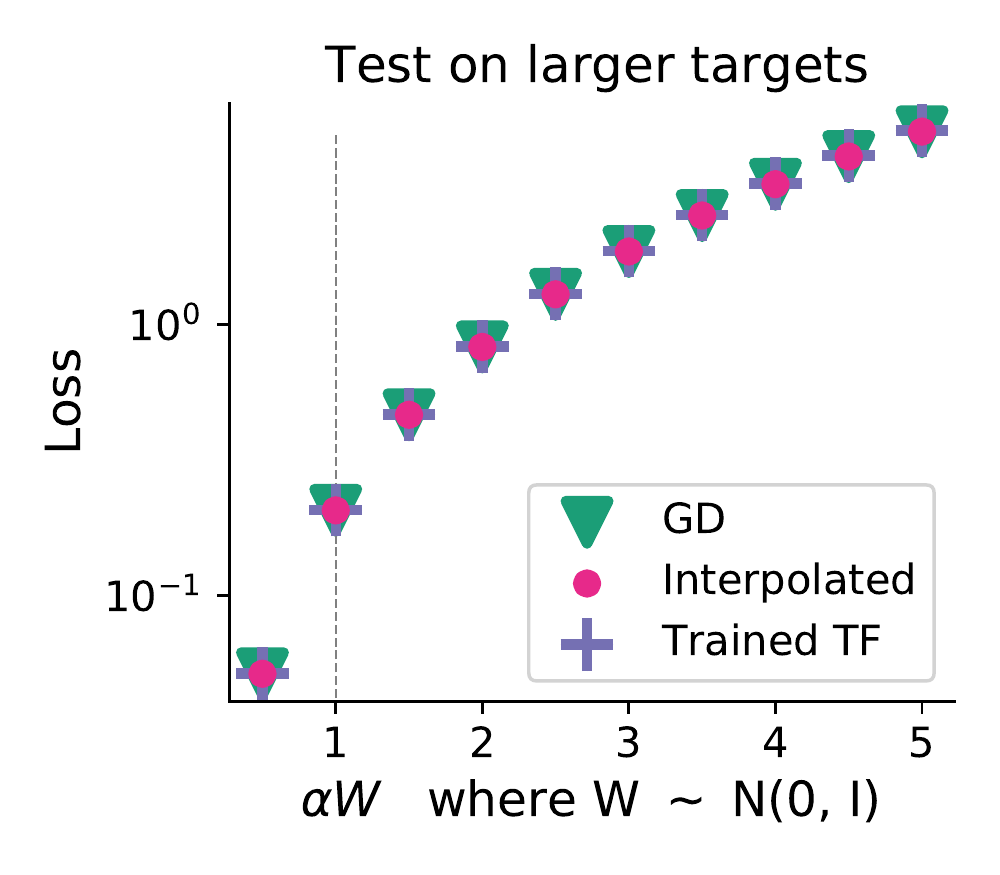}
  \end{center}
  \vspace{-10pt}
\end{minipage}
\begin{minipage}{.24\textwidth}
  \centering
  \begin{center}
    \includegraphics[width=1.\textwidth]{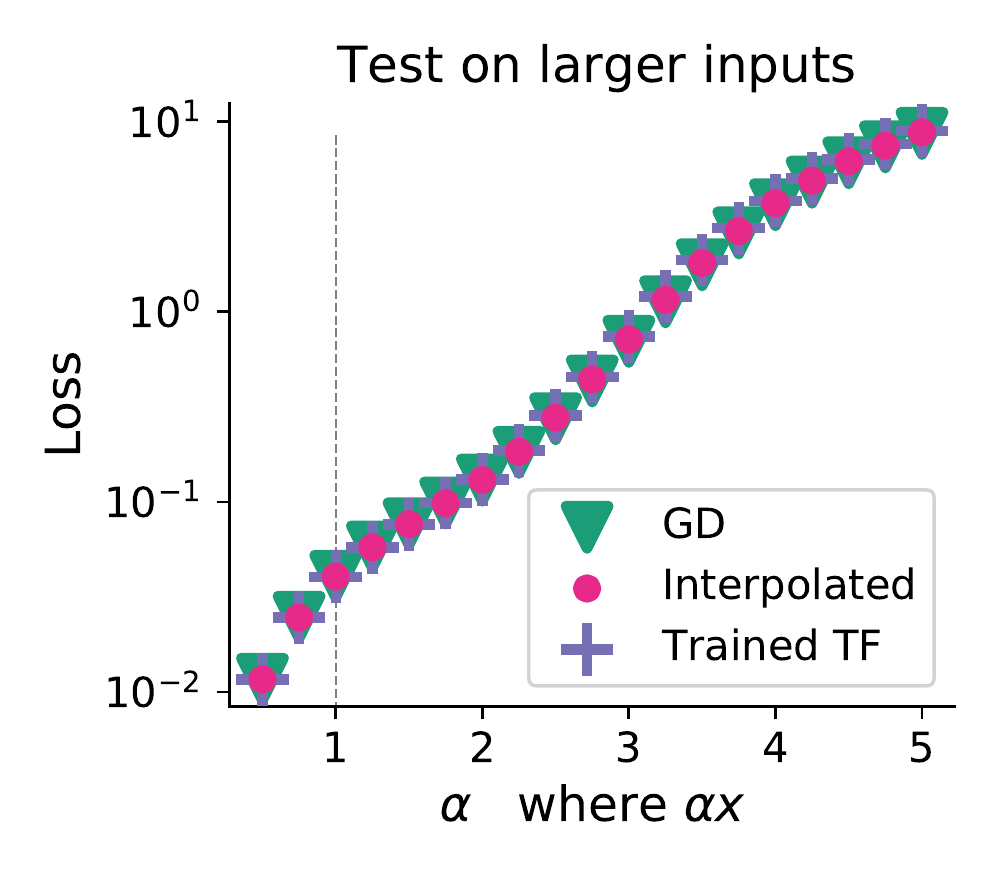}
  \end{center}
  \vspace{-10pt}
\end{minipage}
\begin{minipage}{.24\textwidth}
  \centering
  \begin{center}
    \includegraphics[width=1.\textwidth]{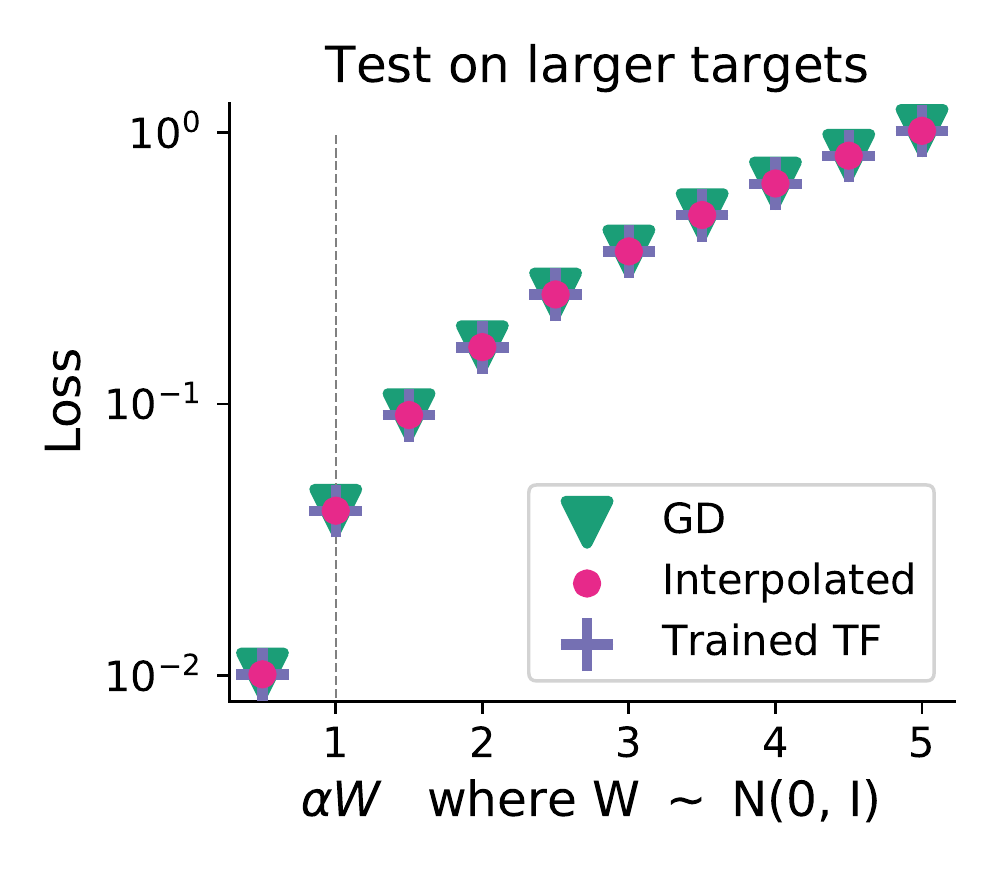}
  \end{center}
  \vspace{-10pt}
\end{minipage}

\end{center}
\textbf{(b) Comparing two steps of gradient descent with trained \textit{recurrent} two layer Transformers on OOD data.}

\begin{center}
\begin{minipage}{.24\textwidth}
  \centering
  \begin{center}
    \includegraphics[width=1.\textwidth]{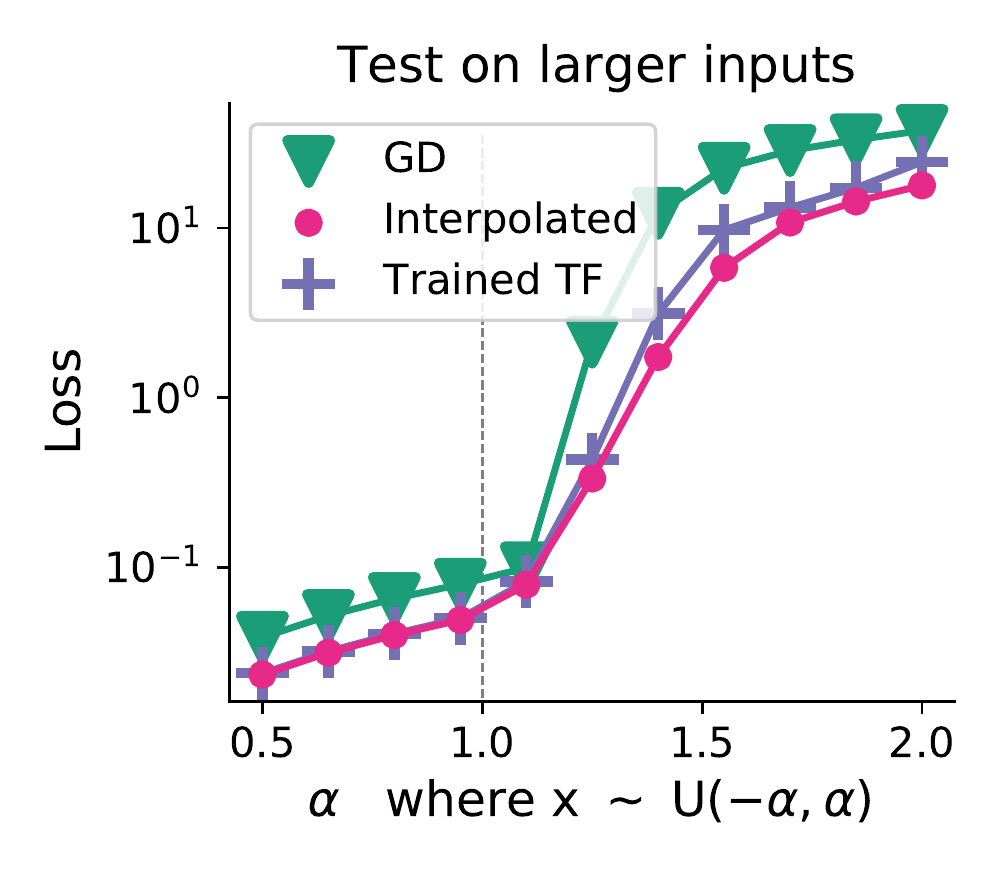}
  \end{center}
  \vspace{-10pt}
\end{minipage}
\begin{minipage}{.24\textwidth}
  \centering
  \begin{center}
    \includegraphics[width=1.\textwidth]{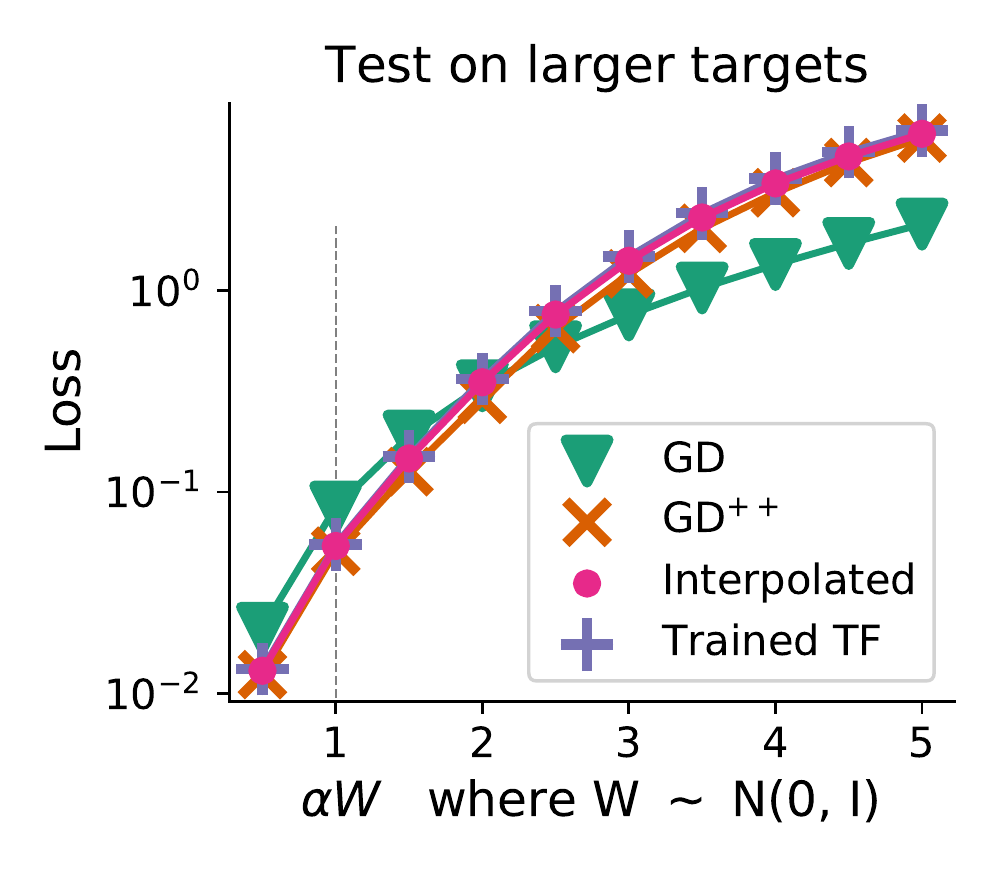}
  \end{center}
  \vspace{-10pt}
\end{minipage}
\begin{minipage}{.24\textwidth}
  \centering
  \begin{center}
    \includegraphics[width=1.\textwidth]{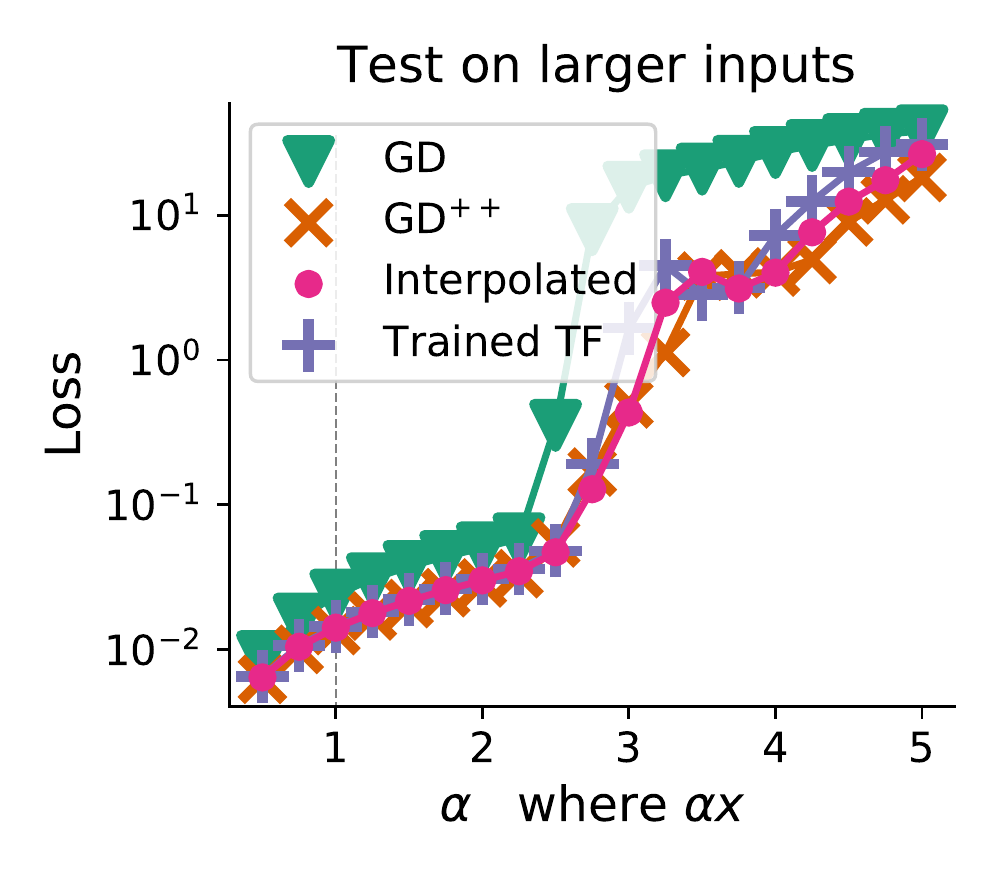}
  \end{center}
  \vspace{-10pt}
\end{minipage}
\begin{minipage}{.24\textwidth}
  \centering
  \begin{center}
    \includegraphics[width=1.\textwidth]{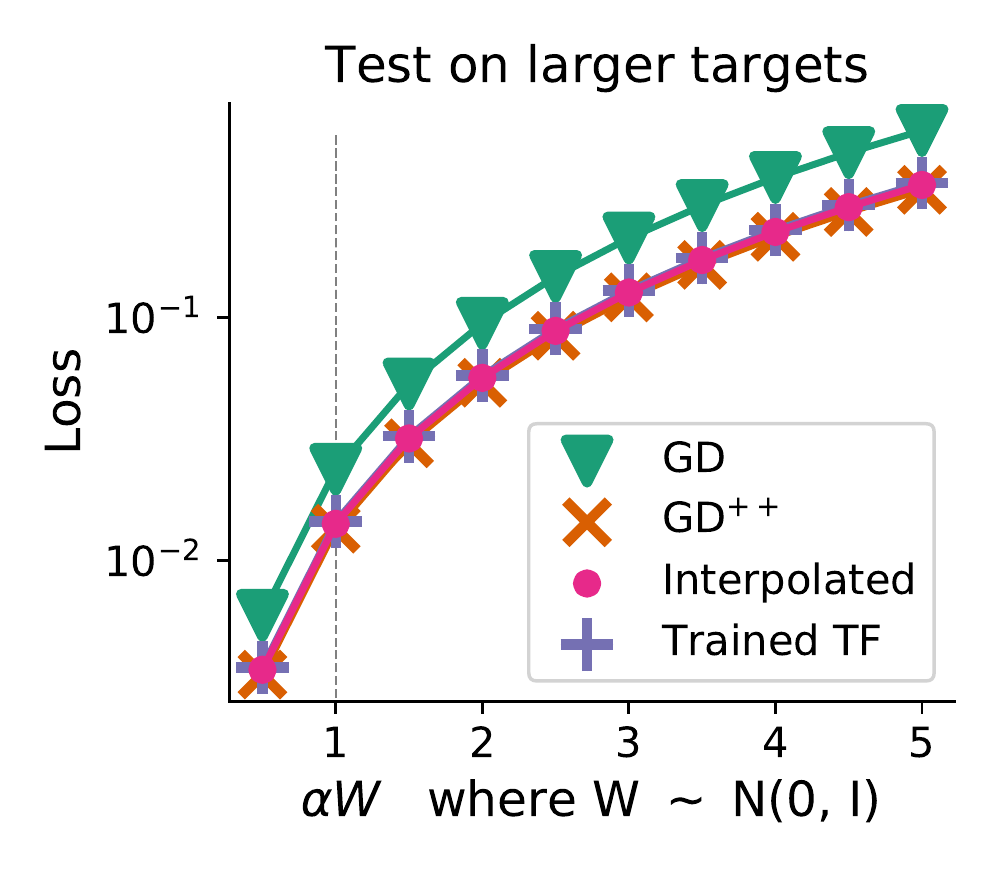}
  \end{center}
  \vspace{-10pt}
\end{minipage}
\end{center}

\textbf{(c) Comparing five steps of gradient descent with trained five layer Transformers on OOD data.}
\begin{center}
\begin{minipage}{.24\textwidth}
  \centering
  \begin{center}
    \includegraphics[width=1.\textwidth]{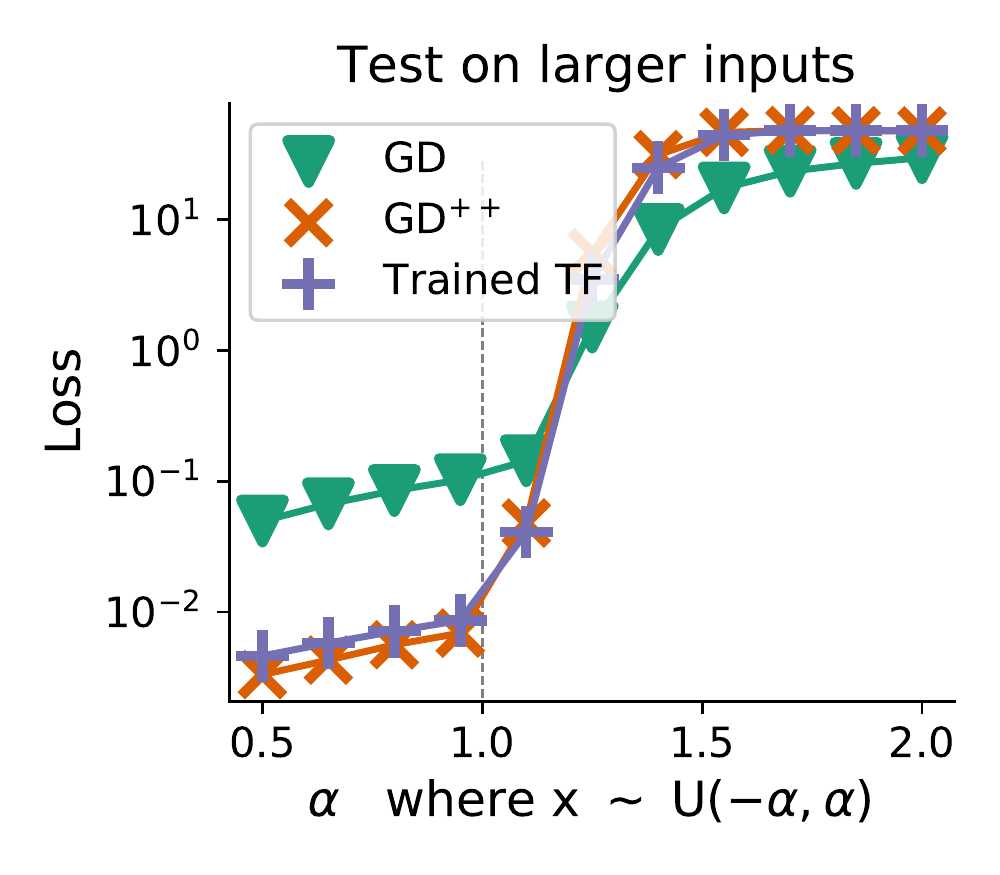}
  \end{center}
  \vspace{-10pt}
\end{minipage}
\begin{minipage}{.24\textwidth}
  \centering
  \begin{center}
    \includegraphics[width=1.\textwidth]{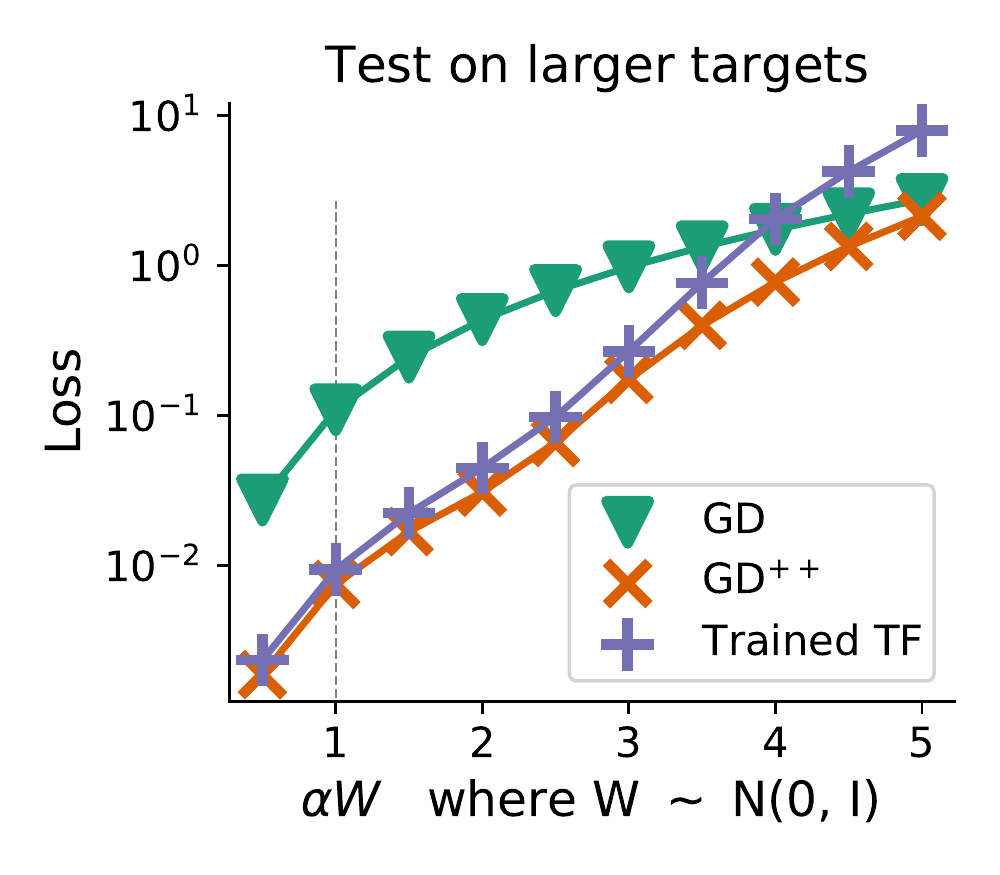}
  \end{center}
  \vspace{-10pt}
\end{minipage}
\begin{minipage}{.24\textwidth}
  \centering
  \begin{center}
    \includegraphics[width=1.\textwidth]{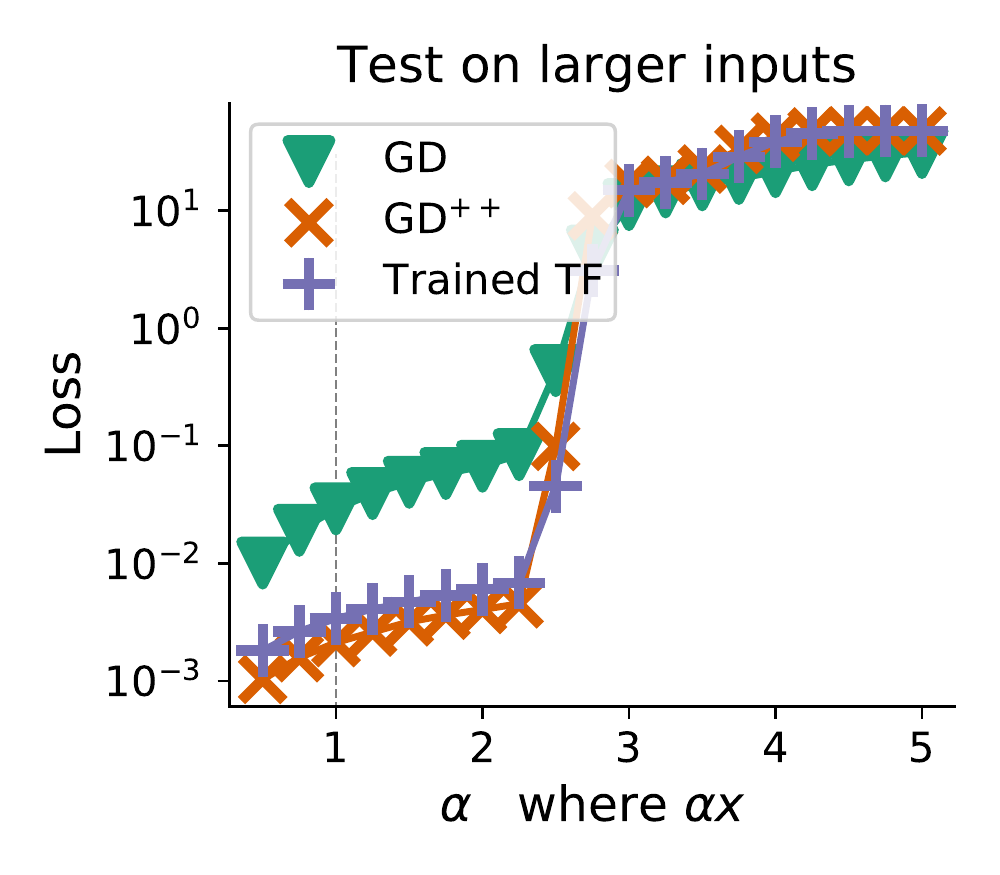}
  \end{center}
  \vspace{-10pt}
\end{minipage}
\begin{minipage}{.24\textwidth}
  \centering
  \begin{center}
    \includegraphics[width=1.\textwidth]{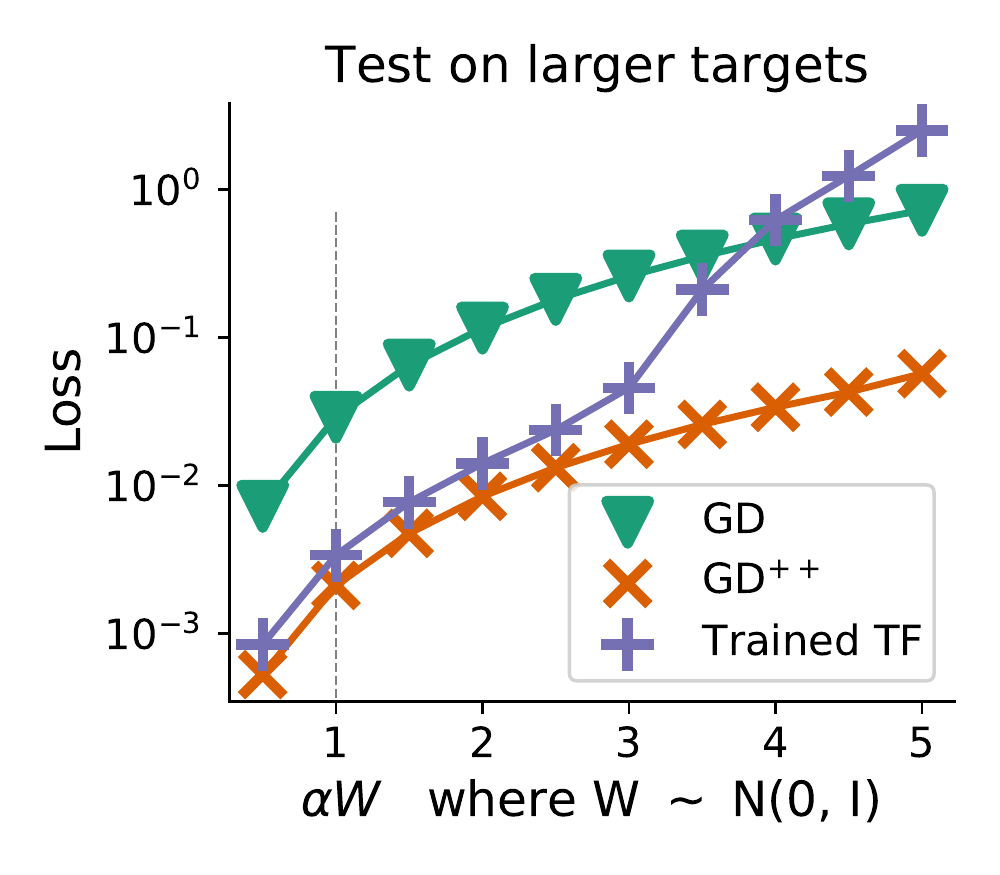}
  \end{center}
  \vspace{-10pt}
\end{minipage}
\end{center}
\vspace{-3pt}
  \caption{\textit{Left \& center left column:}  Comparing Transformers, GD and their weight interpolation on rescaled training distributions. In all setups, the trained Transformer behaves remarkably similar to GD or GD$^{++}$.
  \textit{Right \& center right}: Comparing Transformers, GD and their weight interpolation on data distributions never seen during training. Again, in all setups, the trained Transformer behaves remarkably similar to GD or GD$^{++}$ with less good match for deep non-recurrent Transformers far away from training regimes.}
  \label{fig:ood_big}
  \vspace{-10pt}
\end{figure*}

\begin{figure*}
\begin{center}
\begin{minipage}{.30\textwidth}
  \centering
  \begin{center}
    \includegraphics[width=1.\textwidth]{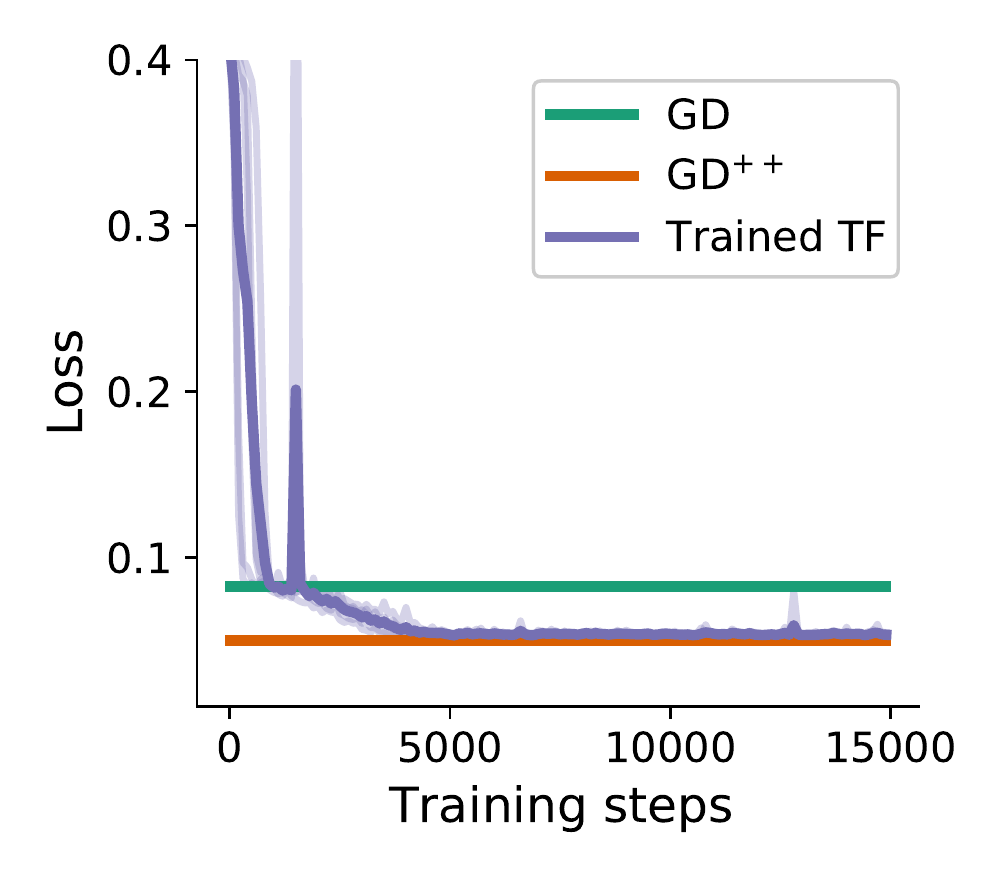}
  \end{center}
  \vspace{-10pt}
\end{minipage}
\begin{minipage}{.30\textwidth}
  \centering
  \begin{center}
    \includegraphics[width=1.\textwidth]{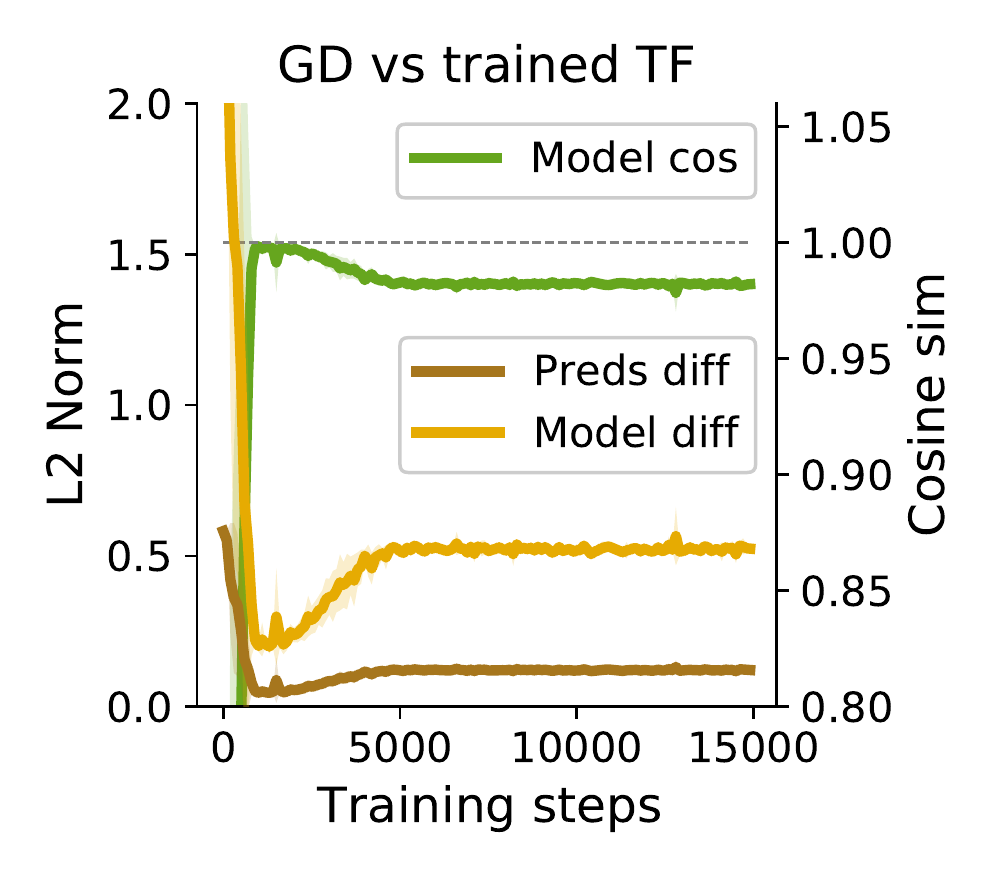}
  \end{center}
  \vspace{-10pt}
\end{minipage}
\begin{minipage}{.30\textwidth}
  \centering
  \begin{center}
    \includegraphics[width=1.\textwidth]{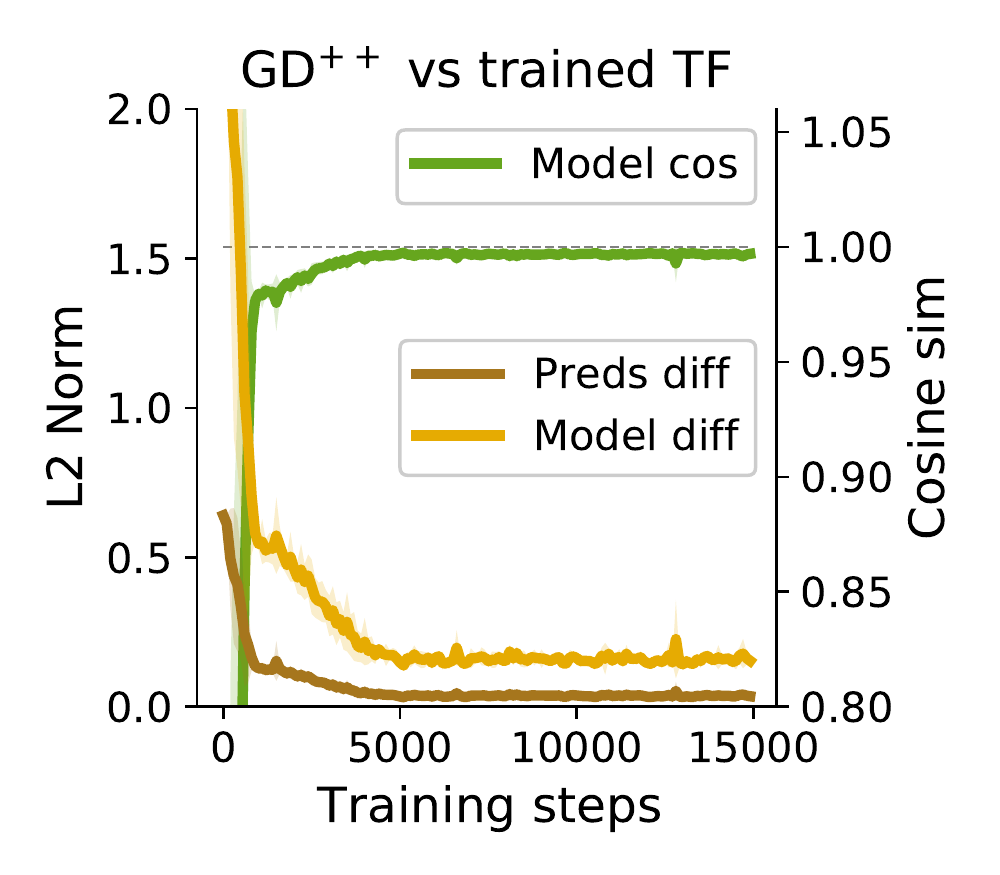}
  \end{center}
  \vspace{-10pt}
\end{minipage}
\end{center}

\begin{center}
\begin{minipage}{.24\textwidth}
  \centering
  \begin{center}
    \includegraphics[width=1.\textwidth]{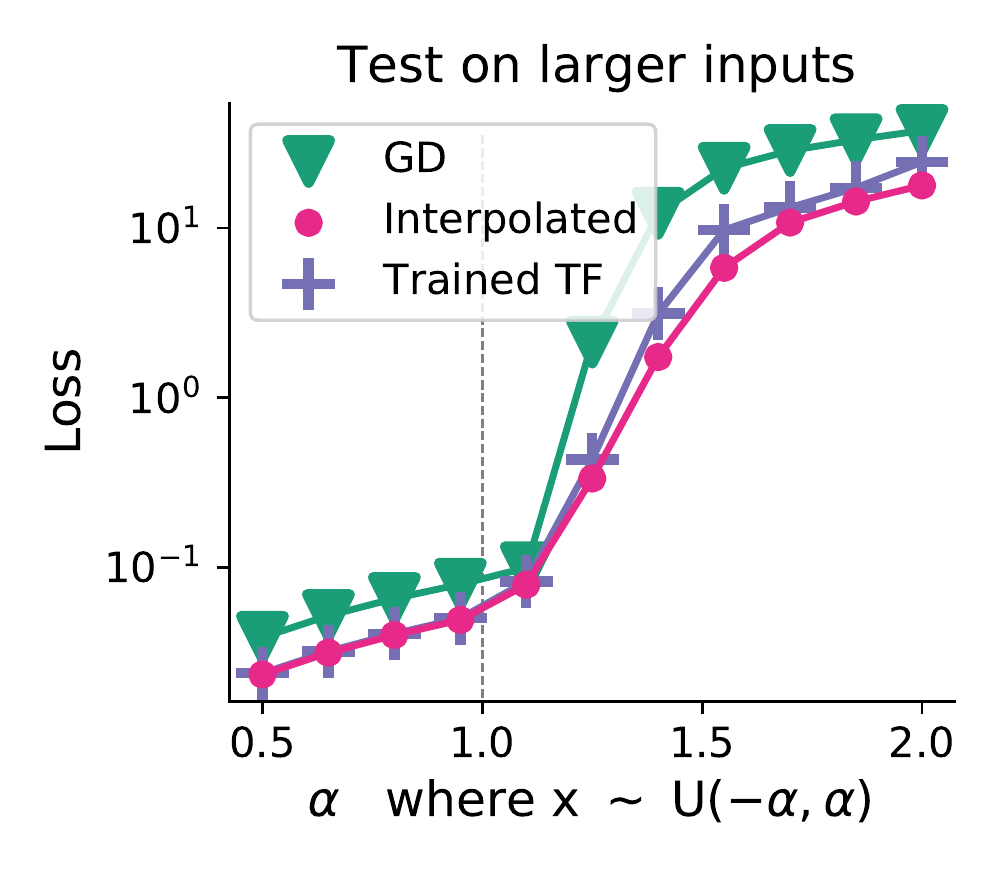}
  \end{center}
  \vspace{-10pt}
\end{minipage}
\begin{minipage}{.24\textwidth}
  \centering
  \begin{center}
    \includegraphics[width=1.\textwidth]{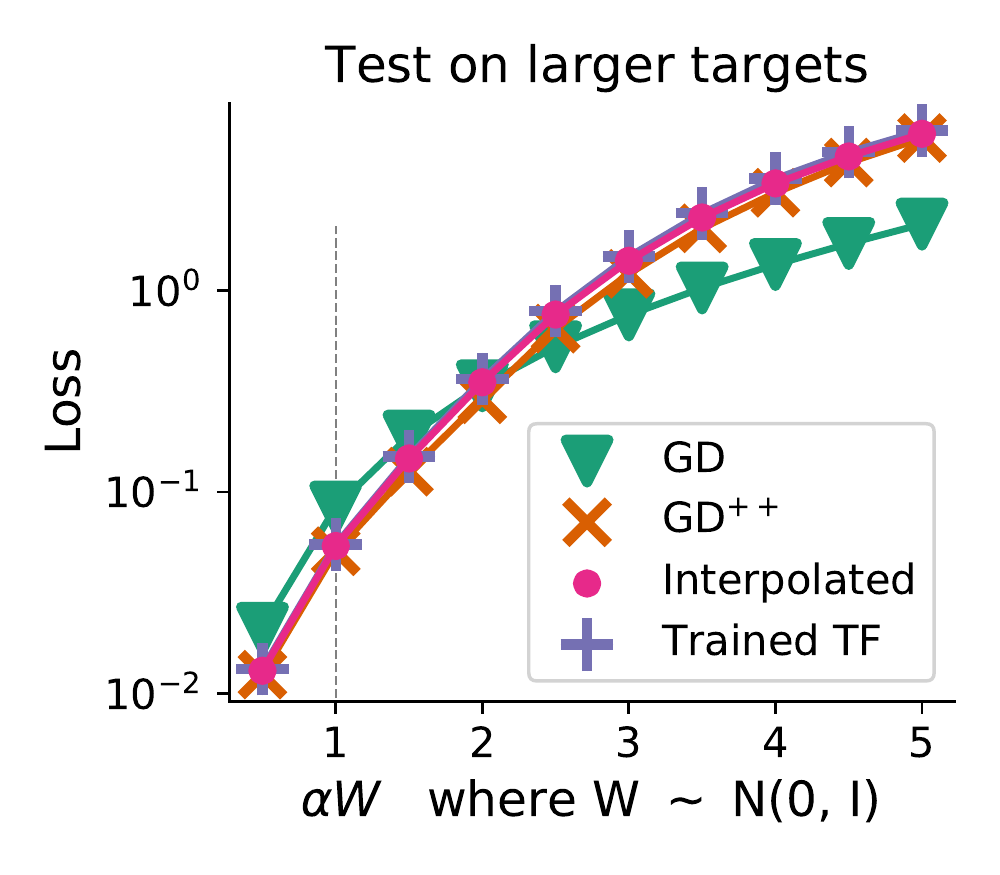}
  \end{center}
  \vspace{-10pt}
\end{minipage}
\begin{minipage}{.24\textwidth}
  \centering
  \begin{center}
    \includegraphics[width=1.\textwidth]{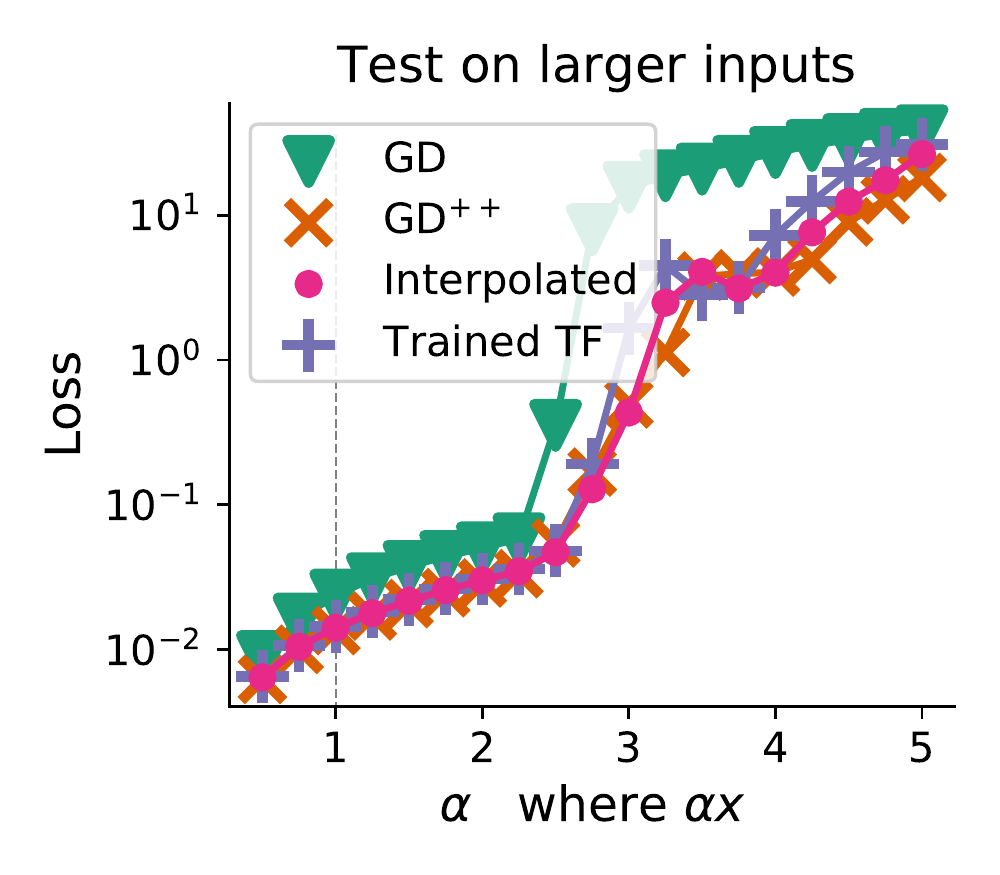}
  \end{center}
  \vspace{-10pt}
\end{minipage}
\begin{minipage}{.24\textwidth}
  \centering
  \begin{center}
    \includegraphics[width=1.\textwidth]{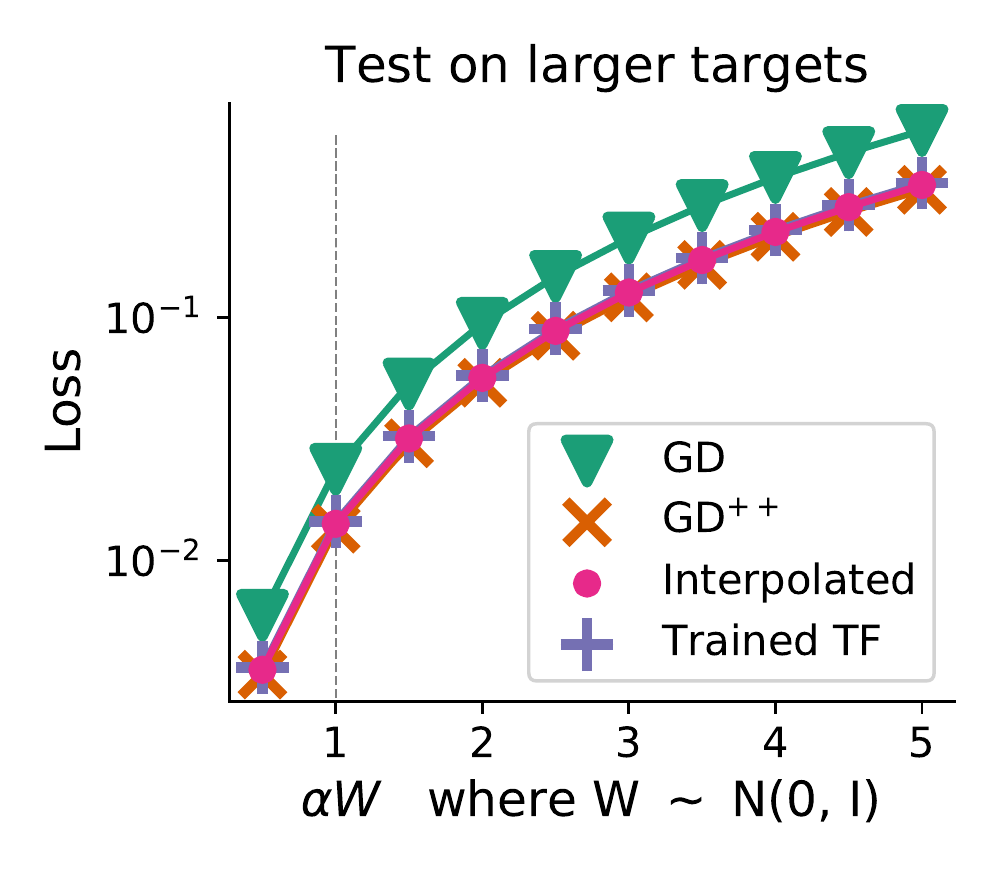}
  \end{center}
  \vspace{-10pt}
\end{minipage}
\end{center}

\vspace{-3pt}
  \caption{\textbf{Comparing ten steps of gradient descent with trained \textit{recurrent} ten-layer Transformers.} Results comparable to recurrent Transformer with two layers, see Figure \ref{fig:multi_layer}, but now with 10 repeated layers. We again observe for deeper recurrent linear self-attention only Transformers that overall GD$^{++}$ and the trained Transformer align very well with one another and are again interpolatable leading to very similar behavior insight as well as outside training situations. Note the inferior performance to the non-recurrent five-layer Transformer which highlights the importance on specific learning rate as well $\gamma$ parameter per layer/step. }
  \label{fig:ten_layer}
  \vspace{-10pt}
\end{figure*}

We provide more experimental results when comparing GD with tuned learning rate $\eta$ and data transformation scalar $\gamma$ 
and the trained Transformer on other data distributions than provided during training, see Figure \ref{fig:ood_big}. We do so by changing the in-context data distribution and measure the loss of both methods
averaged over 10.000 tasks when either changing  $\alpha$ that 1) affects the input data range $x \sim U(-\alpha, \alpha)^{N_x}$ or 2) the teacher by $\alpha W$ with $W \sim \mathcal{N}(0, I)$. This setups leads to results shown in the main text, in the first two columns of Figure \ref{fig:ood_big} and in the corresponding plots of Figure \ref{fig:ten_layer}. Although the match for deeper architectures starts to become worse, overall the trained Transformers behaves remarkably similar to GD and GD$^{++}$ for layer depth greater than 1. 

Furthermore, we try GD and the trained Transformer on input distributions that it never has seen during training. Here, we chose by chance of $1/3$ either a normal, exponential or Laplace distribution (with JAX default parameters) and depict the average loss value over 10.000 tasks where the $\alpha$ value now simply scales the input values that are sampled from one of the distributions $\alpha x$. The teacher scaling is identical to the one described above. See for results the two right columns of Figure \ref{fig:ood_big}, where we see almost identical behavior for recurrent architectures with less good match for deeper non-recurrent architectures far away from the training range of $\alpha = 1$.
Note that for deeper Transformers ($K>2$) the corresponding GD and GD$^{++}$ version, see for more experimental details Appendix section \ref{app:ex_det}, we include a harsh clipping of the token values after every step of transformation between $[-10, 10]$ (for the trained TF and GD) to improve training stability. Therefore, the loss increase is restricted to a certain value and plateaus.

\begin{figure*}
\begin{center}
\begin{minipage}{.30\textwidth}
  \centering
  \begin{center}
    \includegraphics[width=1.\textwidth]{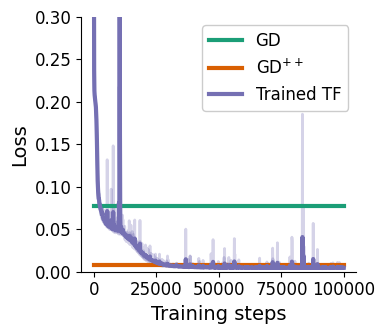}
  \end{center}
  \vspace{-10pt}
\end{minipage}
\begin{minipage}{.30\textwidth}
  \centering
  \begin{center}
    \includegraphics[width=1.\textwidth]{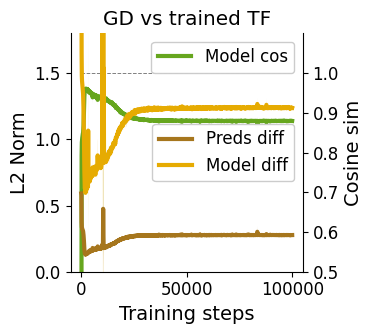}
  \end{center}
  \vspace{-10pt}
\end{minipage}
\begin{minipage}{.30\textwidth}
  \centering
  \begin{center}
    \includegraphics[width=1.\textwidth]{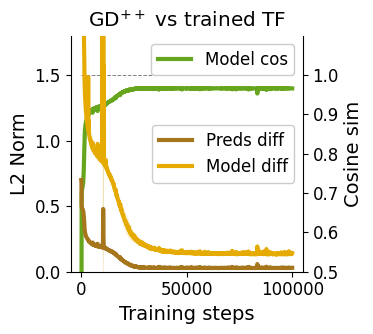}
  \end{center}
  \vspace{-10pt}
\end{minipage}
\end{center}
\vspace{-3pt}
  \caption{\textbf{Comparing twelve steps of GD$^{++}$ with a trained twelve-layer Transformers with MLPs and 4 headed linear self-attention layer.} Results comparable to the deep recurrent Transformer, see Figure \ref{fig:ten_layer}, but now with 12 independent Transformer blocks including MLPs and 4-head linear self-attention. We omit LayerNorm. We again observe a close resemblance of the trained Transformers and GD$^{++}$. We hypotheses that even when equipped with multiple heads and MLPs, Transformers approximate GD$^{++}$.}
  \label{fig:twelve_layer}
\end{figure*}

\subsection{Linear mode connectivity between the weight construction of Prop \ref{prop:self_att_gd} and trained Transformers }
\label{app:mode_connect}

In order to interpolate between the construction $\theta_{\text{GD}}$
and the trained weights of the Transformer $\theta$, we need to correct for some scaling ambiguity. For clarification, we restate here the linear self-attention operation for a single head
\begin{align}
e_j  \leftarrow & e_j + P W_{V}\sum_{i} e_{i} \otimes e_{i} W_{K}^T W_{Q}e_{j} \\
 &= e_j + W_{PV} \sum_{i} e_{i} \otimes e_{i} W_{KQ} e_{j} 
\end{align}
Now, to match the weight construction of Prop. \ref{prop:self_att_gd} we have the aim for the matrix product $W_{KQ}$ to match an identify matrix (except for the last diagonal entry) after re-scaling. Therefore we compute the mean of the diagonal of the matrix product of the trained Transformer weights $W_{KQ}$ which we denote by $\beta$.  After resealing both operations i.e. $W_{KQ} \leftarrow W_{KQ}/\beta$
and $W_{PV} \leftarrow W_{PV}\beta$ we interpolate linearly between the matrix products of GD as well as these rescaled trained matrix products i.e. $W_{I, KQ} = (W_{GD, KQ} + W_{TF, KQ})/2$ as well as 
$W_{I, PV} = (W_{GD, PV} + W_{TF, PV})/2$. We use these parameters to obtain results throughout the paper denote with \textit{Interpolated}.
We do so for GD as well as GD$^{++}$ when comparing to recurrent Transformers. Note that for non-recurrent Transformers, we face more ambiguity that we have to correct for since e.g. scalings influence each other across layer. We also see this in practice and are not able (only for some seeds) to interpolate between weights with our simple correction from above. We leave the search for more elaborate corrections for future work.  

\subsection{Visualizing the trained Transformer weights}

\begin{figure*}
\centering
\begin{minipage}{.52\textwidth}
  \centering
  \begin{center}
    \includegraphics[width=1.\textwidth]{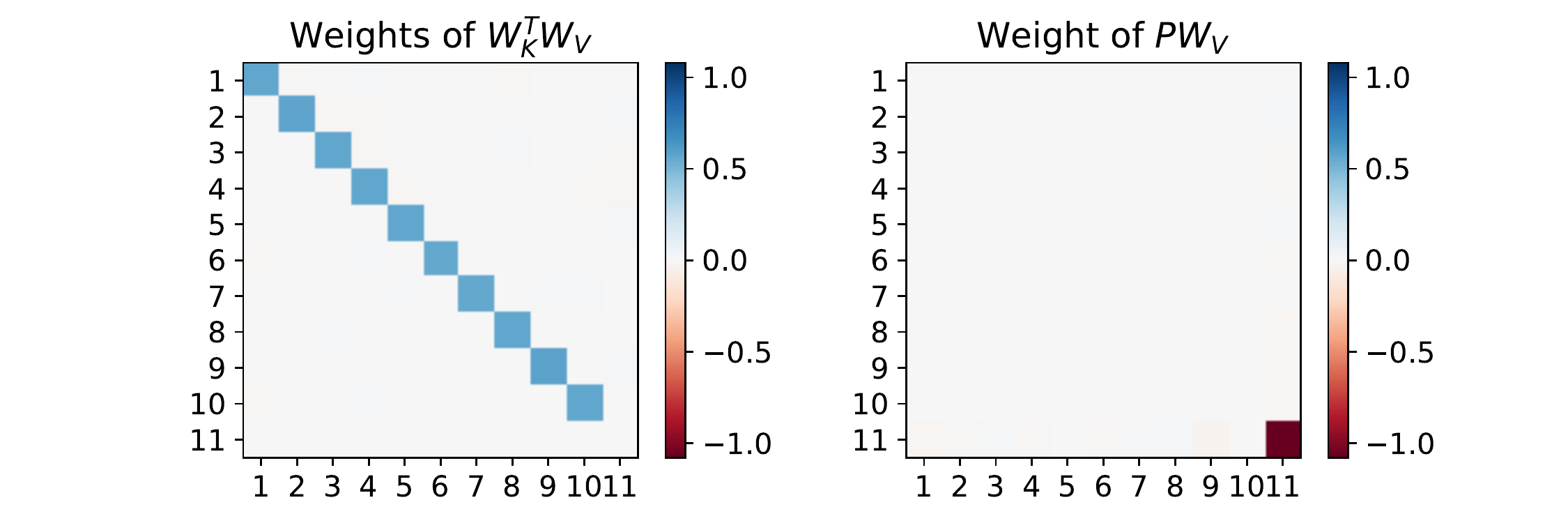}
  \end{center}
\end{minipage}
 \hspace{-40pt}
\begin{minipage}{.52\textwidth}
  \centering
  \begin{center}
    \includegraphics[width=1.\textwidth]{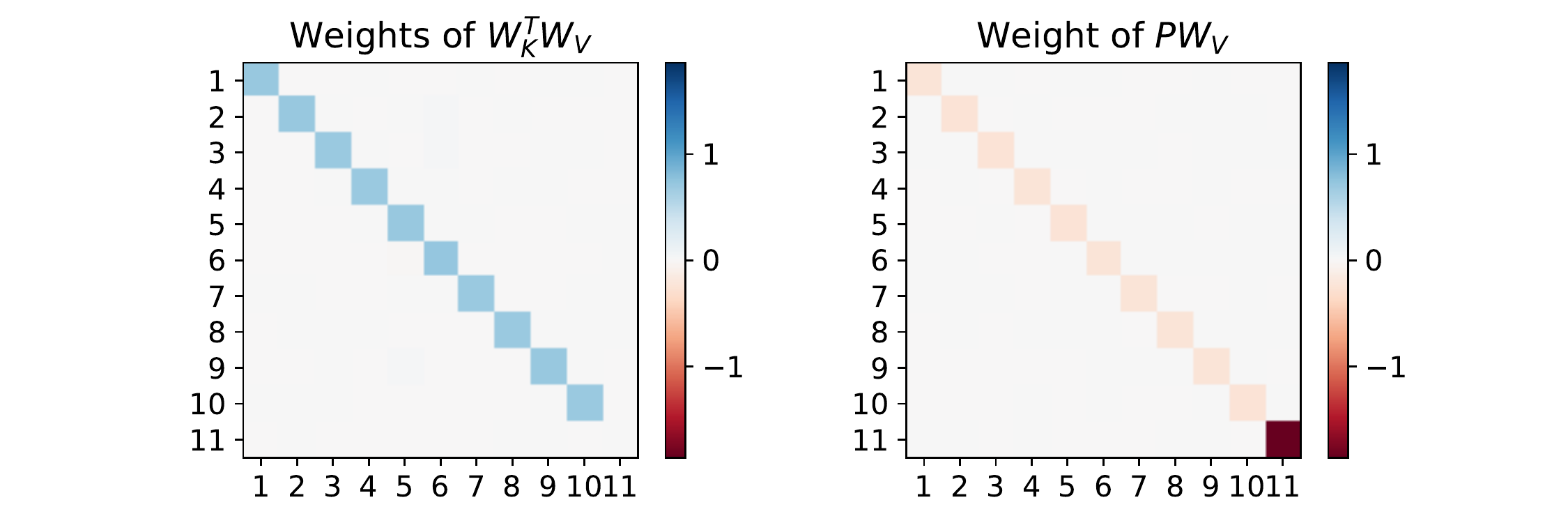}
  \end{center}
\end{minipage}
\hspace{-0pt}
  \caption{\textbf{Visualizing the weight matrices of trained Transformers}. \textit{Left \& outer left:} Weight matrix products of a trained single linear self-attention layer. We see (after scalar correction) a perfect resemblance of our construction. \textit{Right \& outer right:} Weight matrix products of a trained 3-layer recurrent linear self-attention Transformer. Again, we see (after scalar correction) a perfect resemblance of our construction and an additional curvature correction i.e. diagonal values in $PW_V$ of the same magnitude except the last entry that functions as the learning rate.}
  \label{fig:weight_vis}
  \vspace{-10pt}
\end{figure*}

The simplicity of our construction enables us to visually compare trained Transformers and the construction put forward in Proposition \ref{app:prop1} in weight space. As discussed in the previous section \ref{app:mode_connect} there is redundancy in the way the trained Transformer can construct the matrix products leading to the weights corresponding to gradient descent. We therefore visualize $W_{KQ}=W^T_KW_Q$ as well as $W_{PV}=P_KW_V$ in Figure \ref{fig:weight_vis}. 

\subsection{Proof and discussion of Proposition  \ref{prop:token_copy}}
\label{app:prop2}
\setcounter{prop}{2}
We state here again Proposition \ref{prop:token_copy}, provide the necessary construction and a short discussion.
\begin{prop}
Given a 1-head linear- or softmax attention layer and the token construction $e_{2j} = (x_j), e_{2j+1} = (0, y_j)$ with a zero vector $0$ of dim $N_x -N_y$ and concatenated positional encodings, one can construct key, query and value matrix $W_K, W_Q, W_V$ as well as the projection matrix $P$ such that all tokens $e_j$ are transformed into tokens equivalent to the ones required in proposition \ref{prop:self_att_gd}. 
\end{prop}

To get a simple and clean construction, we choose wlog $x_j \in \mathbb{R}^{2N+1}$ and $(0, y_j) \in \mathbb{R}^{2N+1}$ as well as model the positional encodings as unit vectors $p_j \in \mathbb{R}^{2N+1}$ and  concatenate them to the tokens i.e. $e_j = (x_{j/2}, p_j)$. We wish for a construction that realizes 
\begin{align}
     e_j  \leftarrow& \left(\begin{array}{@{}c@{}}
  x_{j/2}\\
  p_j 
\end{array}\right)
 + PVK^{T}W_{Q}  \left(\begin{array}{@{}c@{}}
  x_{j/2}\\
  p_j 
\end{array}\right) \\
        &=  \left(\begin{array}{@{}c@{}}
  x_{j/2}\\
  p_j 
\end{array}\right) + \left(\begin{array}{@{}c@{}}
  0\\
   y_{j/2+1} - p_j
\end{array}\right).
\end{align}
This means that a token replaces its own positional encoding by coping the target data of the next token to itself leading to $e_j=(x_{j/2}, 0, y_{j/2+1})$, with slight abusive of notation. This can simply be realized by (for example) setting $P=I$, $W_V=\left(\begin{array}{@{}c c@{}}
  0
  & 0 \\
  I_x &
  -I_{x, off}
\end{array}\right), W_K=\left(\begin{array}{@{}c c@{}}
  0
  & 0 \\
  0 &
  I_x
\end{array}\right)$ and  $W_Q=\left(\begin{array}{@{}c c@{}}
  0
  & 0 \\
  0 &
  I_{x, off}^T
\end{array}\right)$ with $I_{x, off}$ the lower diagonal identity matrix fo size $N_x$. Note that then simply $K^TW_Qe_j=p_{j+1}$ i.e. it chooses the $j+1$ element of $V$ which stays $p_{j+1}$ if we apply the $\text{softmax}$ operation on $K^Tq_j$. Since the $j+1$ entry of $V$ is $(0, y_{j/2+1} - p_{j})$ we obtain the desired result.

For the (toy-)regression problems considered in this manuscript, the provided result would give $N/2$ tokens for which we also copy (parts) of $x_j$ underneath $y_j$. This is desired for modalities such as language where every two tokens could be considered an in-and output pair for the implicit autoregressive inner-loop loss. These tokens do not have be necessarily next to each other, see for this behavior experimental findings presented in \cite{induction_heads}. For the experiments conducted here, one solution is to zero out these tokens which could be constructed by a two-head self-attention layer that given uneven $j$ simply subtracts itself resulting in a zero token. For all even tokens, we use the construction from above which effectively coincides with the token construction required in Proposition \ref{prop:self_att_gd}.

\subsection{Dampening the self-attention layer}
\label{app:dampening}

\begin{figure*}
\textbf{Rolling out experiment with different dampening strength}
\begin{center}
\begin{minipage}{.30\textwidth}
  \centering
  \begin{center}
    \includegraphics[width=1.\textwidth]{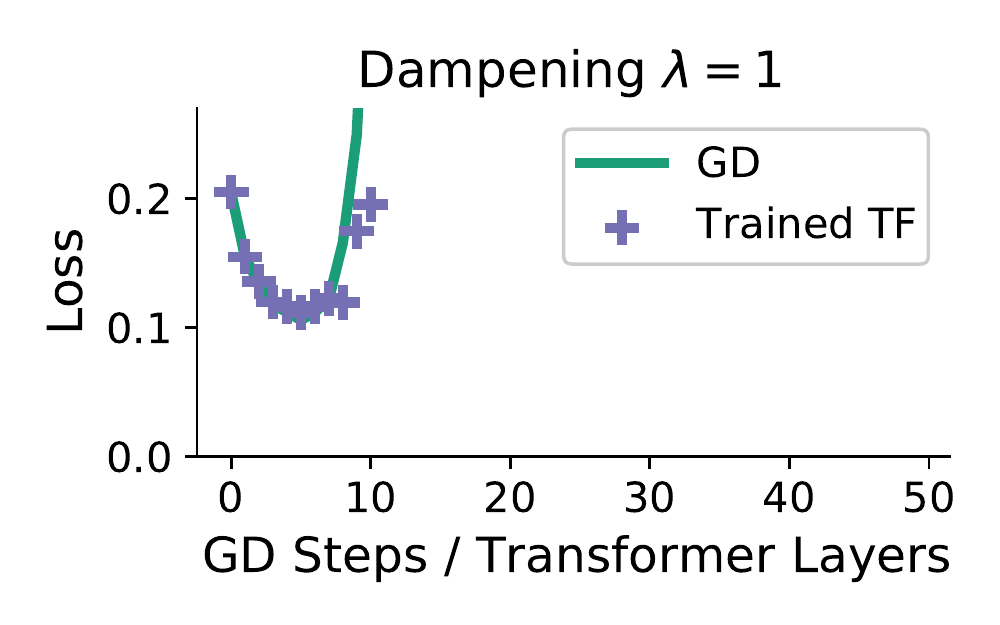}
  \end{center}
  \vspace{-10pt}
\end{minipage}
\begin{minipage}{.30\textwidth}
  \centering
  \begin{center}
    \includegraphics[width=1.\textwidth]{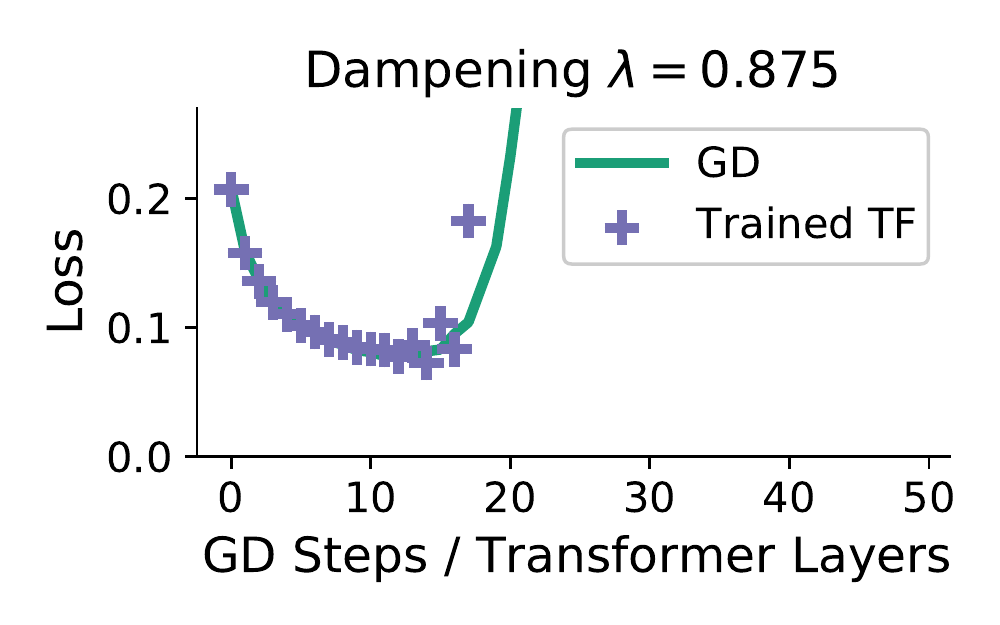}
  \end{center}
  \vspace{-10pt}
\end{minipage}
\begin{minipage}{.30\textwidth}
  \centering
  \begin{center}
    \includegraphics[width=1.\textwidth]{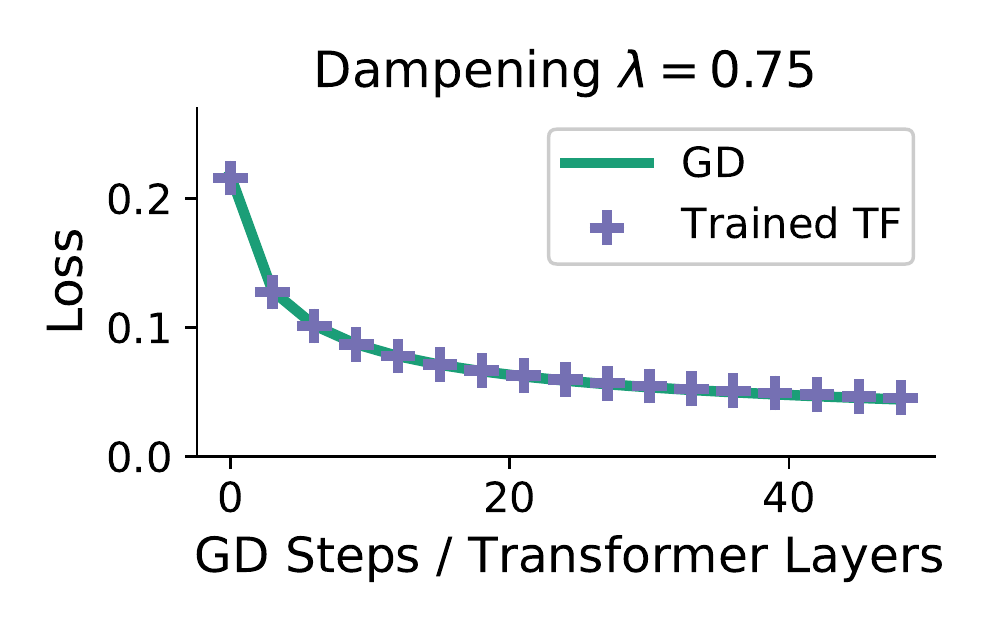}
  \end{center}
  \vspace{-10pt}
\end{minipage}
\end{center}
\vspace{-3pt}
  \caption{\textbf{Roll-out experiments: applying a trained single linear self-attention layer multiple times.} We observe that different dampening strengths affect the generalization of both methods with slightly better robustness for GD which matching performance for 50 steps when $\lambda=0.75$.}
  \label{fig:rolling_out}
\end{figure*}

As an additional out-of-distribution experiment, we test the behavior when repeating a single LSA-layer trained to lower our objective, see \eqref{eq:trans_loss}, with the aim to repeat the learned learning/update rule. Note that GD as well as the self-attention layer were optimized to be optimal for one step. For GD we line search the otpimal learning rate $\eta$ on 10.000 task.
Interestingly, for both methods we observe quick divergence when applied multiple times, see left plot of Figure \ref{fig:rolling_out}. Nevertheless, both of our update functions are described by a linear self-attention layer for which we can control the norm, post training, by a simple scale which we denote as $\lambda$. This results in the new update $y_{\text{test}} + \lambda \Delta Wx_{\text{test}}$ for GD and $y_{\text{test}} + \lambda PVK^TW_Qx_{\text{test}}$ for the trained self-attention layer which effectively re-tunes the learning rate for GD and the trained self-attention layer. Intriguingly, both methods do generalize similarly well (or poorly) on this out-of-distribution experiment when changing $\lambda$, see again Figure \ref{fig:rolling_out}. We show in Figure \ref{fig:illus} the behavior for $\lambda=0.75$ for which we see both methods steadily decreasing the loss within 50 steps. 

\subsection{Sine wave regression}
\label{app:non_linear}

For the sine wave regression tasks, we follow \cite{finn_model-agnostic_2017} and other meta-learning literature and sample for each task an amplitude $a \sim U(0.1, 5)$ and a phase $\rho \sim U(0, \pi)$. Each tasks consist of $N=10$ data points where inputs are sampled $x \sim U(-5, 5)$ and targets computed by $y = a \sin(\rho + x)$.
We choose here for the first time, for GD as well as for the Transformer, an input embedding $\text{emb}$ that maps tokens $e_i = (x_i, y_i)$ into a $40$ dimensional space $\text{emb}(e_i) =  W_{\text{emb}}e_i$ through an affine projection without bias. We skip the first self-attention layer but, as usually done in Transformers, then transform the embedded tokens through an MLP $m$ with a single hidden layer, widening factor of 4 (160 hidden neuros) and GELU nonlinearity \citep{gelu} i.e. $e_j \leftarrow m(\text{emb}(e_j)) + \text{emb}(e_j)$. 

We interpret the last entry of the transformed tokens as the (transformed) targets and the rest as a higher-dimensional input data representation on which we train a model with a single gradient descent step. We compare the obtained meta-learned GD solution with training a Transformer on the same token embeddings but instead learn a self-attention layer. Note that the embeddings of the tokens, including the transformation through the MLP, are not dependent on an interplay between the tokens. Furthermore, the initial transformation is dependent on $e_i = (x_i, y_i)$, i.e., input as well as on the target data except for the query token for which $y_{\text{test}}=0$. This means that this construction is, except for the additional dependency on targets, close to a large corpus of meta-learning literature that aims to find a deep representation optimized for (fast) fine tuning and few-shot learning.
In order to compare the meta-training of the MLP and the Transformer, we choose the same seed to initialize the network weights for the MLPs and the input embedding trained by meta-learning i.e. backprop through training or the Transformer. This leads to the plots and almost identical learned initial function and updated functions shown in Figure \ref{fig:non_linear}.

\subsection{Proposition \ref{prop:kernel} and connections between gradient descent, kernelized regression and kernel smoothing}
\label{app:ker_reg}

Let's consider the data transformation induced by an MLP $\tilde{m}(x)$ and a residual connection commonly used in Transformer blocks i.e. $e_j \leftarrow e_j + \tilde{m}(e_j) = (x_j, y_j) + (\tilde{m}(x_j), 0) = (m(x_j), y_j)$ with $m(x_j) = x_j + \tilde{m}(x_j)$ and $\tilde{m}$ not changing the targets $y$. When simply applying Proposition \ref{prop:self_att_gd}, it is easy to see that given this new token construction, a linear self-attention layer can induce the token dynamics $e_j \leftarrow (m(x_j), y_j) + (0, -\Delta W m(x_j))$ with $\Delta W = -\eta \nabla L(W)$ given the loss function $L(W) = \frac{1}{2N}\sum_{i=1}^N||W m(x_i)  -y_i||^2$.

Interestingly, for the test token $e_{\text{test}} = (x_{\text{test}}, 0)$ this induces, after a multiplication with $-1$, an initial prediction after a single Transformer block given by
\begin{equation}
    \hat{y} = \Delta W m(x_{\text{test}}) = -\eta \nabla_W L(0) m(x_{\text{test}}) = \sum_{i=1}^N y_i m(x_i)^Tm(x_{\text{test}}) = \sum_{i=1}^N y_i k(x_i, x_{\text{test}})
\end{equation}
with $m(x_i)^Tm(x_{\text{test}})=k(x_i, x_{\text{test}}) \in\mathbb{R}$ interpreted as a kernel function.
Concluding, we see that the combination of MLPs and a \textit{single} self-attention layer can lead to dynamics induced when descending a kernelized regression (squared error) loss with a \textit{single} step of gradient-descent. 

Interestingly, when choosing $W_0=0$, we furthermore see that a single self-attention layer or Transformer block can be regarded as doing nonparametric kernel smoothing $\hat{y}=\sum_{i=1}^N y_i k(x_i, x_{\text{test}})$  based on the data given in-context \citep{nadaraya1964estimating,watson1964smooth}. Note that we made a particular choice of kernel function here and that this view still holds when $m(x_j)=\mathbbm{1}$ i.e.~consider Transformers without MLPs or leverage the well-known view of softmax self-attention layer as a kernel function used to measure similarity between tokens \citep[e.g.][]{choromanski2021rethinking,zhang2021dive}. 
Thus, implementing one step of gradient descent through a self-attention layer (w/wo softmax nonlinearity) is equivalent to performing kernel smoothing estimation. We however argue that this nonparametric kernel smoothing view of in-context learning is limited, and arises from looking only at a \textit{single} self-attention layer.
When considering deeper Transformer architectures, we see that multiple Transformer blocks can iteratively transform the targets based on multiple steps of gradient descent leading to minimization of a kernelized squared error loss $L(W)$. One way to obtain a suitable construction is by neglecting MLPs everywhere except in the first Transformer block. 
We leave the study of the exact mechanics, especially how the Transformer makes use of possibility transforming the targets through the MLPs, and the possibility of iteratively changing the kernel function throughout depth for future study.

\subsection{Linear vs.~softmax self-attention as well LayerNorm Transformers}
\label{app:softmax}

\label{ref:linear_vs_softmax} Although linear Transformers and their variants have been shown to be competitive with their softmax counterpart \citep{irie2021going}, the removal of this nonlinearity is still a major departure from classic Transformers and more importantly from the Transformers used in related studies analyzing in-context learning. In this section we investigate whether and when gradient-based learning emerges in trained softmax self-attention layers, and we provide an analytical argument to back our findings.

\begin{figure*}
\textbf{(a) Comparing one step of GD with a trained \textit{softmax one}-headed self-attention layer.}
\begin{center}
\begin{minipage}{.22\textwidth}
  \centering
  \begin{center}
    \includegraphics[width=1.\textwidth]{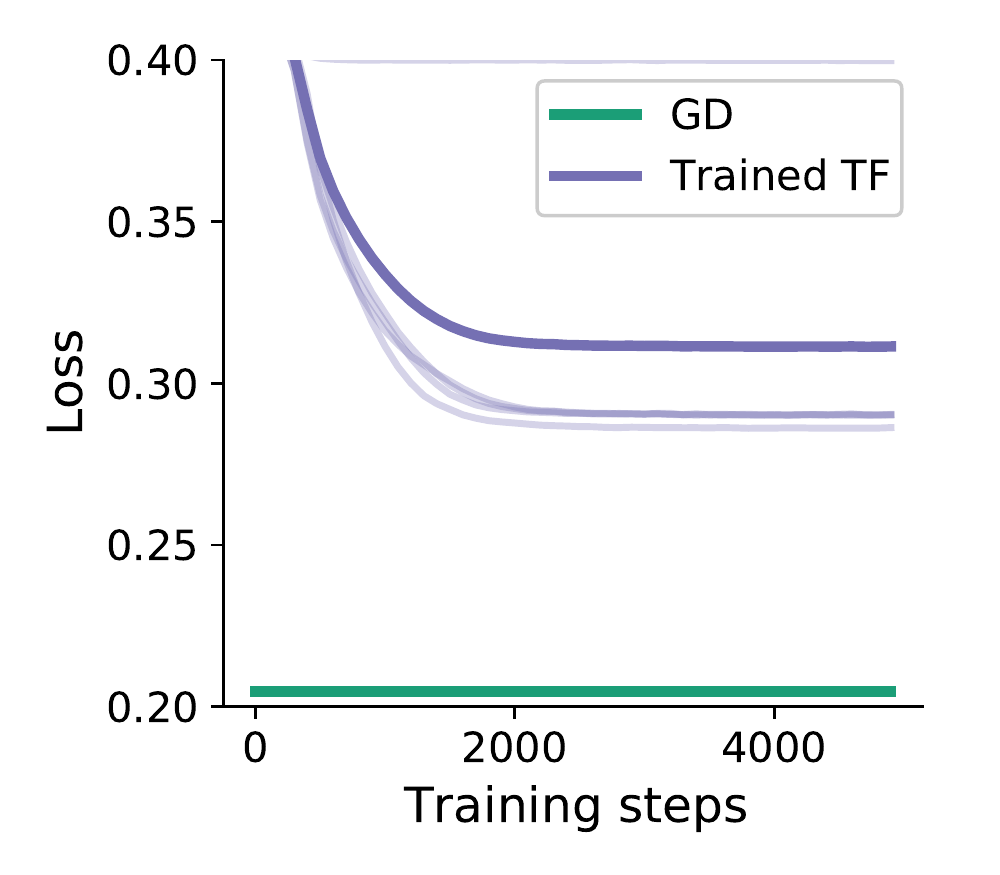}
  \end{center}
  \vspace{-10pt}
\end{minipage}
\begin{minipage}{.28\textwidth}
  \centering
  \begin{center}
    \includegraphics[width=1.\textwidth]{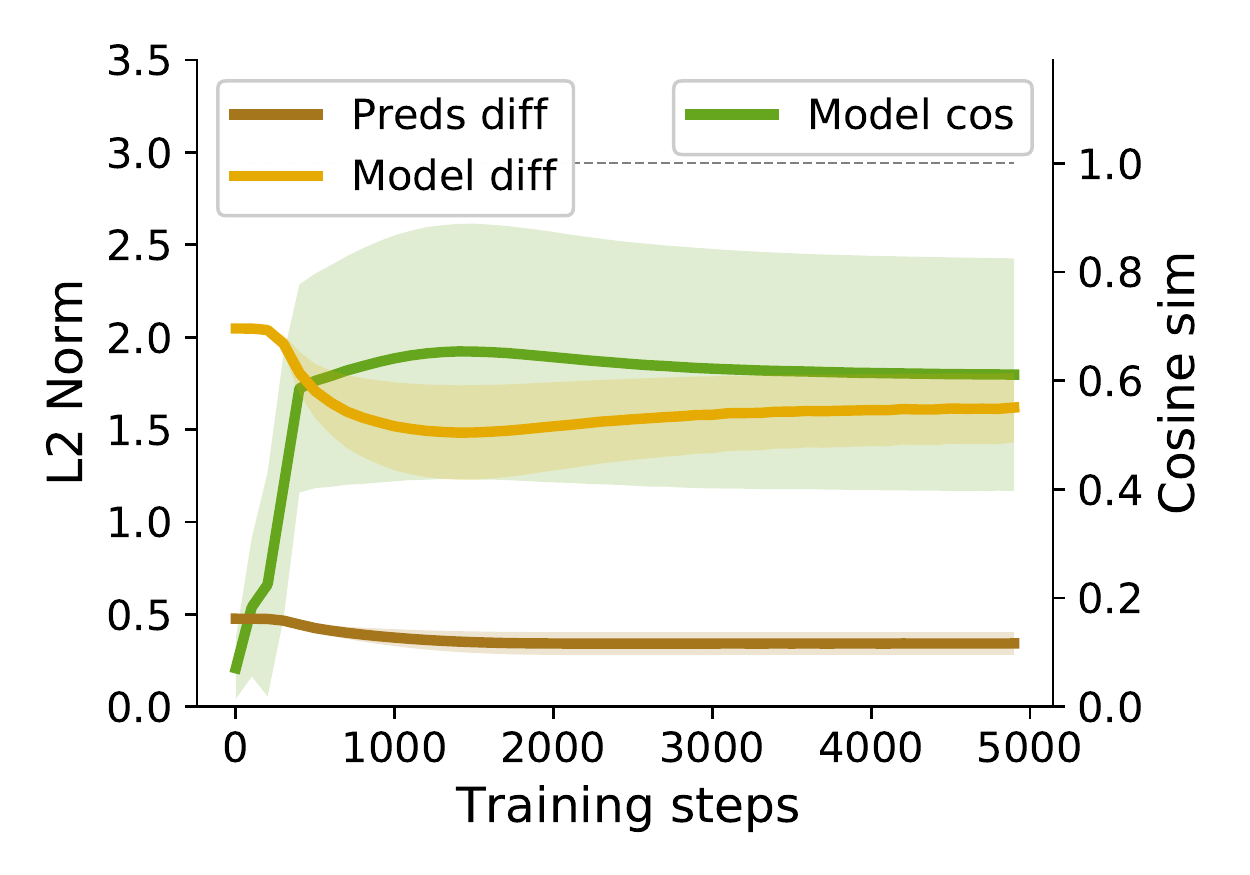}
  \end{center}
  \vspace{-10pt}
\end{minipage}
\begin{minipage}{.23\textwidth}
  \centering
  \begin{center}
    \includegraphics[width=1.\textwidth]{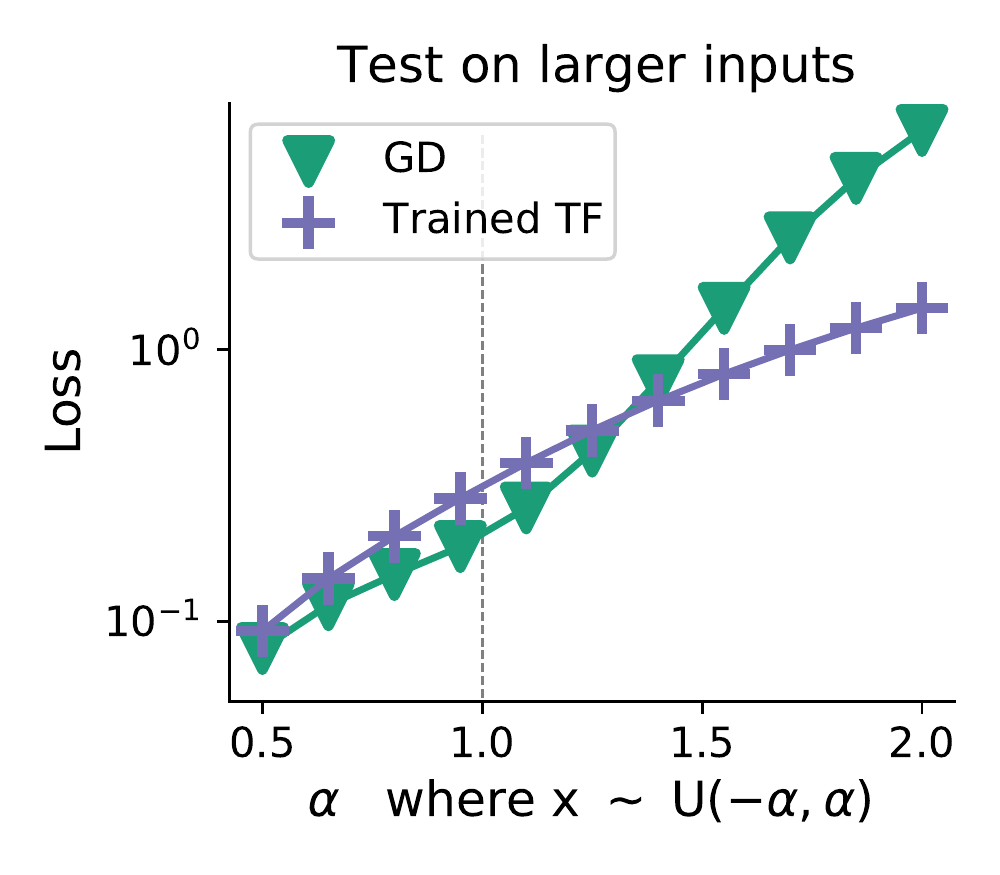}
  \end{center}
  \vspace{-10pt}
\end{minipage}
\begin{minipage}{.23\textwidth}
  \centering
  \begin{center}
    \includegraphics[width=1.\textwidth]{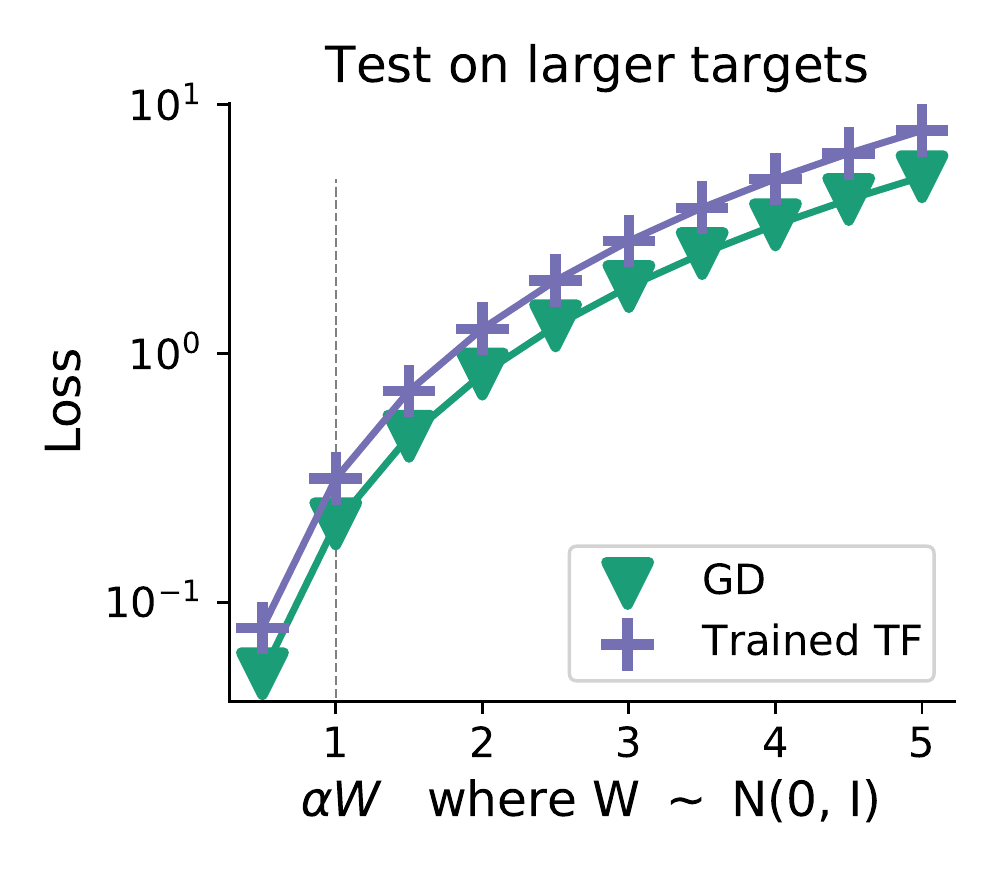}
  \end{center}
  \vspace{-10pt}
\end{minipage}
\end{center}
\textbf{(b) Comparing one step of GD with a trained \textit{softmax two}-headed self-attention layer.}
\begin{center}
\begin{minipage}{.22\textwidth}
  \centering
  \begin{center}
    \includegraphics[width=1.\textwidth]{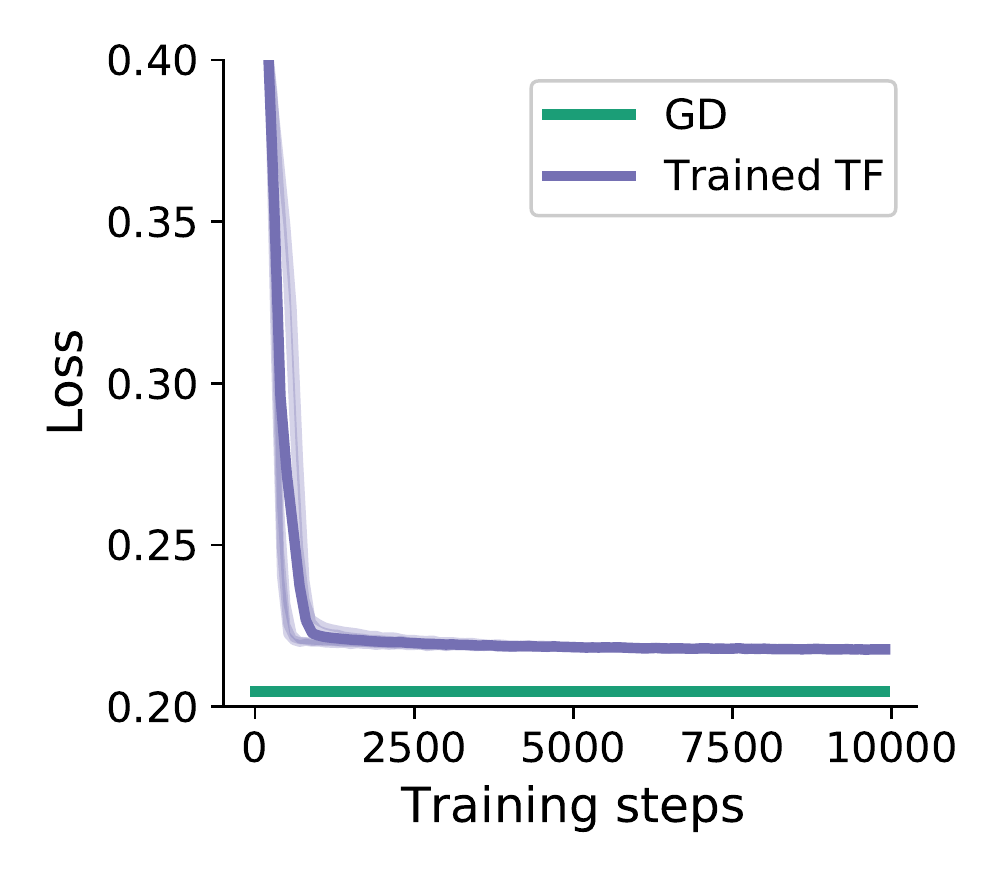}
  \end{center}
  \vspace{-10pt}
\end{minipage}
\begin{minipage}{.28\textwidth}
  \centering
  \begin{center}
    \includegraphics[width=1.\textwidth]{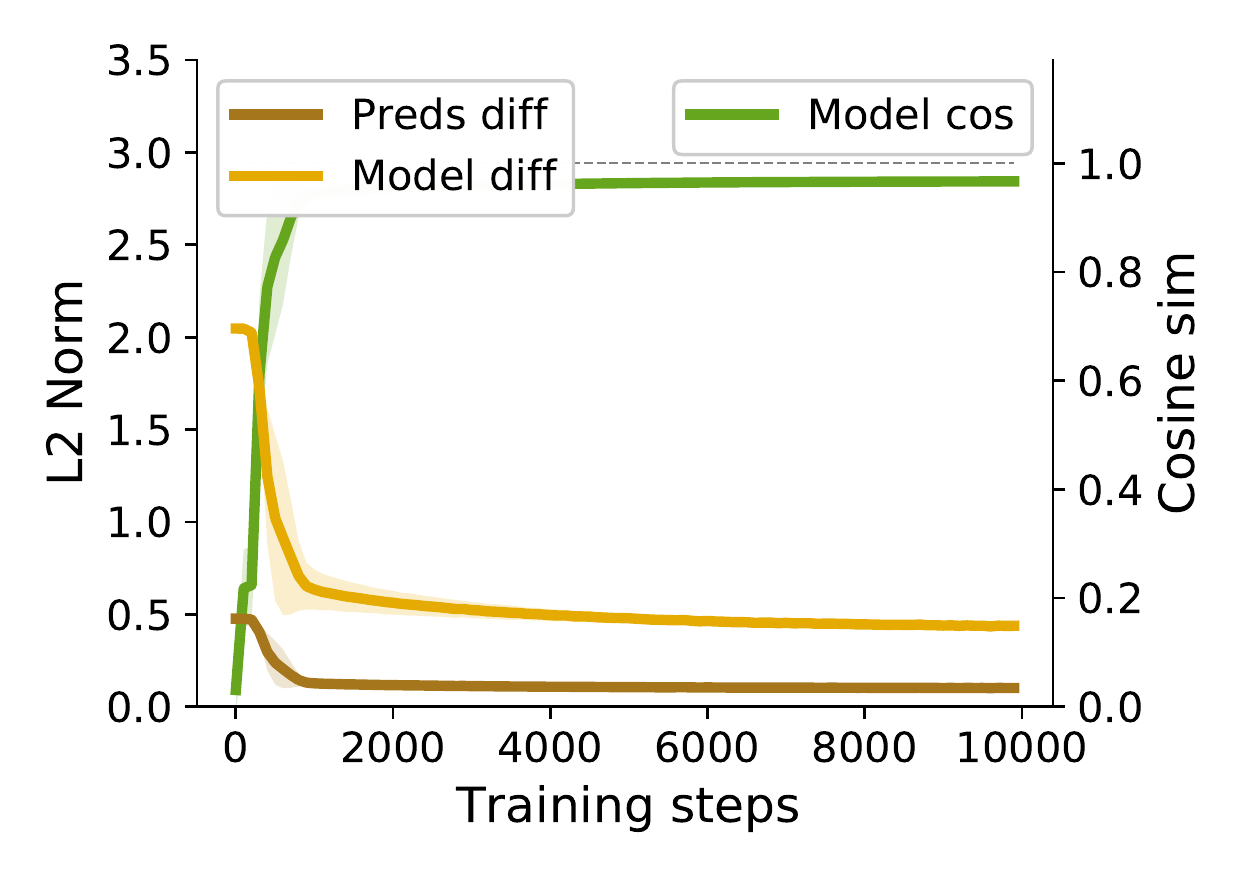}
  \end{center}
  \vspace{-10pt}
\end{minipage}
\begin{minipage}{.23\textwidth}
  \centering
  \begin{center}
    \includegraphics[width=1.\textwidth]{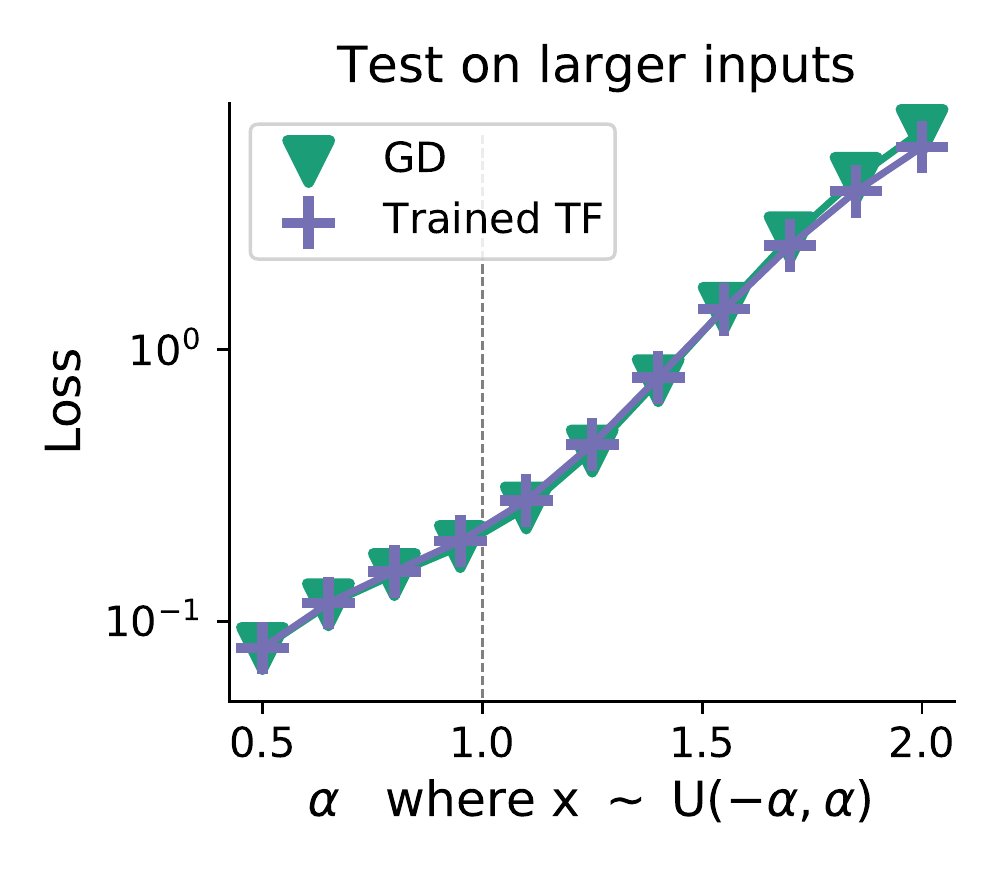}
  \end{center}
  \vspace{-10pt}
\end{minipage}
\begin{minipage}{.23\textwidth}
  \centering
  \begin{center}
    \includegraphics[width=1.\textwidth]{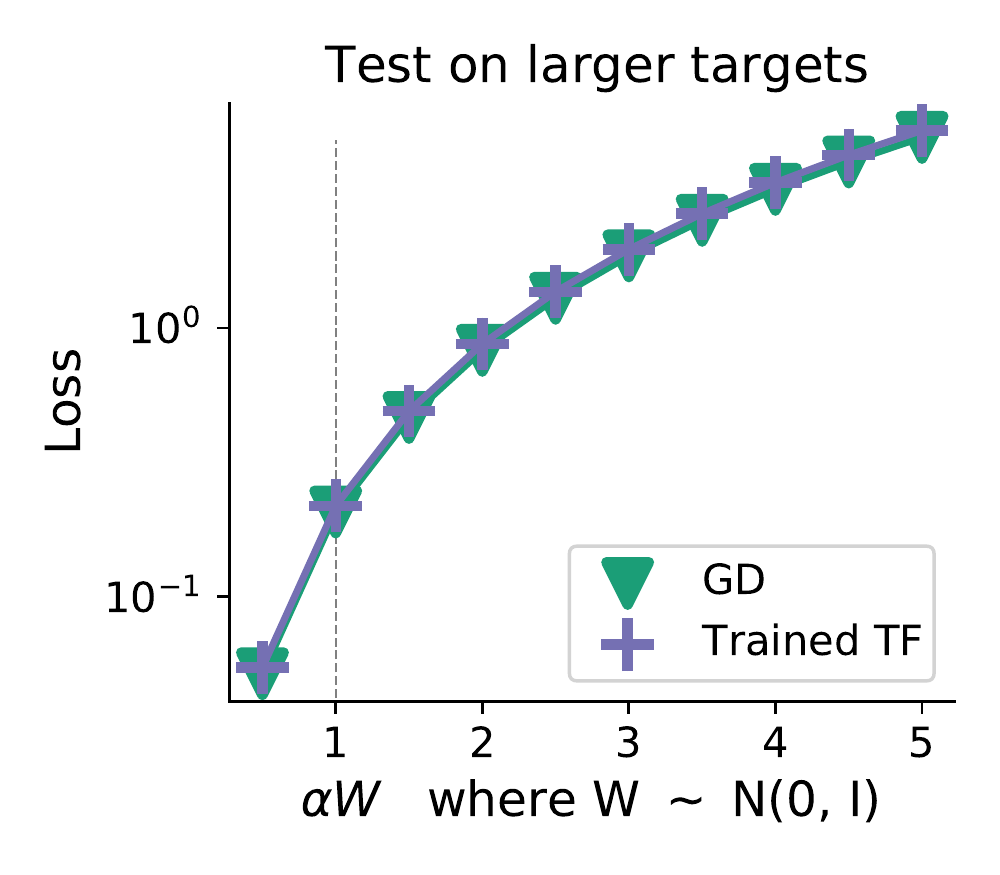}
  \end{center}
  \vspace{-10pt}
\end{minipage}
\end{center}
\vspace{-3pt}
  \caption{\textbf{Comparing trained two-headed and one-headed single-layer \textit{softmax} self-attention with 1 step of gradient descent on linear regression tasks.} \textit{Left column:} Softmax self-attention is not able to match gradient descent performance with hand-tuned learning rate, but adding a second attention head significantly reduces the gap, as expected by our analytical argument.  
  \textit{Center left}: The alignment suffers significantly for single-head softmax SA. We observe good but not as precise alignment when compared to linear Transformers for the two-headed softmax SA layer. \textit{Center right \& right:} The two-headed self-attention compared to the single-head layer shows similar robust out-of-distribution behavior compared to gradient descent.}
  \label{fig:softmax}
  \vspace{-10pt}
\end{figure*}

First, we show, see Figure \ref{fig:softmax}, that a single layer of softmax self-attention is not able to match GD performance. We tuned the learning rate as well as the weight initialization but found no significant difference over the hyperparameters we used througout this study. In general, we hypothesize that GD is an optimal update given the limited capacity of a single layer of (single-head) self-attention. We  therefore argue that the softmax induces (at best) a linear offset of the matrix product of training data and query vector
\begin{align}
    \text{softmax}(K^Tq_j) &= (e^{k_1^Tq_j}, \dots , e^{k_N^Tq_j})^T/ (\sum_i  e^{k_i^Tq_j}) \\
    &= (e^{x_1^TW_{KQ}x_j}, \dots , e^{x_N^TW_{KQ}x_j})^T/ (\sum_i  e^{x_i^TW_{KQ}x_j}) \\
    &\approx (1 + x_1^TW_{KQ}x_j, \dots , 1 +  x_N^TW_{KQ}x_j)^T/ (\sum_i  1 +  x_i^TW_{KQ}x_j) \\
    &\propto K^Tq_j  + \epsilon
\end{align}
proportional to a factor dependent on all $\{x_{\tau, i}\}_{i=1}^{N+1}$. We speculate that the dependency on the specific task $\tau$, for large $N_x$ vanishes or that the $x$-dependent value matrix could introduce a correcting effect. In this case the softmax operation introduces an additive error w.r.t.~to the optimal GD update. To overcome this disadvantageous offset, the Transformer can (approximately) introduce a correction with a second self-attention head by a simple subtraction i.e.
\begin{align}
 & P_1V_1\text{softmax}(K_1^TW_Qx_j) + P_2V_2\text{softmax}(K_2^TW_Qx_j) \\& \approx PV((1 + x_1^TW_{1,KQ}x_j, \dots , 1 +  x_N^TW_{1, KQ}x_j) -  (1 + x_1^TW_{2, KQ}x_j, \dots , 1 +  x_N^TW_{2, KQ}x_j)) \\ 
 &= PV(x_1^T(W_{1,KQ} - W_{2,KQ})x_j, \dots , x_N^T(W_{1,KQ} - W_{2,KQ})x_j) \\ & \propto PVK^Tq_j.
\end{align}

Here we assume that $PV$ 1) subsumes the dividing factor of the softmax and that 2) is the same (up to scaling) for each head. Note that if $(W_{1,KQ} - W_{2,KQ})$ is diagonal, and $P$ and $V$ chosen as in the Proposition of Appendix \ref{app:prop1}, we recover our gradient descent construction.

\begin{figure*}
\begin{center}
\begin{minipage}{.65\textwidth}
 \centering
  \begin{center}
    \includegraphics[width=1.\textwidth]{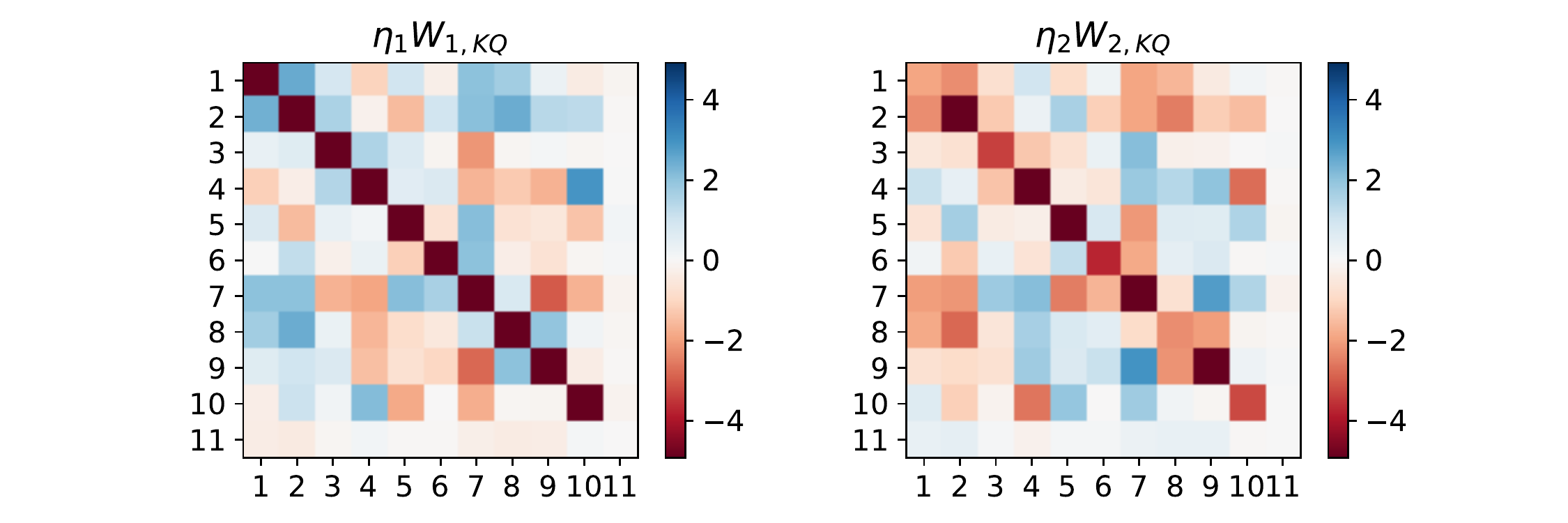}
  \end{center}
\end{minipage}
  \hspace{-30pt}
\begin{minipage}{.32\textwidth}
  \centering
  \begin{center}
    \includegraphics[width=.9\textwidth]{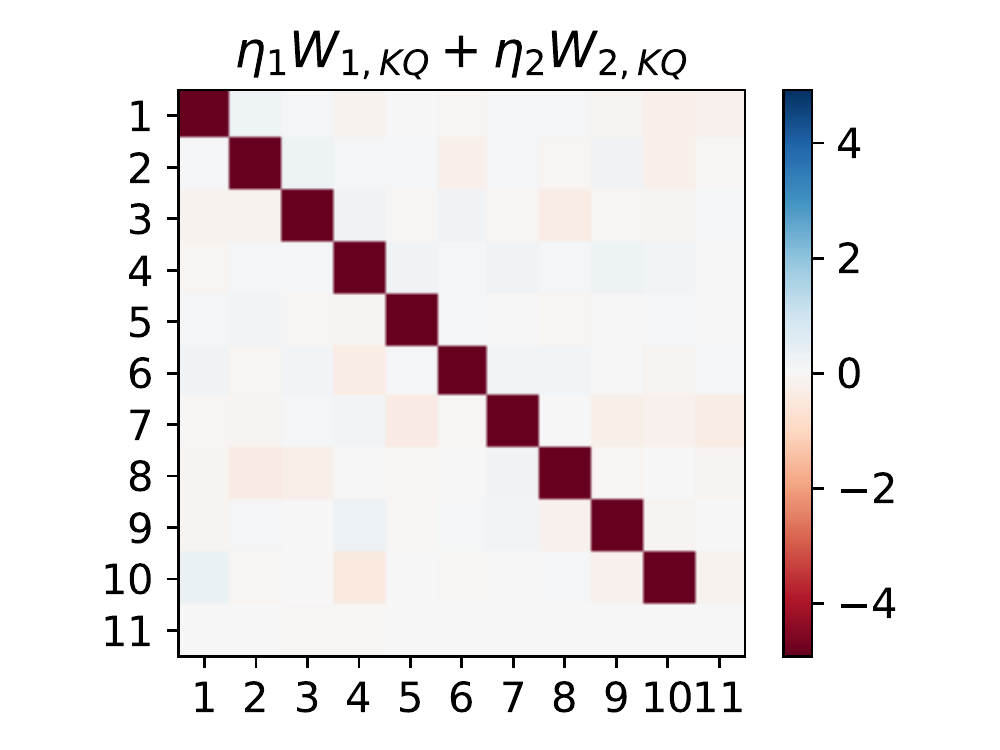}
  \end{center}
\end{minipage}
\end{center}
  \caption{\textbf{Visualizing the correction to the softmax operation when training Transformers on regression tasks.} The left and center plot show the matrix product $W_{KQ}=W_K^TW_Q$ including its scaling by $\eta$ induced through $PW_V$ of the two heads of the trained softmax self-attention layer. We observe that both of the matrices are approximate diagonal almost perfect sign reversed values on the off-diagonal terms. After adding the matrices (right plot), we observe a diagonal matrix and therefore to much improved approximation of our construction and therefore gradient descent dynamics.}
  \label{fig:softmax_weight_c}
  \vspace{-10pt}
\end{figure*}

We base this derivation on empirical findings, see Figure \ref{fig:softmax}, that, first of all, show the softmax self-attention performance increases drastically when using two heads instead of one. Nevertheless, the self-attention layer has difficulties to match the loss values of a model trained with GD. Furthermore, this architecture change leads to a very much improved alignment of the trained model and GD. Second, we can observe that when training a two-headed softmax self-attention layer on regression tasks the correction proposed above is actually observed in weight space, see Figure \ref{fig:softmax_weight_c}. Here, we visualize the matrix product within the softmax operation $W_{h, KQ}$ per head which we scale with the last diagonal entry of $P_hW_{h,V}$ which we denote by $\eta_h=P_hW_{h,V}(-1, -1)$. Intriguingly, this results in an almost perfect cancellation (right plot) of the off-diagonal terms and therefore in sum to an improved approximation of our construction, see the derivation above.


We would like to reiterate that the stronger inductive bias for copying data of the softmax layer remains, and is not invalidated by the analysis above. Therefore, even for our shallow and simple constructions they indeed fulfill an important role in support for our hypotheses: The ability to merge or copy input and target data into single tokens allowing for their dot product computation necessary for the construction in Proposition~\ref{prop:self_att_gd}, see Section~\ref{sect:softmax-builds-tokens} in the main text.

\begin{figure*}
\textbf{(a) Comparing one step of GD with a single-layer LSA Transformer with LayerNorm.}
\begin{center}
\begin{minipage}{.22\textwidth}
  \centering
  \begin{center}
    \includegraphics[width=1.\textwidth]{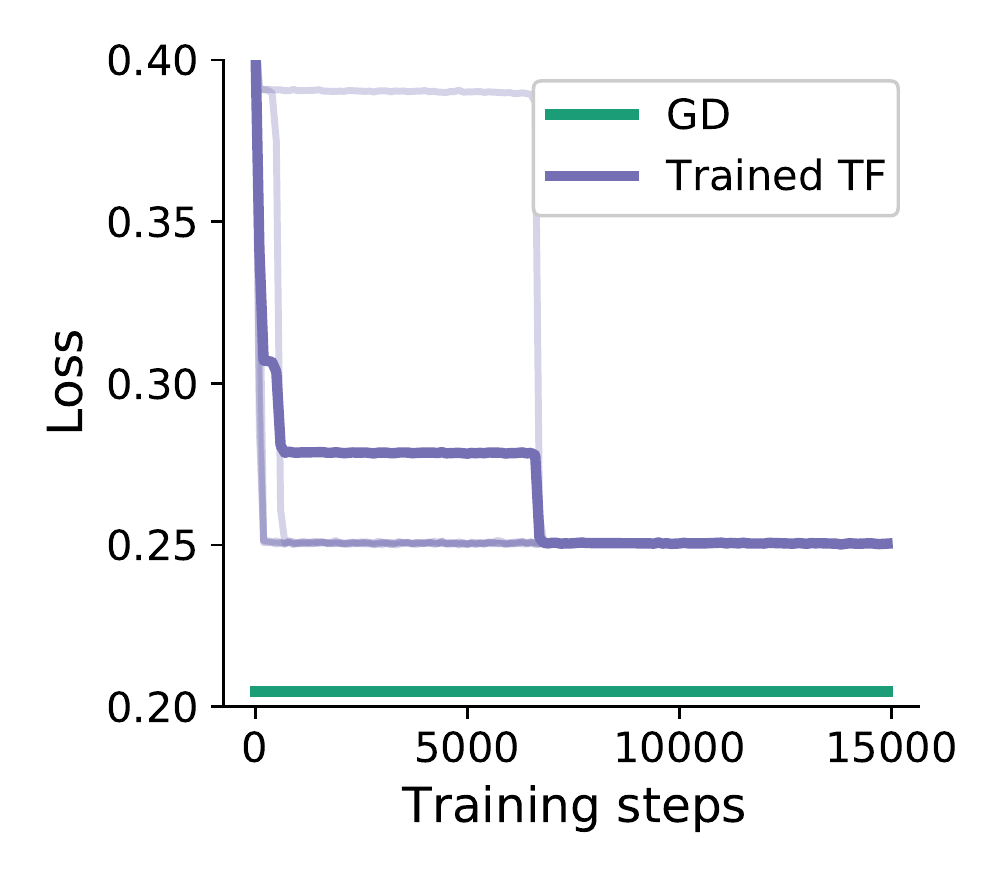}
  \end{center}
  \vspace{-10pt}
\end{minipage}
\begin{minipage}{.28\textwidth}
  \centering
  \begin{center}
    \includegraphics[width=1.\textwidth]{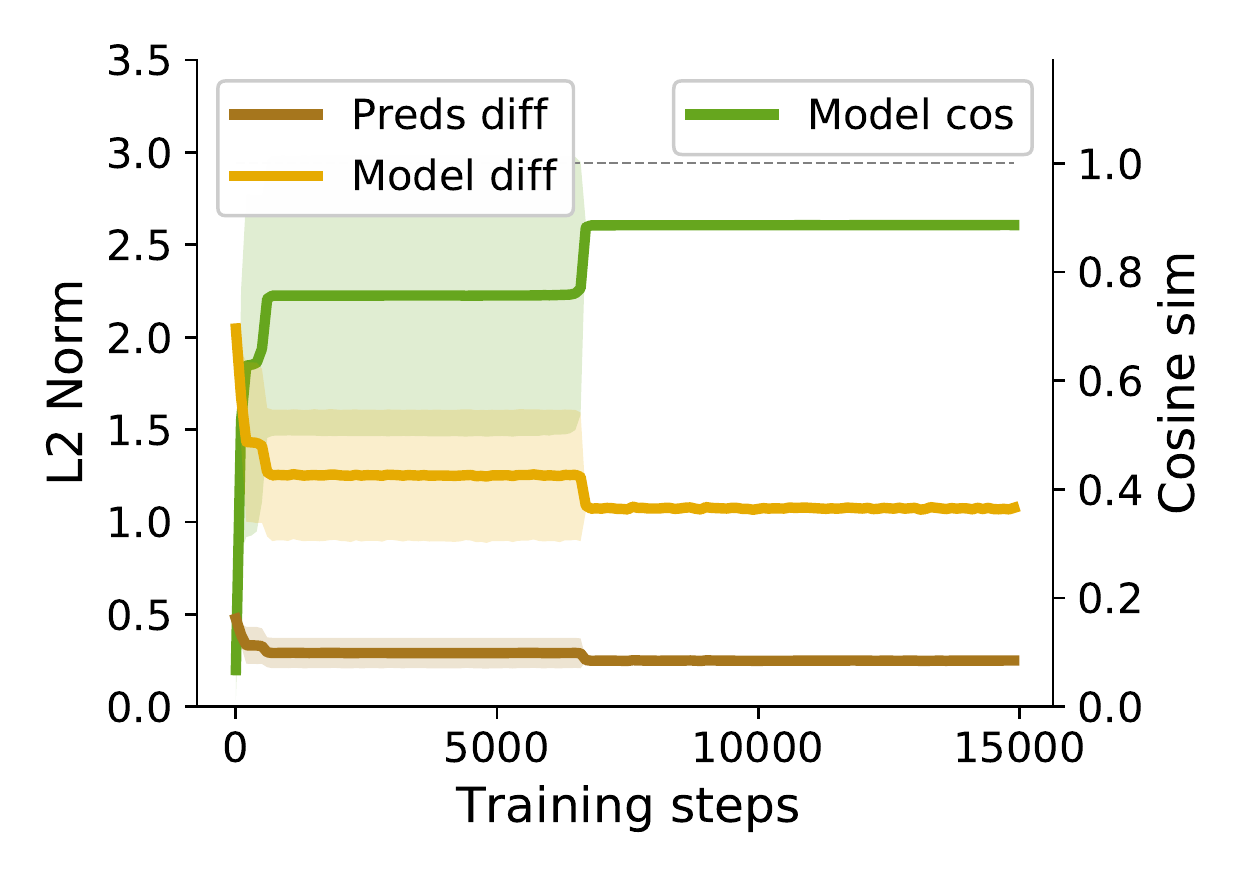}
  \end{center}
  \vspace{-10pt}
\end{minipage}
\begin{minipage}{.23\textwidth}
  \centering
  \begin{center}
    \includegraphics[width=1.\textwidth]{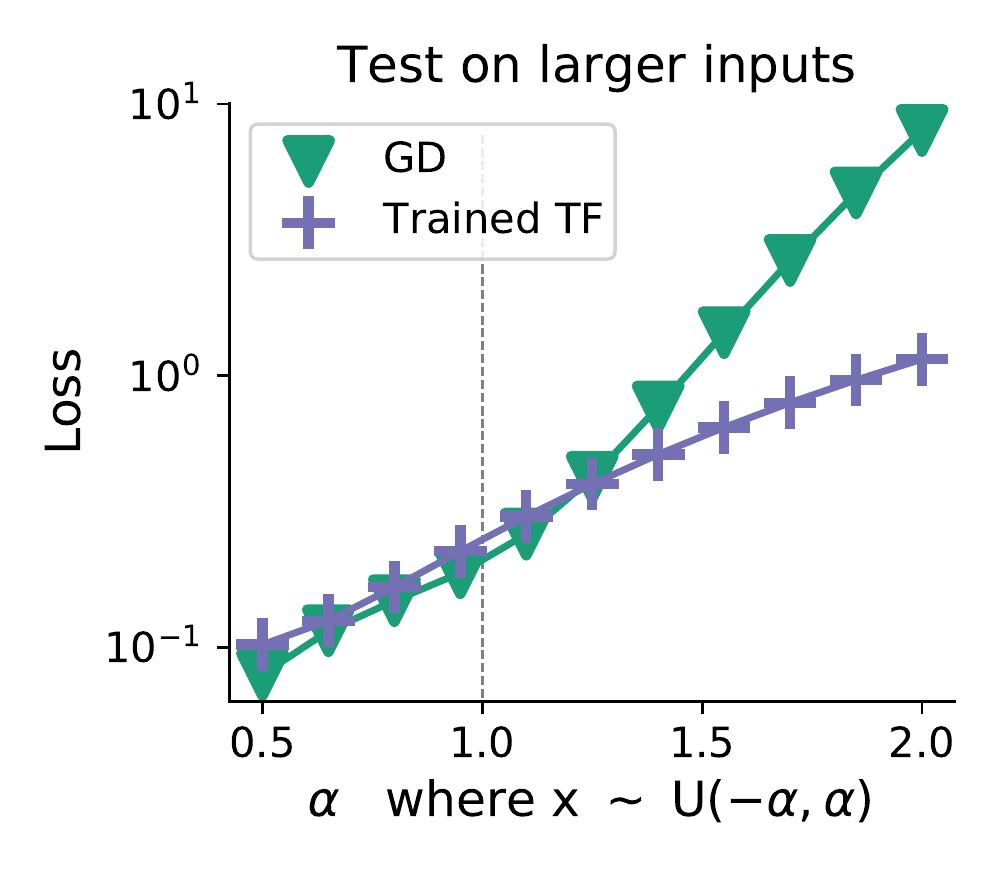}
  \end{center}
  \vspace{-10pt}
\end{minipage}
\begin{minipage}{.23\textwidth}
  \centering
  \begin{center}
    \includegraphics[width=1.\textwidth]{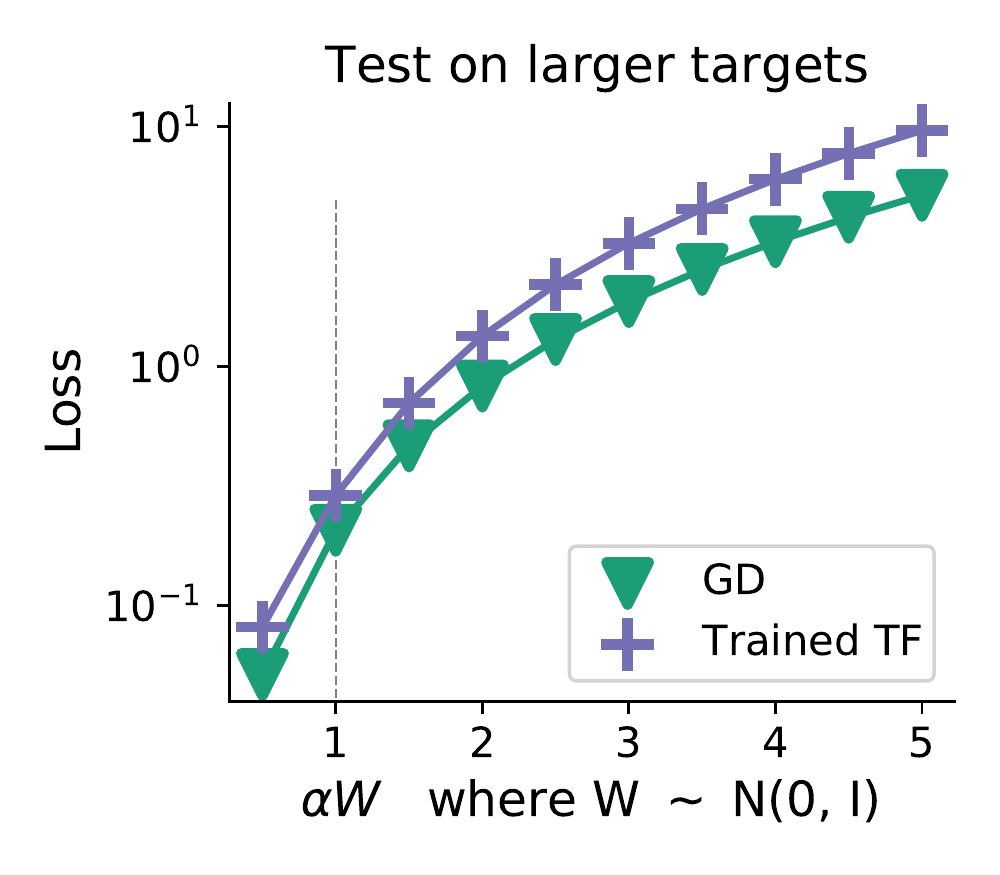}
  \end{center}
  \vspace{-10pt}
\end{minipage}
\end{center}
\textbf{(b) Comparing two steps of GD with a two-layer LSA Transformer with LayerNorm.}
\begin{center}
\begin{minipage}{.24\textwidth}
  \centering
  \begin{center}
    \includegraphics[width=1.\textwidth]{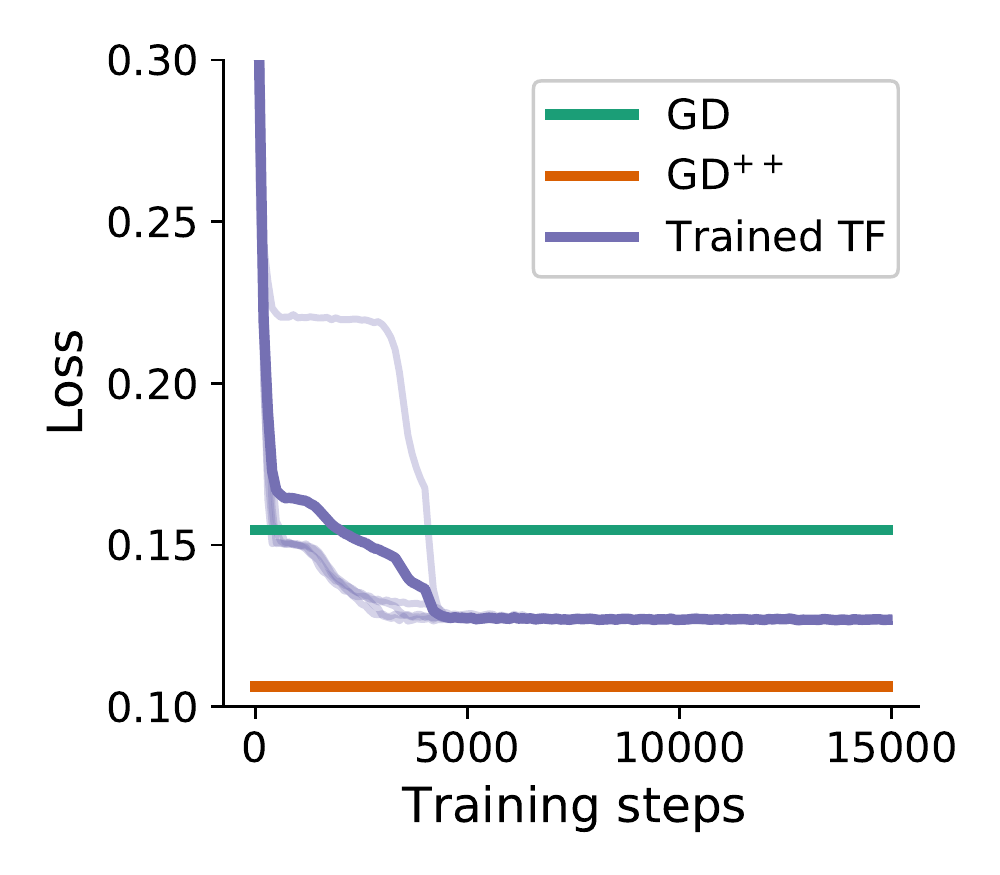}
  \end{center}
  \vspace{-10pt}
\end{minipage}
\begin{minipage}{.24\textwidth}
  \centering
  \begin{center}
    \includegraphics[width=1.\textwidth]{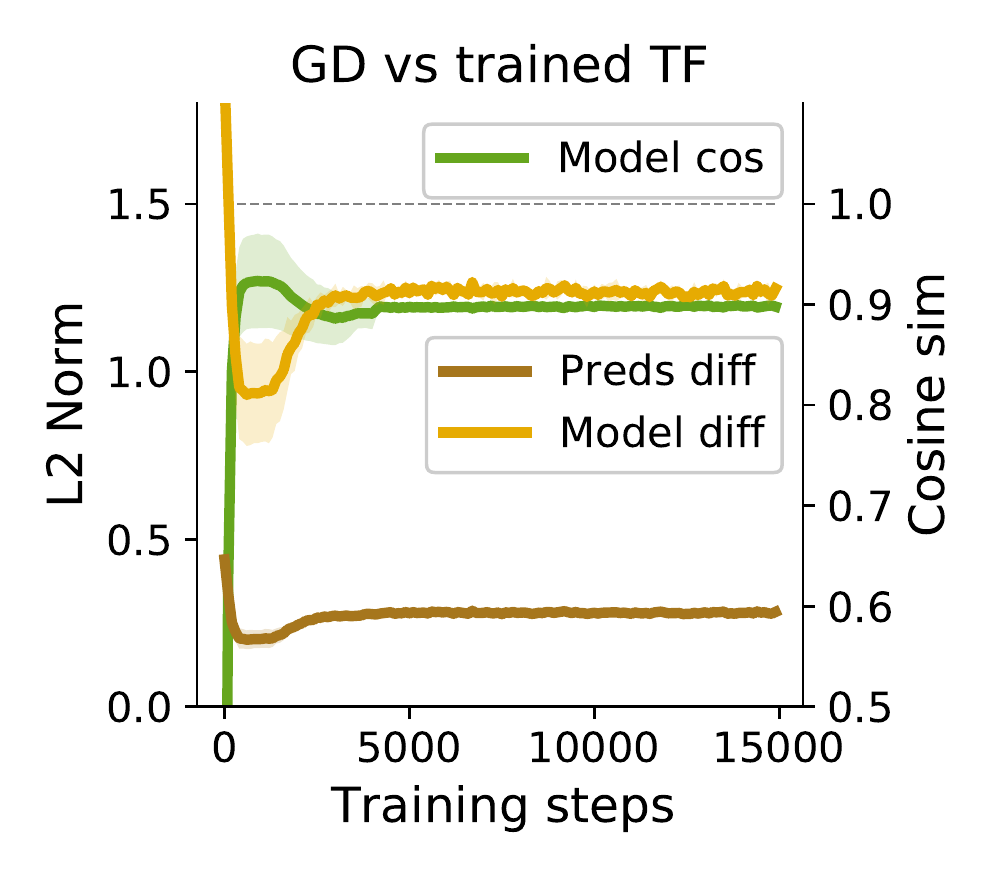}
  \end{center}
  \vspace{-10pt}
\end{minipage}
\begin{minipage}{.24\textwidth}
  \centering
  \begin{center}
    \includegraphics[width=1.\textwidth]{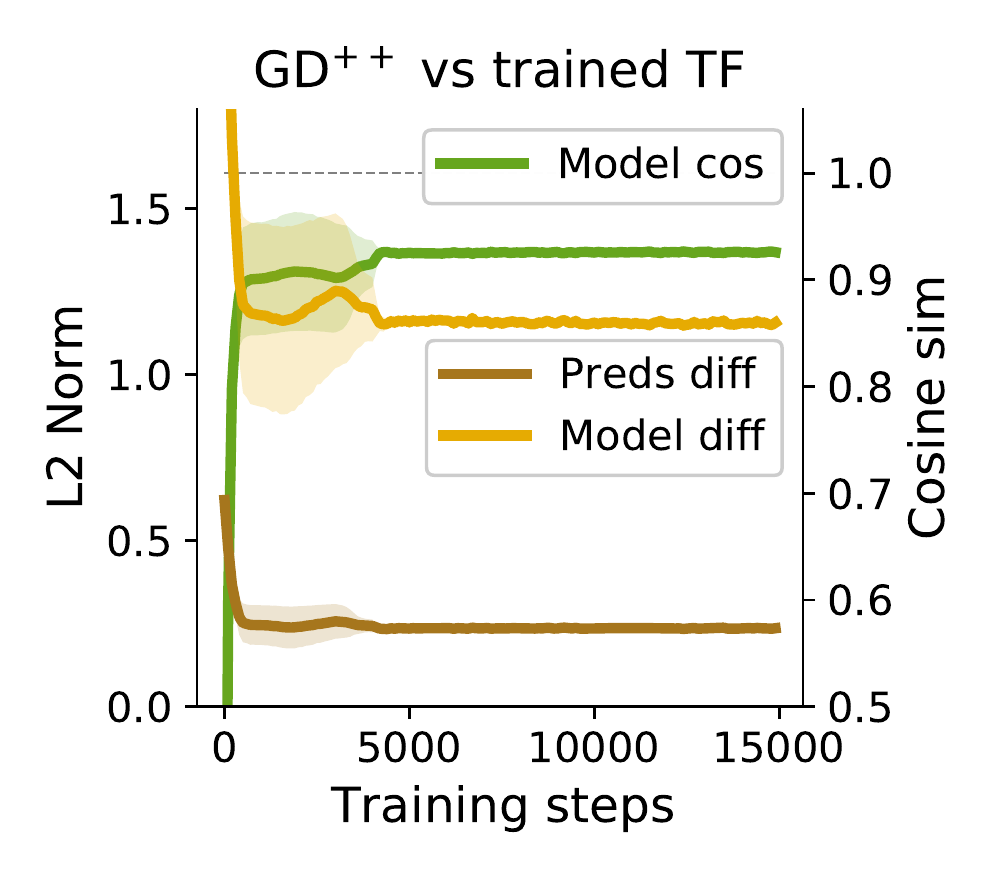}
  \end{center}
  \vspace{-10pt}
\end{minipage}
\begin{minipage}{.24\textwidth}
  \centering
  \begin{center}
    \includegraphics[width=1.\textwidth]{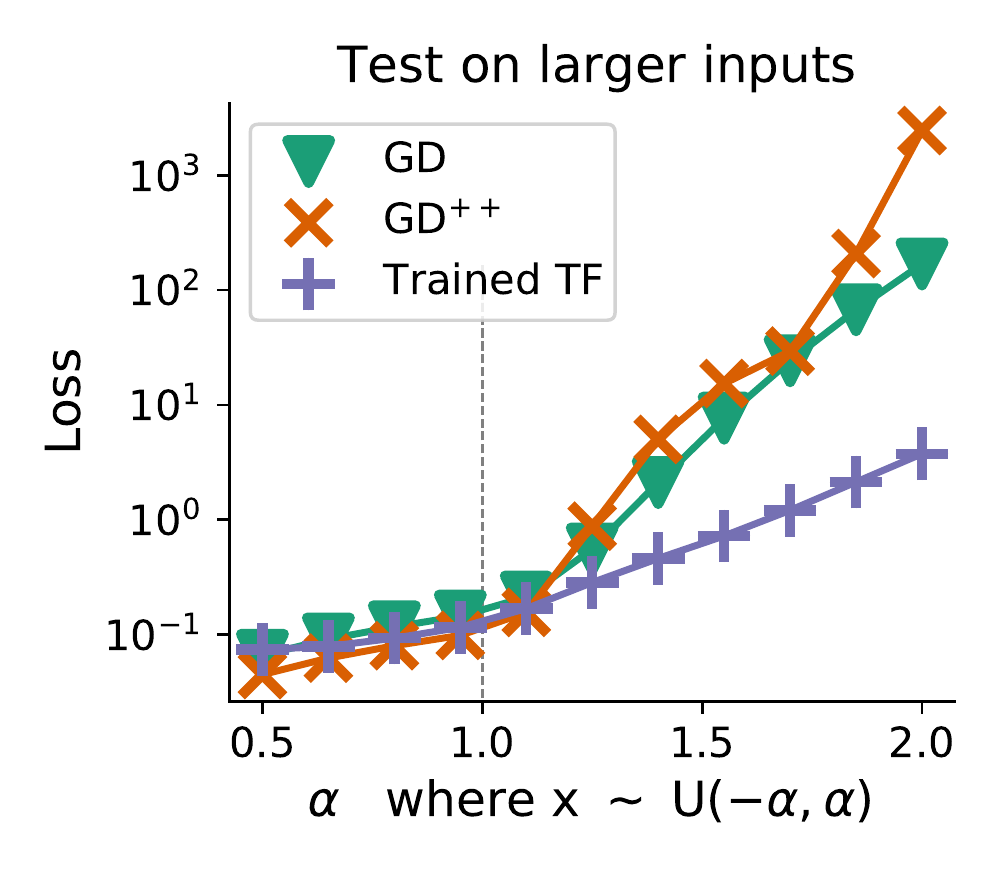}
  \end{center}
  \vspace{-10pt}
\end{minipage}
\end{center}
\vspace{-3pt}
  \caption{\textbf{Comparing trained 1-layer and 2-layer Transformers with  \textit{LayerNorm} and 1 step or 2 steps of gradient descent resp.} \textit{Left column:} The Transformers is not able to match the gradient descent performance with hand-tuned learning rate.  
  \textit{Alignment plots}: The alignment suffers significantly when comparing to linear self-attention layers although still reasonable alignment is obtained which decreases slightly when comparing to GD$^{++}$ for the two-layer Transformer.\textit{Center right \& right:} The LayerNorm Transformer outperforms when GD when providing training input data that is significantly larger than the data provided during training.}
  \label{fig:layernorm}
  \vspace{-10pt}
\end{figure*}

We end this section by analysing Transformers equipped with LayerNorm which we apply as usually done before the self-attention layer: Overall, we observe qualitatively similar results to Transformers with softmax self-attention layer i.e. a decrease in performance compared to GD accompanied with a decrease in alignment between models generated by the Transformer and models trained with GD, see Figure \ref{fig:layernorm}.
Here, we test again a single linear self-attention layer succeeding LayerNorm as well as two layers where we skip the first LayerNorm and only include a LayerNorm between the two. Including more heads does not help substantially. 
We again assume the optimality of GD and argue that information of  targets and inputs present in the tokens is lost by averaging when applying LayerNorm. This naturally leads to decreasing performance compared to GD, see first row of Figure \ref{fig:layernorm}. Although the alignment to GD and GD$
^{++}$, especially for two layers, is high, we overall see inferior performance to one or two steps of GD or two steps of GD$^{++}$.
Nevertheless, we speculate that LayerNorm might not only stabilize Transformer training but could also act as some form of data normalization procedure that implicitly enables better generalization for larger inputs as well as targets provided in-context, see OOD experiments in Figure \ref{fig:layernorm}.

Overall we conclude that common architecture choices like softmax and LayerNorm seem supoptimal for the constructed in-context learning settings when comparing to GD or linear self-attention. Nevertheless, we speculate that the potentially small performance drops of in-context learning are negligible when turning to deep and wide Transformers for which these architecture choices have empirically proven to be superior. 

\subsection{Details of curvature correction}
\label{app:curvature_details}

We give here a precise construction showing how to implement in a single head, a step of GD and the discussed data transformation, resulting in GD$^{++}$. Recall again the linear self-attention operation with a single head
\begin{align}
e_j  \leftarrow & e_j + P W_{V}\sum_{i} e_{i} \otimes e_{i} W_{K}^T.
\end{align}
We provide again the weight matrices in block form of the construction of Prop. \ref{prop:self_att_gd} but now enabling additionally our described data transformation: $W_{K} = W_{Q} = \left(\begin{array}{@{}c c@{}}
  I_x
  & 0 \\
  0 &
  0
\end{array}\right)
$ with $I_x$ the identity matrix of size $N_x$, $I_y$ od size $N_y$ resp. Furthermore, we set $W_{V} = \left(\begin{array}{@{}c c@{}}
  I_x
  & 0 \\
  W &
  -I_y
\end{array}\right)$ with the weight matrix $W \in \mathbb{R}^{N_y \times N_x}$ of the linear model we wish to train and 
$P = \left(\begin{array}{@{}c c@{}}
  -\gamma I_x
  & 0 \\
  0 &
  \frac{\eta}{N}
\end{array}\right)$. This leads to the following update

\begin{align}
\left(\begin{array}{@{}c@{}}
  x_j\\
  y_j 
\end{array}\right)  \leftarrow & \left(\begin{array}{@{}c@{}}
  x_j\\
  y_j 
\end{array}\right)
+ \left(\begin{array}{@{}c c@{}}
  - \gamma I_x
  & 0 \\
  0 &
  \frac{\eta}{N}
\end{array}\right) \sum_{i=1}^N
\left(\left(\begin{array}{@{}c c@{}}
  I_x
  & 0 \\
  W &
  -I_y
\end{array}\right)
\left(\begin{array}{@{}c@{}}
  x_i\\
  y_i 
\end{array}\right)\right)
 \otimes\left(
\left(\begin{array}{@{}c c@{}}
  I_x
  & 0 \\
  0 &
  0
\end{array}\right)\left(\begin{array}{@{}c@{}}
  x_i\\
  y_i 
\end{array}\right)\right)
\left(\begin{array}{@{}c c@{}}
  I_x 
  & 0 \\
  0 &
  0
\end{array}\right)
\left(\begin{array}{@{}c@{}}
  x_j\\
  y_j 
\end{array}\right)
 \nonumber\\
&= \left(\begin{array}{@{}c@{}}
  x_j\\
  y_j 
\end{array}\right)
+ \left(\begin{array}{@{}c c@{}}
  - \gamma I_x
  & 0 \\
  0 &
  \frac{\eta}{N}
\end{array}\right) \sum_{i=1}^N
\left(\begin{array}{@{}c@{}}
  x_i\\
Wx_i - y_i 
\end{array}\right)
 \otimes\left(\begin{array}{@{}c@{}}
  x_i\\
  0 
\end{array}\right)
\left(\begin{array}{@{}c@{}}
  x_j\\
  0
\end{array}\right) = \left(\begin{array}{@{}c@{}}
  x_j\\
  y_j 
\end{array}\right) + \left(\begin{array}{@{}c@{}}
  - \gamma XX^T x_j\\
  -\Delta W x_j 
\end{array}\right).
\end{align}
for every token $e_j = (x_j, y_j)$ including the query token $e_{N+1} = e_{\text{test}} = (x_{\text{test}}, 0) $ which will give us the desired result.

\begin{figure*}
\begin{center}
\begin{minipage}{.32\textwidth}
 \centering
  \begin{center}
    \includegraphics[width=.9\textwidth]{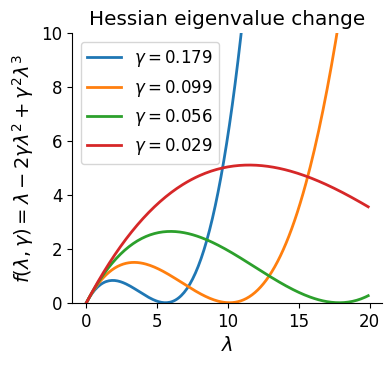}
  \end{center}
\end{minipage}
\begin{minipage}{.32\textwidth}
  \centering
  \begin{center}
    \includegraphics[width=.9\textwidth]{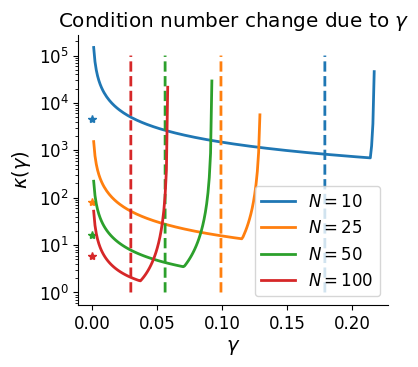}
  \end{center}
\end{minipage}
\begin{minipage}{.3\textwidth}
  \centering
  \begin{center}
    \includegraphics[width=.9\textwidth]{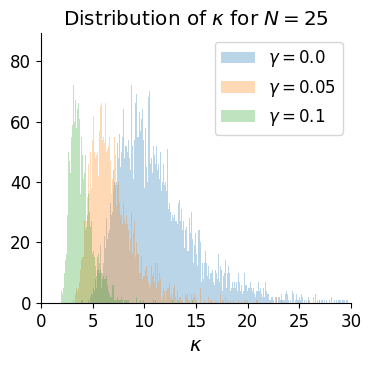}
  \end{center}
\end{minipage}
\end{center}
  \caption{\textbf{GD$^{++}$ analyses}. \textit{Left:} We visualize the change of the eigenspectrum induced by the input data transformation of GD$^{++}$ for different $\gamma$ observed in practice. \textit{Center:} Given we know the maximum and minimum of eigenvalues $\lambda_1, \lambda_n$ of the loss Hessian $XX^T$ with $X=(x_0, \dots, x_N)$ for different $N$, we compare the original condition number (depicted by *'s at $\gamma=0$) and the condition number (in log scale) of the GD$^{++}$ altered loss Hessian when varying $\gamma$. We plot in dotted lines the $\gamma$ values that we observe in practice which are close the optimal ones i.e. the local minimum derived through our analysis.
  \textit{Right:} For$N=25$, we plot for different $\gamma$ values the distribution of condition numbers $\kappa = \lambda_1/\lambda_n$ for 10000 tasks and observe favorable $\kappa$ values close to 1 when approaching the $\gamma=0.099$ value was found in practice. The $\kappa$ values quickly explode for $\gamma > 0.1$.}
  \label{fig:gd_pp}
  \vspace{-10pt}
\end{figure*}

\textbf{Why does GD$^{++}$ perform better?} We give here one possible explanation of the superior performance of GD$^{++}$ compared to GD. Note that there is a close resemblance of the GD transformation and a heavily truncated Neuman series approximation of the inverse $XX^T$. We provide here a more heuristic explanation for the observed acceleration. 

Given $\gamma \in \mathbb{R}$, GD$^{++}$ transforms every input according to $x_i \leftarrow x_i-\gamma XX^Tx_i = (I-\gamma XX^T)x_i$. We can therefore look at the change of squared regression loss $L(W) = \frac{1}{2}\sum_{i=0}^N (Wx_i -y_i)^2$ induced by this transformation i.e. $L^{++}(W) = \frac{1}{2}\sum_{i=0}^N (W(I-\gamma XX^T)x_i -y_i)^2 = \frac{1}{2}(W(I-\gamma XX^T)X-Y)^2$ which in turn leads to a change of the loss Hessian from $\nabla^2 L = XX^T$ to 
$\nabla^2 L^{++} = (I-\gamma XX^T)X((I-\gamma XX^T)X)^T$.

Given the original Hessian $H = XX^T = U\Sigma U^T$ with it's set of sorted eigenvalues $\{\lambda_1, \dots , \lambda_n\}$ and $\lambda_i \geq 0$ on the diagonal matrix $\Sigma$ we can express the new Hessian through $U, \Sigma$ i.e. $H^{++} = (I-\gamma XX^T)X((I-\gamma XX^T)X)^T = (I-\gamma U\Sigma U^T)U\Sigma U^T(I-\gamma U\Sigma U^T)^T$. 

We can simplify $H^{++}$ further as
\begin{align}
    H^{++} &= (I-\gamma U\Sigma U^T)U\Sigma U^T(I-\gamma U\Sigma U^T)^T = U (\Sigma - \gamma \Sigma^2) U^T U (I - \gamma \Sigma) U^T \\ &= U (\Sigma - 2 \gamma \Sigma^2 + \gamma^2 \Sigma^3) U^T
\end{align}

Given the eigenspectrum $\{\lambda_1, \dots , \lambda_n\}$ of $H$, we obtain an (unsorted) eigenspecturm for $H^{++}$ with  $\{\lambda_1 - 2\gamma \lambda_1^2 + \gamma^2 \lambda_1^3, \dots , \lambda_n - 2\gamma \lambda_n^2 + \gamma^2 \lambda_n^3\}$ which we visualize in Figure  \ref{fig:gd_pp} for different $\gamma$ observed in practice. We hypotheses that the Transformer chooses $\gamma$ in a way that on average, across the distribution of tasks, the data transformation (iteratively) decreases the condition number $\lambda_1/\lambda_n$ leading to accelerated learning. This could be achieved, for example, by keeping the smallest eigenvalue $\lambda_n \approx \lambda_n^{++}$ fixed and choosing $\gamma$ such that the largest eigenvalue of the transformed data $\lambda_1^{++}$ is reduced, while the original $\lambda_1$ stays within $[\lambda_1^{++}, \lambda_n^{++}]$.

To support our hypotheses empirically, we computed the minimum and maximum eigenvalues of $XX^T$ across $10000$ tasks while changing the number of datapoints $N \in [10, 25, 50, 100]$ i.e. $X=(x_0, \dots, x_N)$ leading to better conditioned loss Hessians i.e. $[1\mathrm{e}{-10}, 0.097, 0.666, 2.870]$ and $[4.6, 7.712, 10.845, 17.196]$ as the minimum and maximum eigenvalues of $XX^T$ across all tasks where we cut the smallest eigenvalue for $N=10$ at $1\mathrm{e}{-10}$.
Furthermore, we extract the $\gamma$ values from the weights of optimized recurrent 2-layer Transformers trained on different task distributions and obtain $\gamma$ values of $[0.179, 0.099, 0.056, 0.029]$, see again Figure \ref{fig:gd_pp}. Note that the observed eigenvalues stay within $[0, 1/\gamma]$ i.e. the two roots of $f(\lambda, \gamma) = \lambda - 2\gamma \lambda^2 + \gamma^2 \lambda^3$.

Given the derived function of eigenvalue change $f(\lambda, \gamma)$, we compute the condition number of $H^{++}$ by dividing the novel maximum eigenvalues $\lambda_1^{++} = f(1/(3\gamma), \gamma)$ where $\lambda = 1/(3\gamma)$ as the local maximum of $f(\lambda, \gamma)$, for fixed $\gamma$, and the novel minimum eigenvalue $\lambda_n^{++} = \min (f(\lambda_1, \gamma),f(\lambda_n, \gamma))$. Note that with too small $\gamma$, we move the original $\lambda_n$ closer to the root of $f(\lambda, \gamma)$ i.e. $\lambda = 1/\gamma$  and therefore can change the smallest eigenvalue. 

Given the task distribution and its corresponding eigenvalue distribution, we see that choosing $\gamma$ reduces the new condition number $\kappa^{++} = \lambda_1^{++}/\lambda_n^{++}$ which leads to better conditioned learning, see center plot of Figure \ref{fig:gd_pp}. Note that the optimal $\gamma$ based on our derivation above is based on the maximum and minimum eigenvalue across all tasks and does not take the change of the eigenvalue distribution into account. We argue therefore that the simplicity of the arguments above does not capture the task statistics and distribution shifts entirely and therefore obtains a slightly larger $\gamma$ as an optimal value. We furthermore visualize the condition number change for $N=25$ and 10000 tasks in the right plot of Figure \ref{fig:gd_pp} and observe the distribution moving to desirable $\kappa$ values close to $1$. For $\gamma$ values larger than 0.1 the distribution quickly exhibits exploding condition numbers.

\subsection{Phase transitions}
\label{app:phase_transitions}

\begin{figure*}
\begin{center}
\begin{minipage}{.32\textwidth}
 \centering
  \begin{center}
    \includegraphics[width=.9\textwidth]{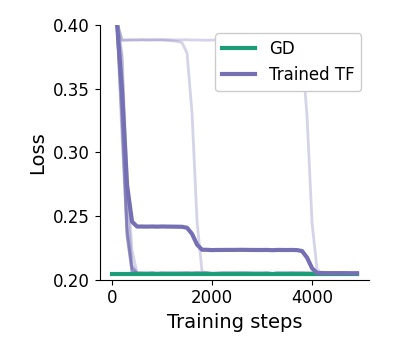}
  \end{center}
\end{minipage}
\begin{minipage}{.32\textwidth}
  \centering
  \begin{center}
    \includegraphics[width=.9\textwidth]{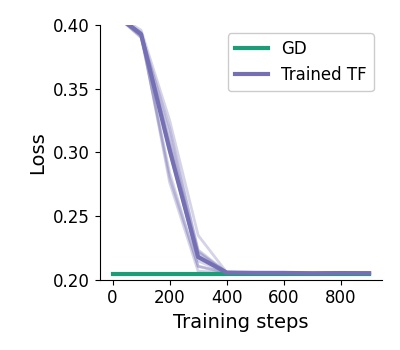}
  \end{center}
\end{minipage}
\begin{minipage}{.3\textwidth}
  \centering
  \begin{center}
    \includegraphics[width=.9\textwidth]{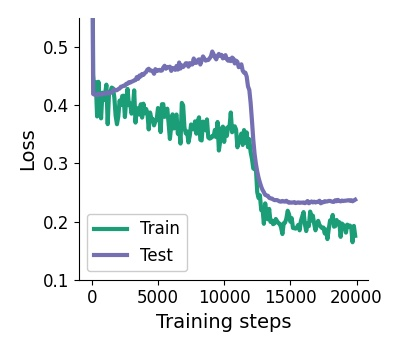}
  \end{center}
\end{minipage}
\end{center}
  \caption{\textbf{Phase transitions during training}. \textit{Left:} Loss based on 10 different random seeds when optimizing a single-headed self-attention layer. We observe for some seeds very long initial phases of virtually zero progress after which the loss drops suddenly to the desired GD loss. \textit{Center:} The same experiment but optimizing a \textit{two}-headed self-attention layer. We observe fast and robust convergence to the loss of GD.
  \textit{Right:} Training a single Transformer block i.e. a self-attention layer with MLP and a reduced training set size of 8192 tasks. We observe grokking like train and test loss phase transitions where test set first increases drastically before experiencing a sudden drop in loss almost matching the desired GD loss of $~0.2$.  }
  \label{fig:grokking}
  \vspace{-10pt}
\end{figure*}

We comment shorty on the curiously looking phase transitions of the training loss observed in many of our experiments, see Figure \ref{fig:training_trans_lin}. Nevertheless, simply switching from a single-headed self-attention layer to a two-headed self-attention layer mitigates the random seed dependent training instabilities in our experiments presented in the main text, see left and center plot of Figure \ref{fig:grokking}.

Furthermore, these transitions look reminiscent of the recently observed "grokking" behaviour \cite{grocking}. Interestingly, when carefully tuning the learning rate and batchsize we can also make the Transformers trained in these linear regression tasks \textit{grokk}. For this, we train a single Transformer block (self-attention layer and MLP) on a limited amount of data (8192 tasks), see right plot of Figure \ref{fig:grokking}, and observe grokking like train and test loss phase transitions where test set first increases drastically before experiencing a sudden drop in loss almost matching the desired GD loss of $~0.2$. We leave a thorough investigation of these phenomena for future study. 

\subsection{Experimental details}

We use for most experiments identical hyperparameters that were tuned by hand which we list here

\label{app:ex_det}
\begin{itemize}
    \item Optimizer: Adam \citep{adam} with default parameters and learning rate of 0.001 for Transformer with depth $K<3$ and 0.0005 otherwise. We use a batchsize of 2048 and applied gradient clipping to obtain gradients with global norm of $10$. We used the Optax library. 
    \item Haiku weight initialisation (fan-in) with truncated normal and std $0.002/K$ where $K$ the number of layers.
    \item We did not use any regularisation and observed for deeper Transformers with $K>2$ instabilities when reaching GD performance. We speculate that this occurs since the GD performance is, for the given training tasks, already close to divergence as seen when providing tasks with larger input ranges. 
    Therefore, training Transformers also becomes instable when we approach GD with an optimal learning rate. In order to stabilize training, we simply clipped the token values to be in the range of $[-10, 10]$.
    \item  When applicable we use standard positional encodings of size $20$ which we concatenated to all tokens.
    \item  For simplicity, and to follow the provided weight construction closely, we did use square key, value and query parameter matrix in all experiments.
    \item The training length varied throughout our experimental setups and can be read off our training plots in the article. 
    \item When training meta-parameters for gradient descent i.e. $\eta$ and $\gamma$ we used an identical training setup but usually training required much less iterations.
    \item In all experiments we choose inital $W_0=0$ for gradient descent trained models.
\end{itemize}

\begin{figure*}
\textbf{(a) Comparing 1 step of gradient descent with training a LSA-layer on 128 tasks.}
\begin{center}
\begin{minipage}{.24\textwidth}
  \centering
  \begin{center}
    \includegraphics[width=1.\textwidth]{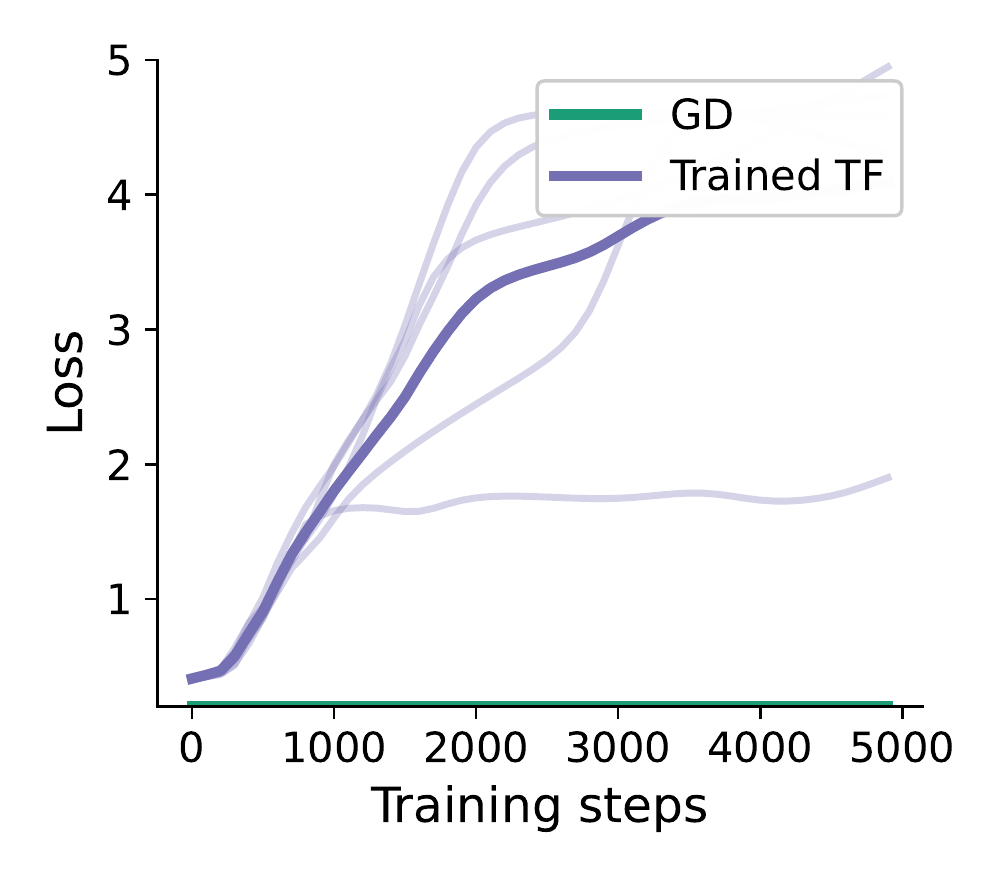}
  \end{center}
  \vspace{-10pt}
\end{minipage}
\begin{minipage}{.24\textwidth}
  \centering
  \begin{center}
    \includegraphics[width=1.\textwidth]{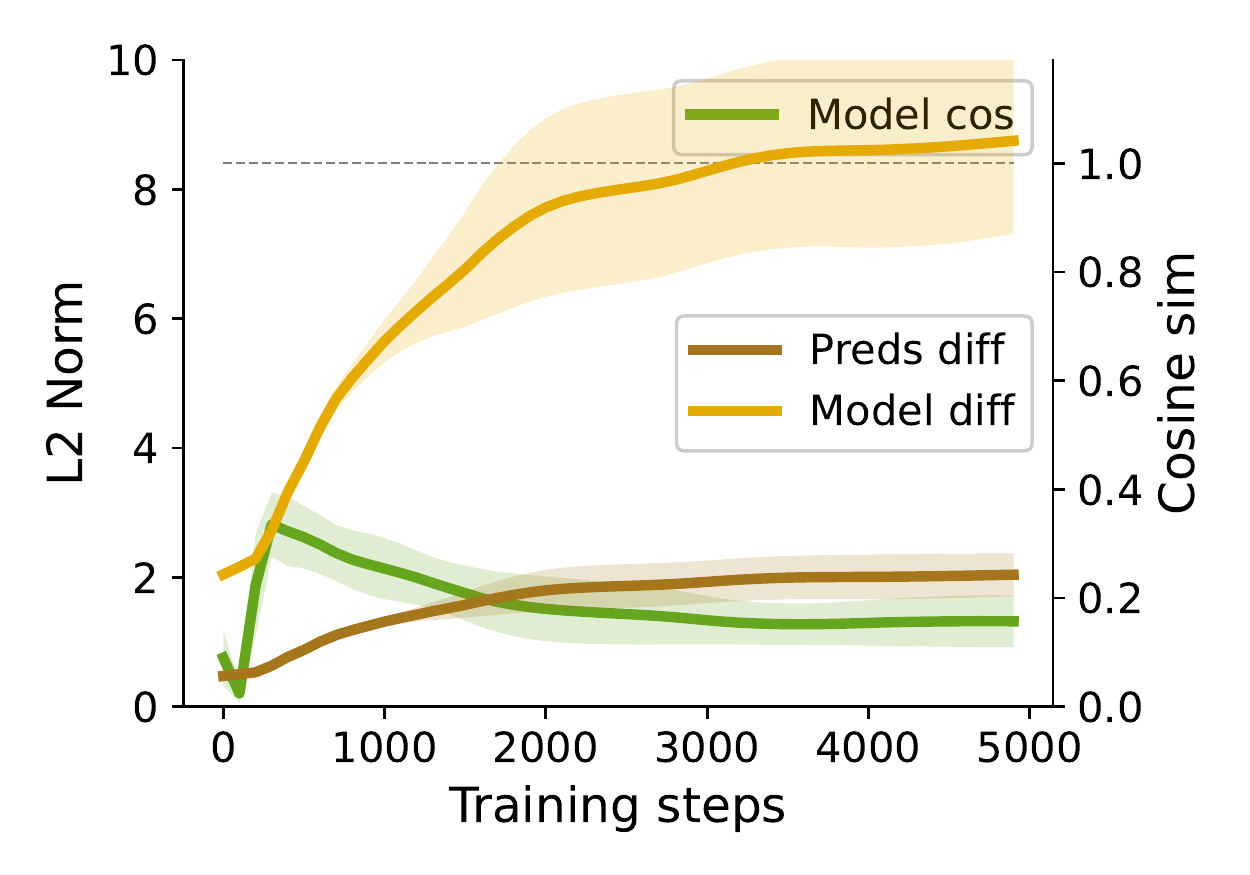}
  \end{center}
  \vspace{-10pt}
\end{minipage}
\begin{minipage}{.24\textwidth}
  \centering
  \begin{center}
    \includegraphics[width=1.\textwidth]{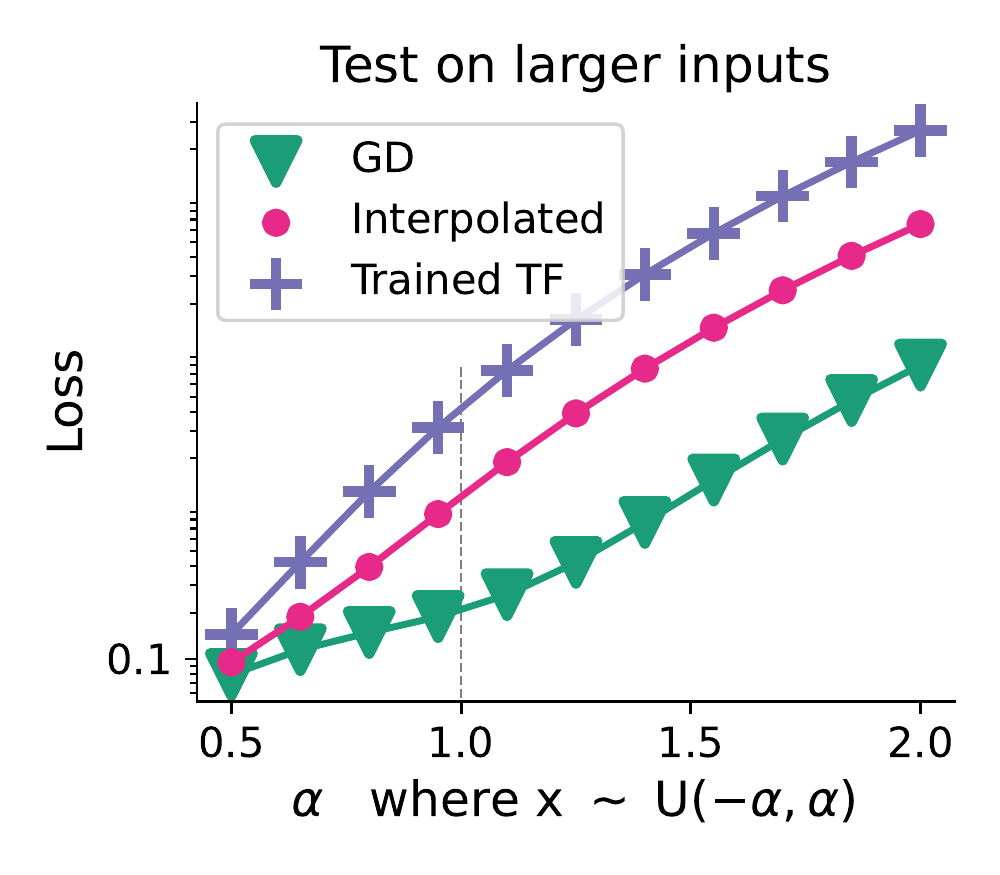}
  \end{center}
  \vspace{-10pt}
\end{minipage}
\begin{minipage}{.24\textwidth}
  \centering
  \begin{center}
    \includegraphics[width=1.\textwidth]{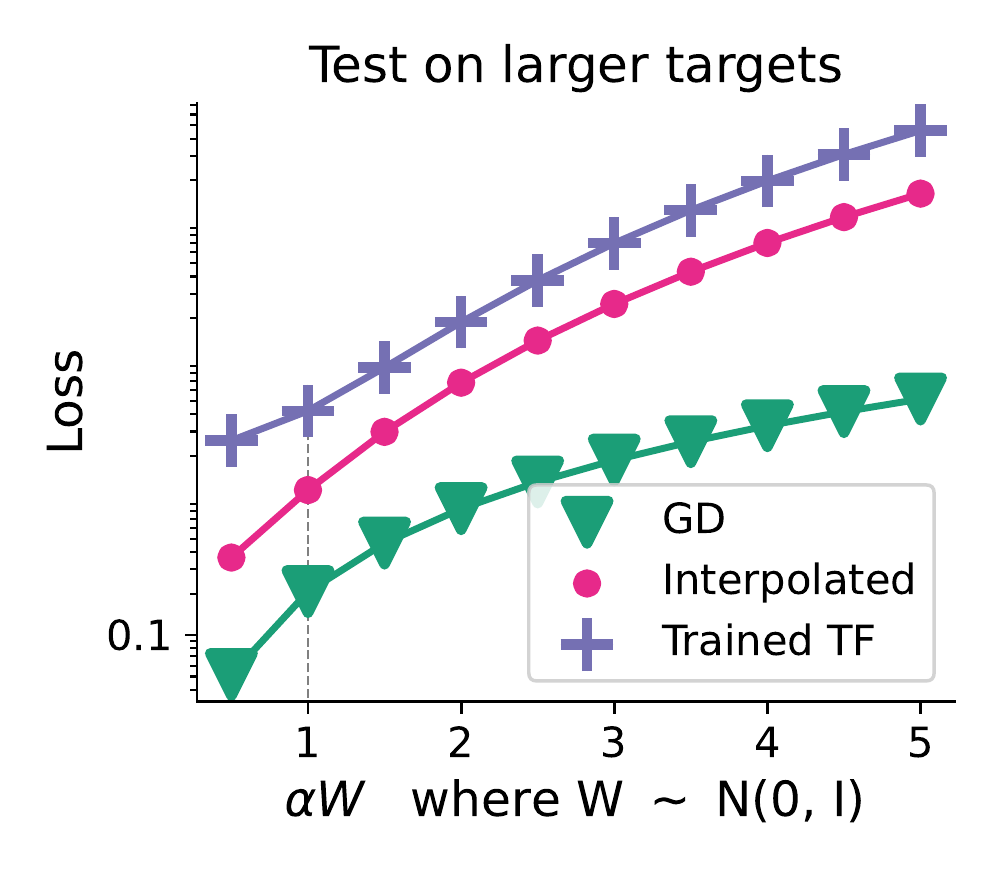}
  \end{center}
  \vspace{-10pt}
\end{minipage}
\end{center}

\textbf{(b) Comparing 1 step of gradient descent with training a LSA-layer on 512 tasks.}

\begin{center}
\begin{minipage}{.24\textwidth}
  \centering
  \begin{center}
    \includegraphics[width=1.\textwidth]{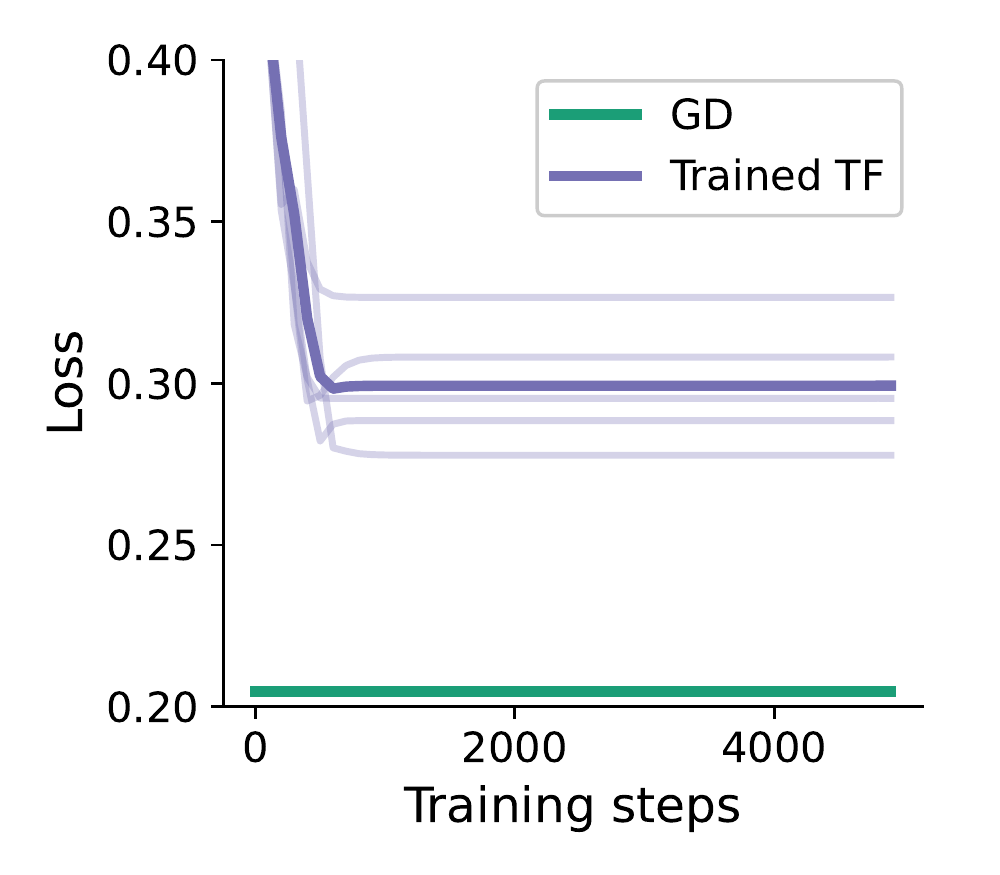}
  \end{center}
  \vspace{-10pt}
\end{minipage}
\begin{minipage}{.24\textwidth}
  \centering
  \begin{center}
    \includegraphics[width=1.\textwidth]{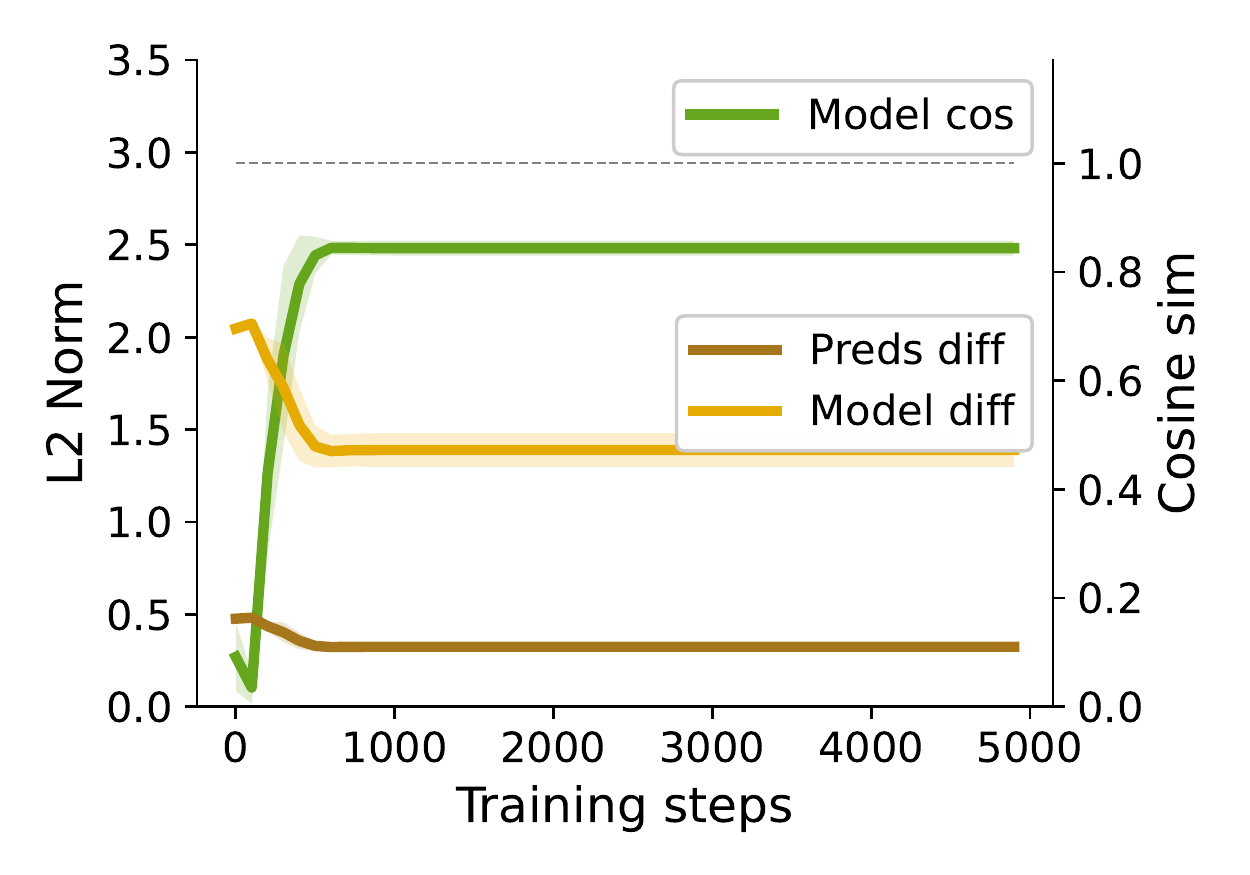}
  \end{center}
  \vspace{-10pt}
\end{minipage}
\begin{minipage}{.24\textwidth}
  \centering
  \begin{center}
    \includegraphics[width=1.\textwidth]{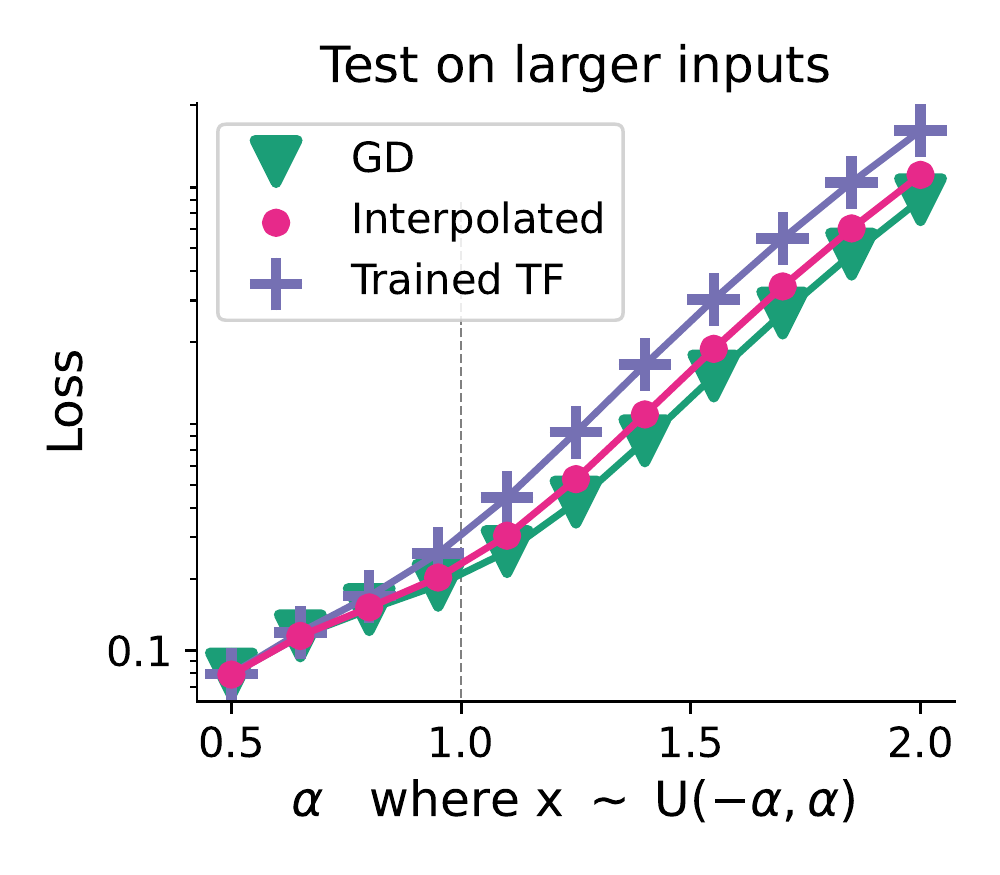}
  \end{center}
  \vspace{-10pt}
\end{minipage}
\begin{minipage}{.24\textwidth}
  \centering
  \begin{center}
    \includegraphics[width=1.\textwidth]{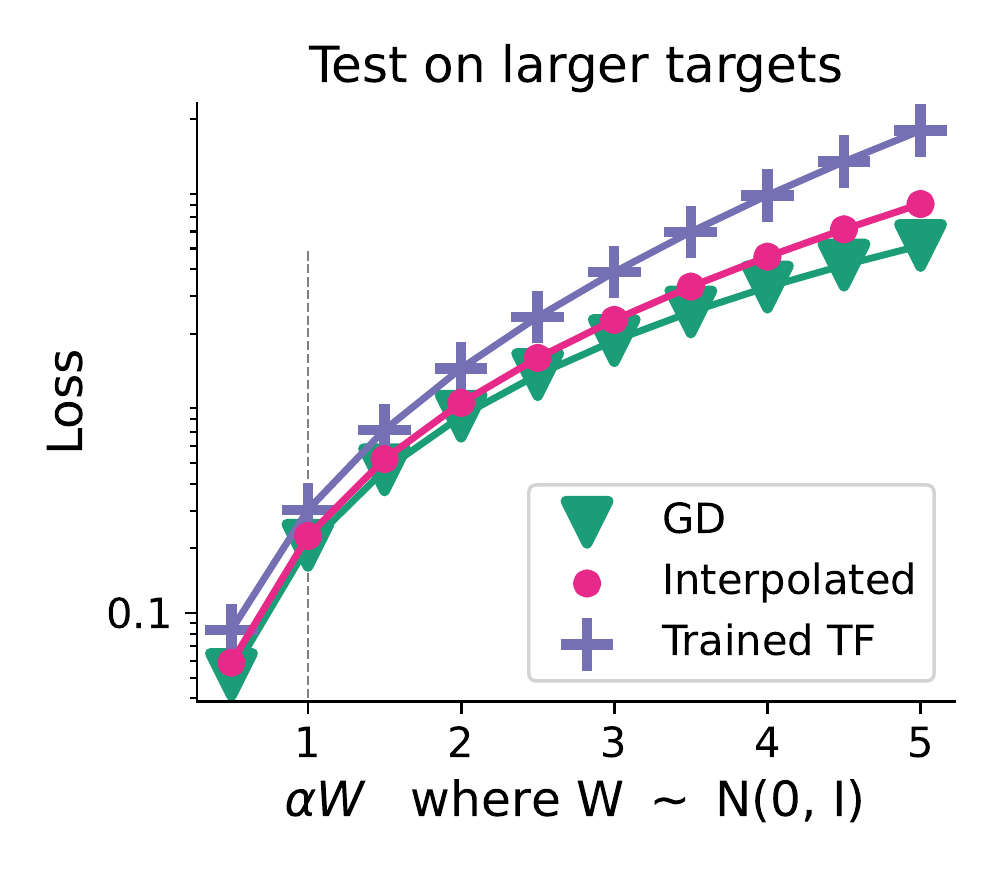}
  \end{center}
  \vspace{-10pt}
\end{minipage}
\end{center}

\textbf{(c) Comparing 1 step of gradient descent with training a LSA-layer on 2048 tasks.}

\begin{center}
\begin{minipage}{.24\textwidth}
  \centering
  \begin{center}
    \includegraphics[width=1.\textwidth]{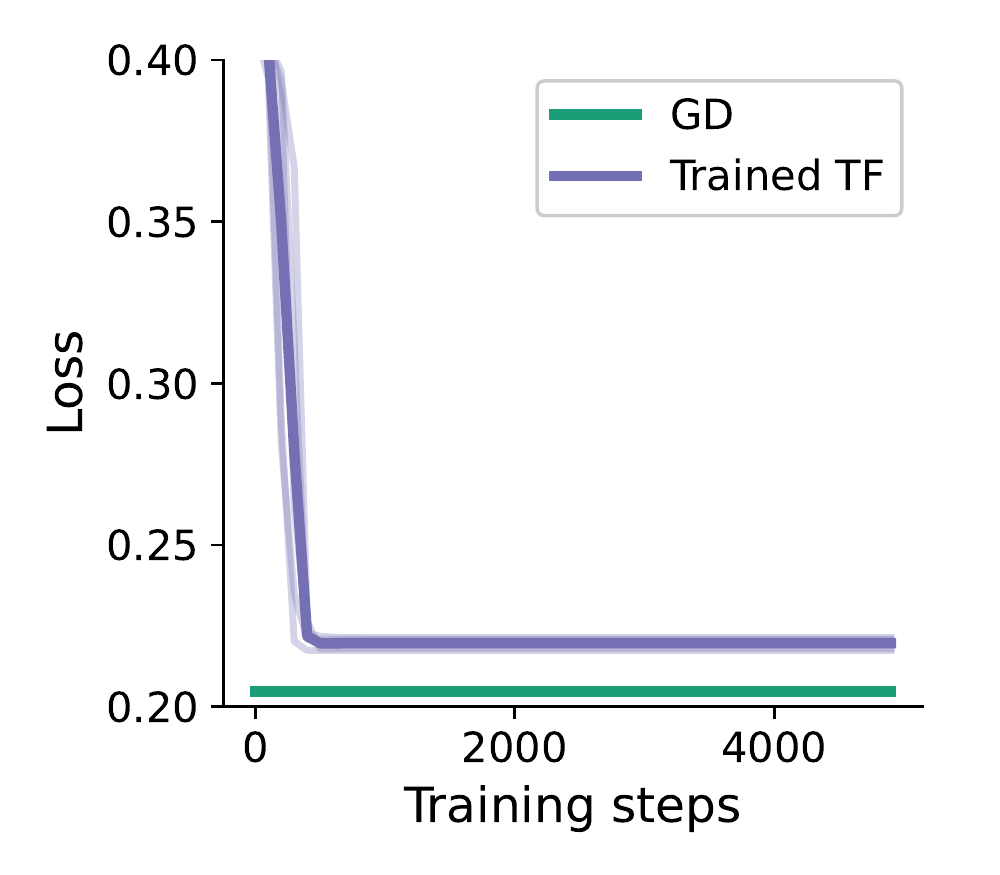}
  \end{center}
  \vspace{-10pt}
\end{minipage}
\begin{minipage}{.24\textwidth}
  \centering
  \begin{center}
    \includegraphics[width=1.\textwidth]{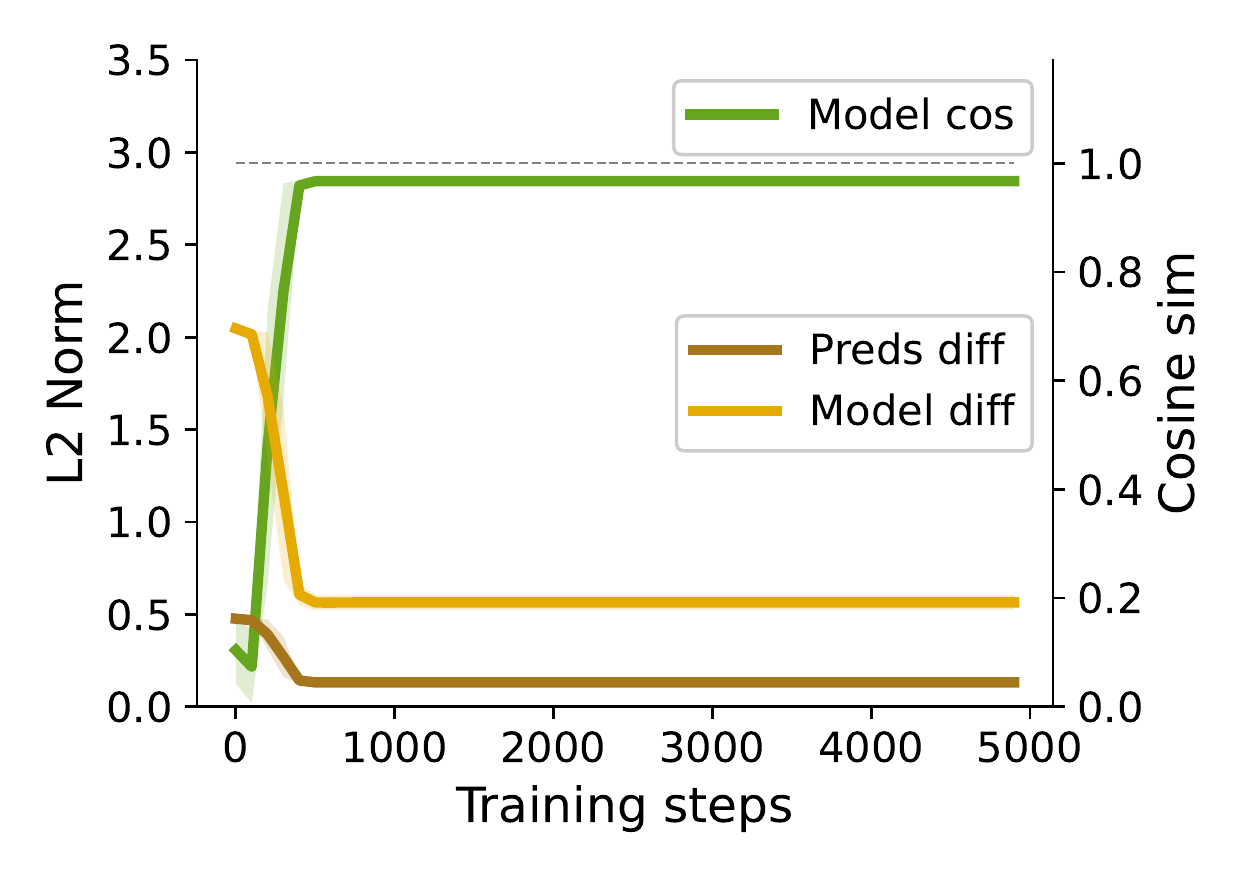}
  \end{center}
  \vspace{-10pt}
\end{minipage}
\begin{minipage}{.24\textwidth}
  \centering
  \begin{center}
    \includegraphics[width=1.\textwidth]{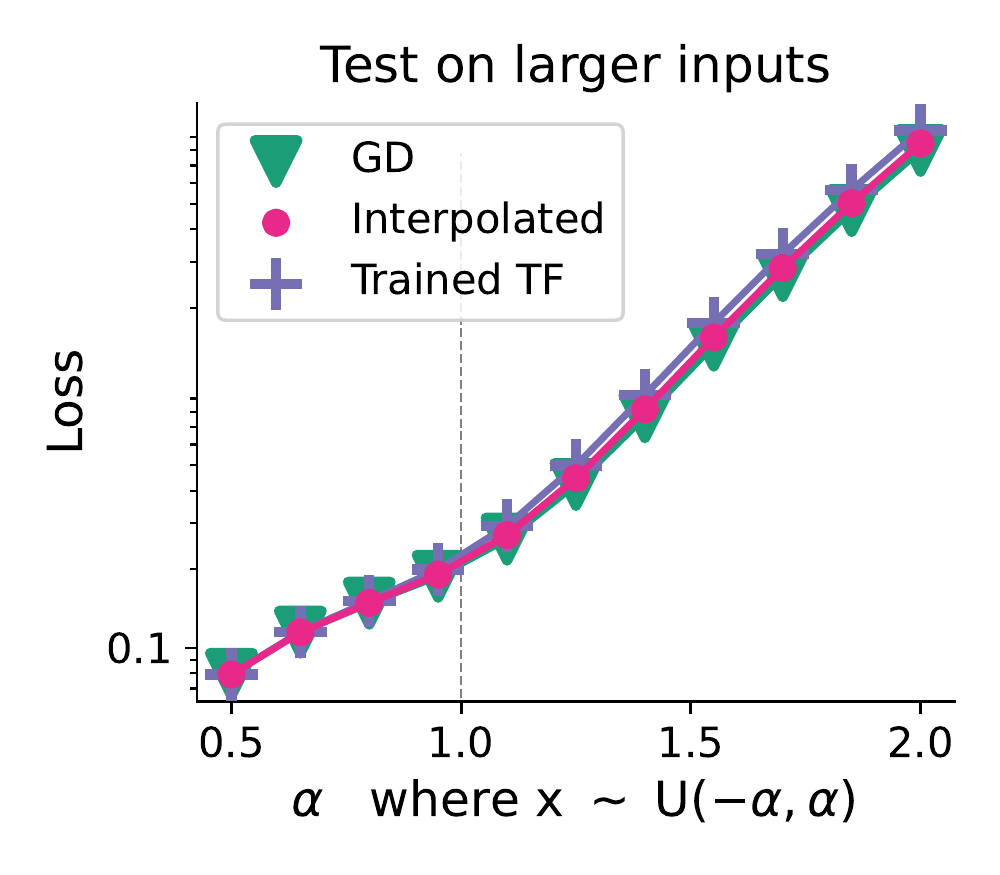}
  \end{center}
  \vspace{-10pt}
\end{minipage}
\begin{minipage}{.24\textwidth}
  \centering
  \begin{center}
    \includegraphics[width=1.\textwidth]{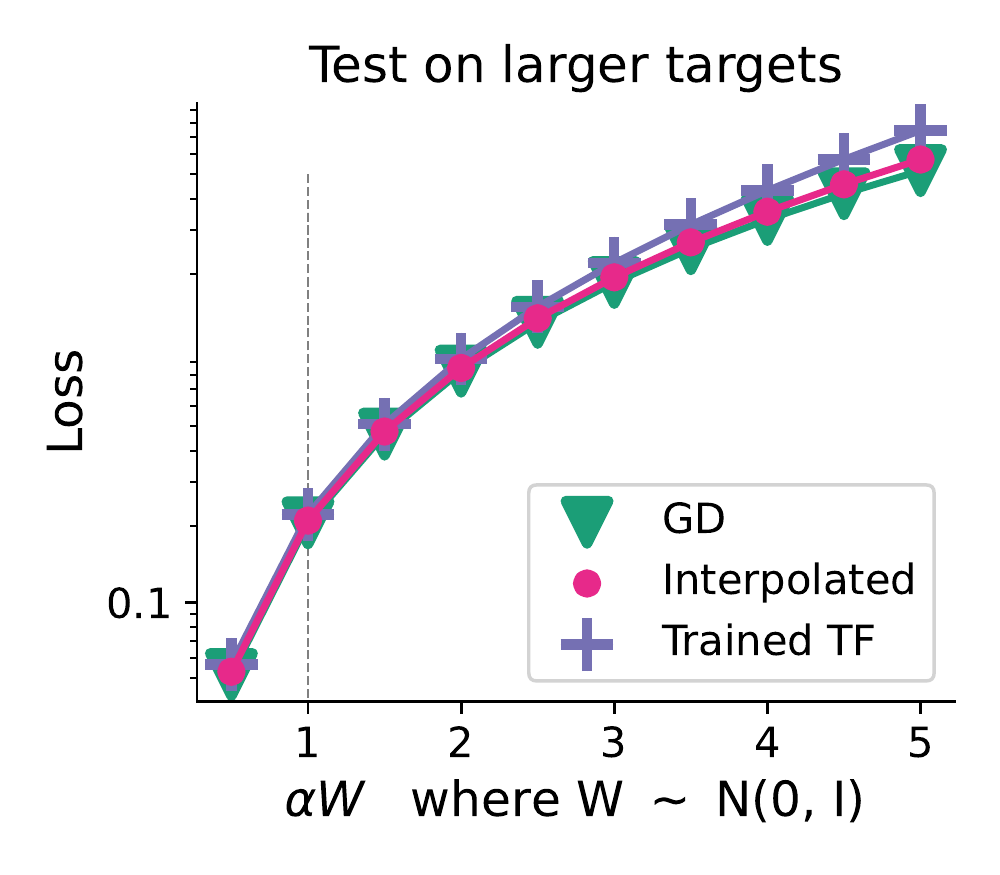}
  \end{center}
  \vspace{-10pt}
\end{minipage}
\end{center}

\textbf{(d) Comparing 1 step of gradient descent training a LSA-layer on 8192 tasks.}
\begin{center}
\begin{minipage}{.24\textwidth}
  \centering
  \begin{center}
    \includegraphics[width=1.\textwidth]{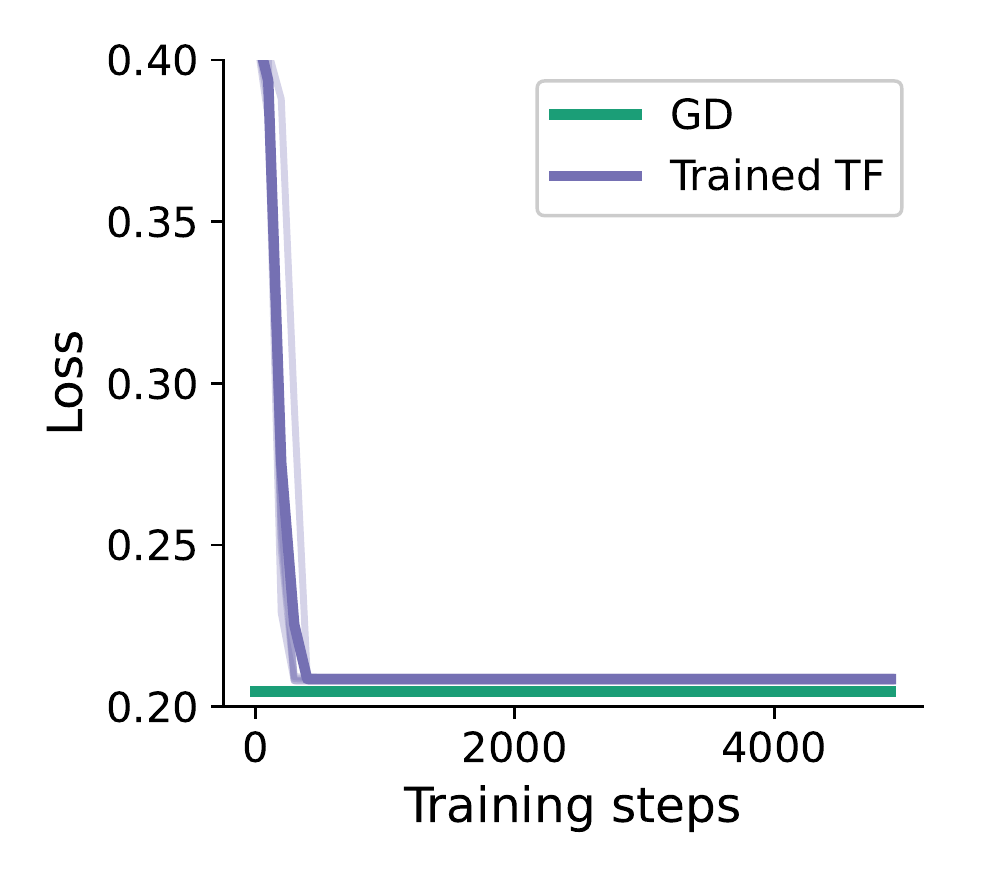}
  \end{center}
  \vspace{-10pt}
\end{minipage}
\begin{minipage}{.24\textwidth}
  \centering
  \begin{center}
    \includegraphics[width=1.\textwidth]{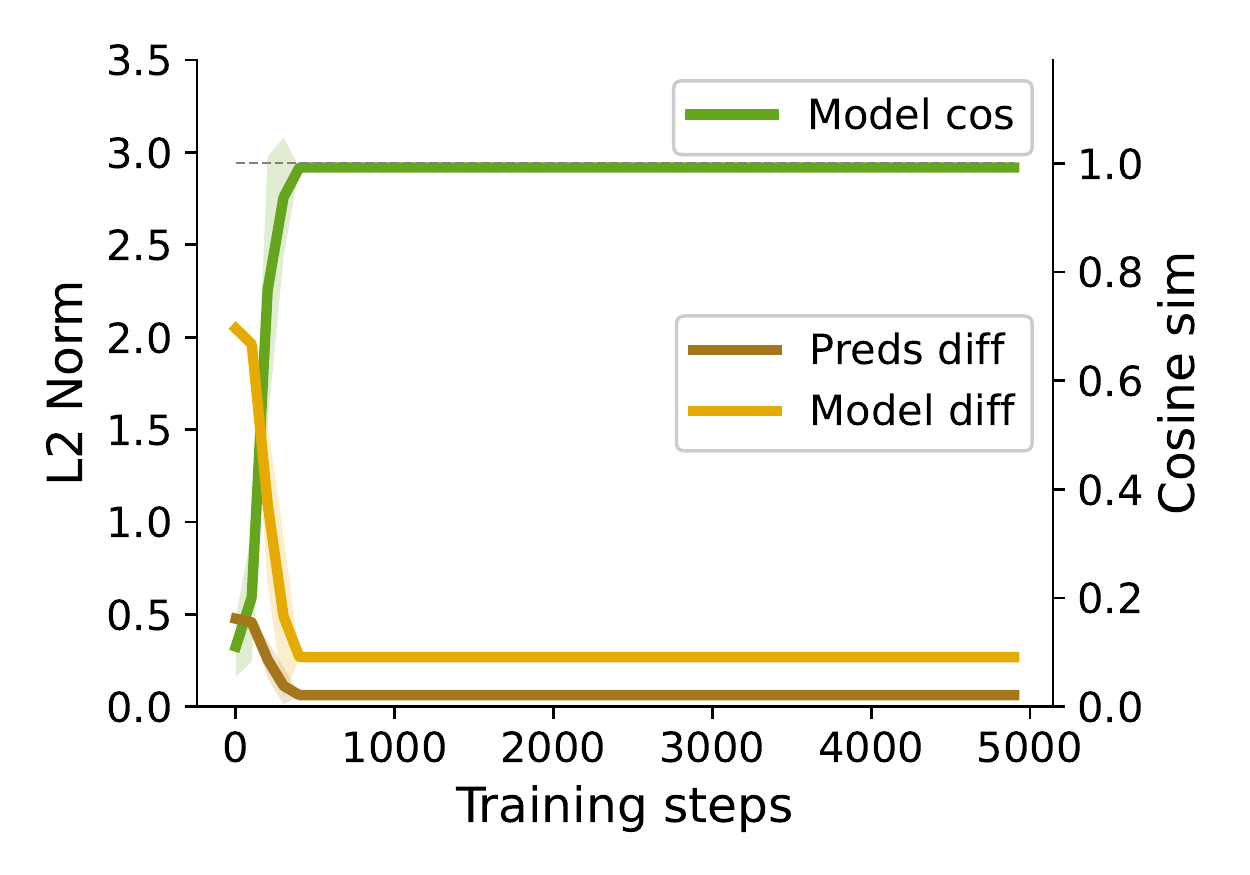}
  \end{center}
  \vspace{-10pt}
\end{minipage}
\begin{minipage}{.24\textwidth}
  \centering
  \begin{center}
    \includegraphics[width=1.\textwidth]{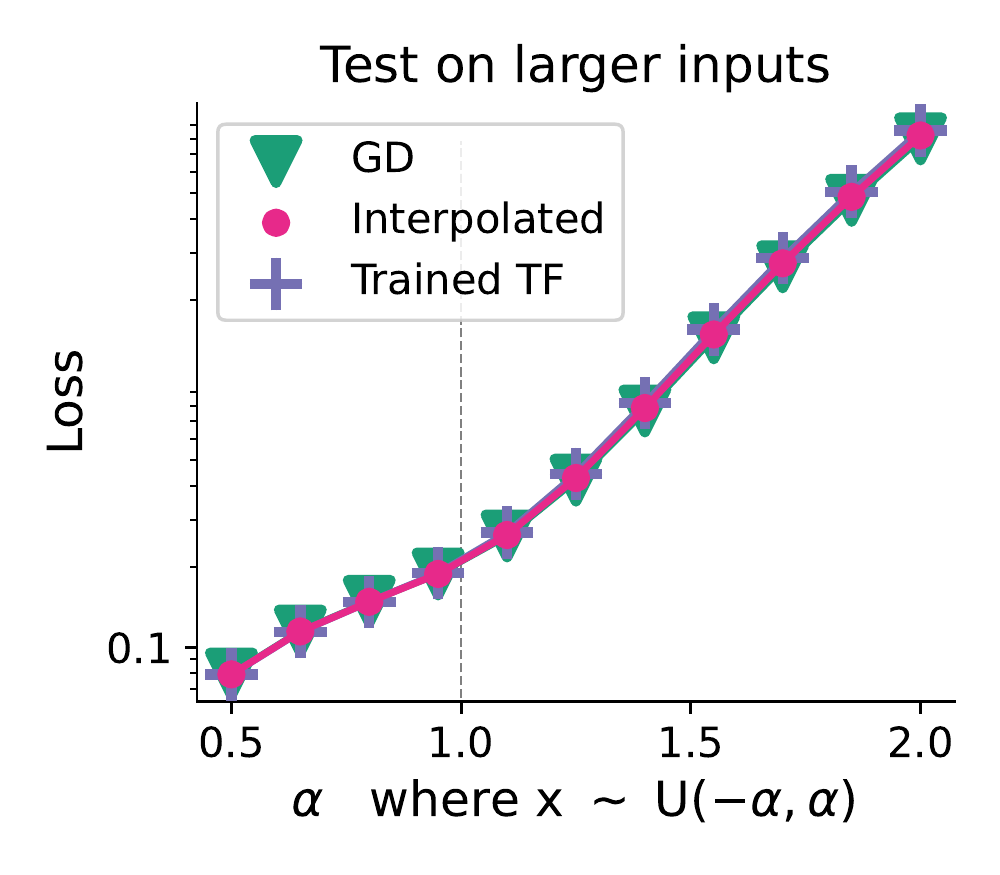}
  \end{center}
  \vspace{-10pt}
\end{minipage}
\begin{minipage}{.24\textwidth}
  \centering
  \begin{center}
    \includegraphics[width=1.\textwidth]{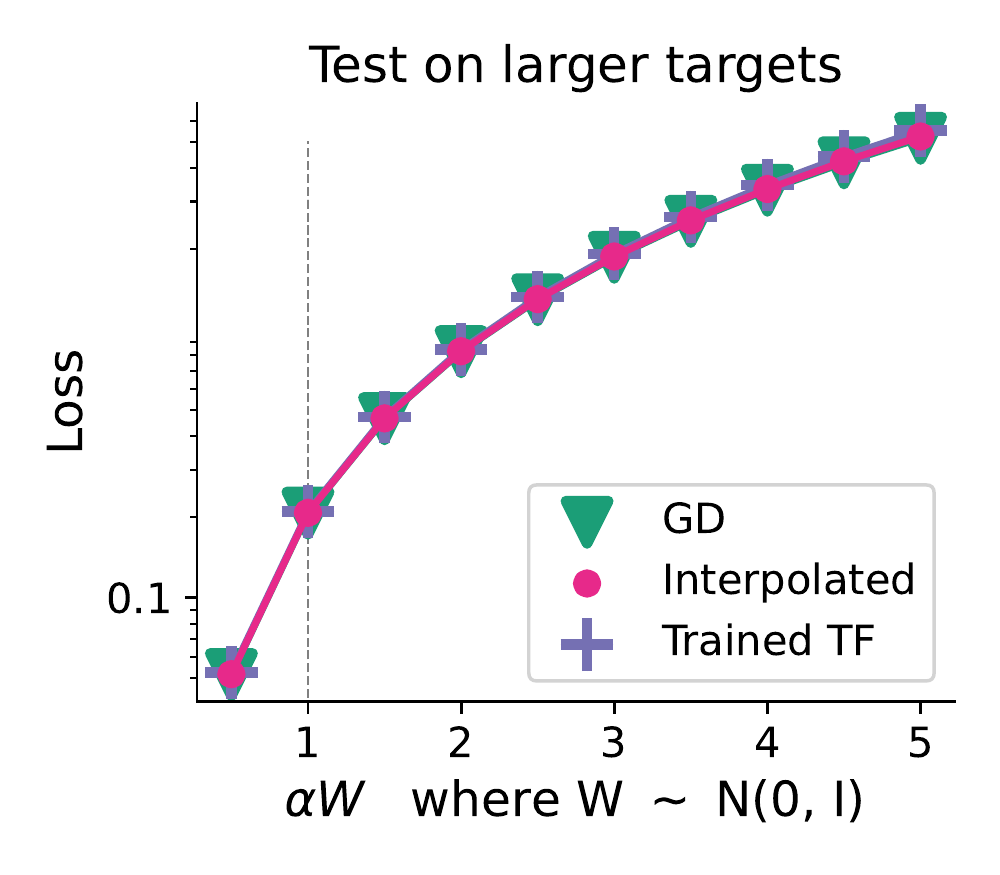}
  \end{center}
  \vspace{-10pt}
\end{minipage}
\end{center}
\vspace{-3pt}
  \caption{ \textbf{Comparing trained Transformers with GD and their weight interpolation when training the Transformer on a fixed training set size $B$}. Across our alignment measures as well as our tests on out-of-training behaviour, trained Transformers fail to align with GD when provided with a very small amount of tasks. Nevertheless, we see already almost perfect alignment in our base setting $N=N_x=10$ when provided with $B>2048$ tasks. In all settings, we train the Transformer on (non-stochastic) gradient descent iterating over a single batch of tasks of size $B$ equal to the number provided in the Figure titles (128, 512, 2048, 8192).}
  \label{fig:cycling}
  \vspace{-10pt}
\end{figure*}

Inspired by \cite{simple_case_study}, we additionally provide results when training a single linear self-attention layer on a fixed number of training tasks. Therefore, we iterate over a single fixed batch of size $B$ instead of drawing new batch of tasks at every iteration. Results can be found in Figure \ref{fig:cycling}. Intriguingly, we find that (meta-)gradient descent finds Transformer weights that align remarkable well with the provided construction and therefore gradient descent even when provided with an arguably very small number of training tasks. We argue that this again highlights the strong inductive bias of the LSA-layer to match (approximately) gradient descent learning in its forward pass.

\end{document}

%% file: math_commands.tex
\usepackage{amsmath,amsfonts,bm}

\newcommand{\sattn}{\text{SA}}
\newcommand{\lsattn}{\text{LSA}}

\def\eqref#1{equation~\ref{#1}}

\def\1{\bm{1}}

\DeclareMathAlphabet{\mathsfit}{\encodingdefault}{\sfdefault}{m}{sl}
\SetMathAlphabet{\mathsfit}{bold}{\encodingdefault}{\sfdefault}{bx}{n}

